\documentclass{article}

\usepackage[preprint,nonatbib]{neurips_2022}

\usepackage[utf8]{inputenc} 
\usepackage[T1]{fontenc}    
\usepackage{hyperref}       
\usepackage{url}            
\usepackage{booktabs}       
\usepackage{amsfonts}       
\usepackage{nicefrac}       
\usepackage{microtype}      
\usepackage{xcolor}         
\usepackage{caption} 
\usepackage{graphicx}
\usepackage{booktabs} 
\usepackage{amssymb}
\usepackage{amsmath}
\usepackage{multirow}
\usepackage{float}
\usepackage{mathtools}
\usepackage{bbold}
\usepackage{subcaption}

\usepackage{amsthm}

\newcommand{\norm}[1]{\left\lVert#1\right\rVert}

\usepackage{rotating,amsmath}

\definecolor{Highlight}{HTML}{39b54a}  

\newcommand{\del}[1]{\textcolor{Highlight}{{\small\textbf{#1}}}}
\newcommand{\dellarge}[1]{\textcolor{Highlight}{{{#1}}}}
\usepackage{pifont}
\newcommand{\cmark}{\ding{51}}%
\newcommand{\xmark}{\ding{55}}%

\author{}
\date{}
\title{Optimizing Relevance Maps of Vision\\ Transformers Improves Robustness}

\author{%
  Hila Chefer \quad Idan Schwartz \quad Lior Wolf\\
  School of Computer Science\\
  Tel-Aviv University\\
}

\begin{document}

\maketitle

\begin{abstract}

It has been observed that visual classification models often rely mostly on the image background, neglecting the foreground, which hurts their robustness to distribution changes.  
To alleviate this shortcoming, we propose to monitor the model's relevancy signal and manipulate it such that the model is focused on the foreground object.
This is done as a finetuning step, involving relatively few samples consisting of pairs of images and their associated foreground masks. Specifically, we encourage the model's relevancy map (i) to assign lower relevance to background regions, (ii) to consider as much information as possible from the foreground, and (iii) we encourage the decisions to have high confidence. When applied to Vision Transformer (ViT) models, a marked improvement in robustness to domain shifts is observed. Moreover, the foreground masks can be obtained automatically, from a self-supervised variant of the ViT model itself; therefore no additional supervision is required.
Our code is available at \url{https://github.com/hila-chefer/RobustViT}.
\end{abstract}

\section{Introduction}

\begin{figure*}[t]
  \centering
\begin{tabular}{c@{~~~~~~~~~~}c@{~~~~~~~~~~}c@{~}c}
ViT-base~\cite{dosovitskiy2020image} & AR-base~\cite{Steiner2021HowTT} & DeiT-base~\cite{touvron2020training}\\
~\\
\includegraphics[width=0.3\linewidth, clip]{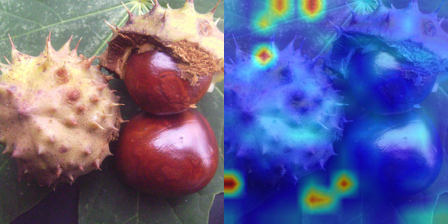}&
\includegraphics[width=0.3\linewidth, clip]{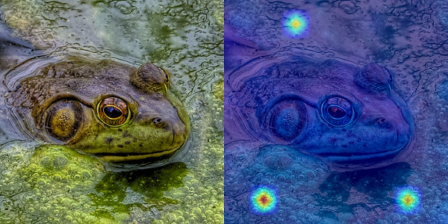}&
\includegraphics[width=0.3\linewidth, clip]{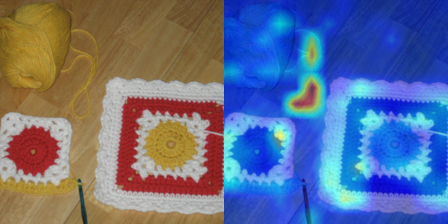}
\\
\small{Classification: Chestnut} &
\small{Classification: Bullfrog} &
\small{Classification: Dishcloth}
\\
\small{Confidence: $100\%$} &
\small{Confidence: $99.1\%$}
&
\small{Confidence: $91.6\%$}
\\
\includegraphics[width=0.3\linewidth, clip]{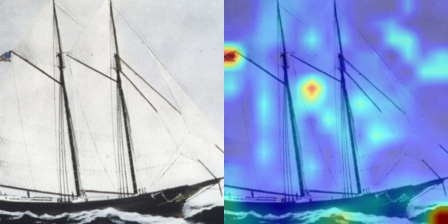}&
\includegraphics[width=0.3\linewidth, clip]{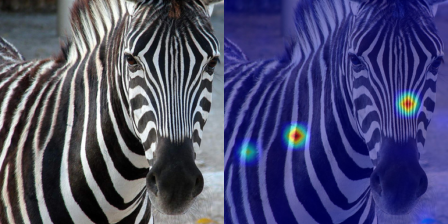}&

\includegraphics[width=0.3\linewidth, clip]{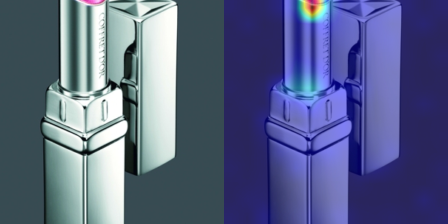}
\\
\small{Classification: Schooner} &
\small{Classification: Zebra} &
\small{Classification: Lipstick}
\\
\small{Confidence: $97\%$} &
\small{Confidence: $99.9\%$} &
\small{Confidence: $96.1\%$}
\end{tabular}
\caption{Examples of the salient issues with ViTs. Each pair depicts an input image and its corresponding relevance map. The first row demonstrates examples of background-centered relevance, the second row shows examples of sparse foreground relevance. Both issues occur in all models- ViT~\cite{dosovitskiy2020image}, ViT AugReg~\cite{Steiner2021HowTT} (AR), and DeiT~\cite{touvron2020training}, even if the confidence of the model is above $90\%$.
\label{fig:motivation}}
\vspace{-14px}
\end{figure*}

The reliance on simple image-level classification supervision, together with the sampling biases of object recognition datasets, leads to vision models that exhibit unintuitive behavior, as depicted in  Fig~\ref{fig:motivation}. First, the models we tested (ViT~\cite{dosovitskiy2020image}, ViT AugReg~\cite{Steiner2021HowTT}, and DeiT~\cite{touvron2020training}) tend to give disproportional high weight to the background of the image in the decision-making process.  
Second, the tested models occasionally regard a sparse subset of the pixels in the foreground object for the classification, disregarding much of the object's data. As argued by Geirhos et al.~\cite{Geirhos2020ShortcutLI}, and stated in~\cite{Hendrycks_2021_CVPR} ``image classification datasets contain `spurious cues' or `shortcuts' . For instance, cows tend to co-occur with green pastures, and even though the background is inessential to the identity of the object, models may predict `cow', using primarily the green pasture background cue.''

There is considerable evidence that context is a useful cue~\cite{lapuschkin2019unmasking, rosenfeld2018elephant}. However, many of the associated background elements and foreground shortcuts are only relevant to the specific data distribution, which leads to lack of robustness to distribution shifts~\cite{nguyen2014deep,shetty2019not}. There are many methods for overcoming domain shifts, including domain adaptation techniques~\cite{ganin2016domain,arjovsky2020invariant} and methods that augment the training set or the training procedure~\cite{ilyas2019adversarial,nuriel2021permuted}. In this work, however, we opt for a direct approach, which monitors the relevancy score of the model for each image region, and manipulates the relevancy map to be focused on the regions within the foreground mask.

The method is based on a finetuning procedure, which is applied to a pretrained Vision Transformer (ViT) model. A relatively small set of samples, for which the foreground is given, is employed during this phase. In most of our experiments, we use three samples for half the classes, following work that examined the effect of transfer learning on half of the classes~\cite{yosinski2014transferable}. The ground-truth foreground mask is either human annotated~\cite{gao2021luss}, or estimated from a self-supervised ViT model~\cite{wang2022tokencut}.

The finetuning procedure employs three loss terms. The first encourages the relevance map to assign lower values to the background of the image. The second term aims to remedy the sparse relevance issue by making larger parts of the foreground part of the relevancy map. The third term is a regularizer that ensures that the classification accuracy of the original model remains unimpaired.

Unsurprisingly, applying this method leads to a (modest) drop in accuracy on the original training dataset and on datasets with very similar distributions. 
In an extensive battery of experiments we show that (i)  the classification accuracy on datasets from shifted domains increases considerably. This includes real-world unbiased and adversarial datasets, as well as synthetic ones that were created specifically to measure the robustness of the classification model, (ii) the resulting relevance maps demonstrate a significant improvement in focusing on the foreground of the image, i.e. the object, rather than on its background.

\section{Related Work}

Image classification datasets are becoming increasingly challenging, while at the same time models are growing more complex~\cite{Deng2009ImageNetAL, krizhevsky2012imagenet,7298594, resnets}. With the rapid advancements in object recognition, the measuring stick is being narrowed down to a single number, accuracy~\cite{torralba2011unbiased}. By relying solely on accuracy, classifiers introduce biases, since they utilize shortcuts to select the right class~\cite{lapuschkin2019unmasking, Geirhos2020ShortcutLI}. The shortcomings of models that rely on shortcuts have been demonstrated for domains other than vision, such as natural language processing~\cite{feng2018pathologies} and multi-modal learning~\cite{gat2021perceptual}.

One way to assess the salient behavior of a model is to study Sufficient Input Subsets (SIS), i.e., the minimal number of pixels necessary for a confident prediction~\cite{carter2019made,overinterpretation}. Finding an SIS for a class can imply that the classifier has overinterpreted its input, since it can make a confident accurate decision using a small, sparse subset of pixels, which does not seem meaningful to humans. We study the SIS with gradients approach for ViT models, and find that it can be misleading. Specifically, SIS can be regarded as an adversarial method that can lead to high-confidence classification of {\em any label} from a sparse set of pixels, see Appendix~\ref{sec:sis}. Therefore, we opt to use datasets designed specifically for evaluating accuracy when facing distribution shifts, to assess the model's resilience and ability to generalize. Several alternatives have been proposed to ImageNet~\cite{ILSVRC15}: (i) ImageNet-v2~\cite{recht2019imagenet}, a new test set sampled from the same distribution, which reduces adaptive overfitting; (ii) ImageNet-A~\cite{Hendrycks_2021_CVPR}, a test set of natural adversarial samples; (iii) ImageNet-R~\cite{Hendrycks_2021_ICCV} and ImageNet-Sketch~\cite{NEURIPS2019_3eefceb8}, which contain renditions of objects (e.g., art, sculptures, sketches); (iv) ObjectNet~\cite{barbu2019objectnet}, a real-world set with controls on object locations and view points, and (v) SI-Score~\cite{Djolonga2021OnRA}, which is a synthetic dataset designed specifically for testing robustness to object location, rotation and size. 

Explainability methods may be used to determine the reasons for the decisions made by classifiers. As an example, we may find that models tend to overlook objects with relevance maps (see Fig.~\ref{fig:motivation}). Gradients are a dominant and useful signal for model interpretation~\cite{simonyan2014deep, dabkowski2017real, mahendran2016visualizing}. By adding input signals to the gradients, relevance maps were refined~\cite{gu2018recent, shrikumar2017learning, smilkov2017smoothgrad, srinivas2019full}. Alternatives to gradient propagation for explanation include attribution propagation - a theory-driven method based on axioms, and permutation based on Shapley values\cite{bach2015pixel, montavon2017explaining, shrikumar2017learning, lundberg2017unified}. For transformer architectures, the combination of gradients and attention values has been shown to produce a viable interpretation of the model's prediction~\cite{Chefer_2021_ICCV, chefer2020transformer}.

Our method optimizes the relevancy maps of the model as a regularization term. Other works that investigated the use of relevancy to alleviate overfitting include Ross et al.~\cite{Ross2017RightFT}, who introduced a regularization term for the input gradient, which reduces reliance on irrelevant cues, e.g., background pixels. Additional work in this vein has been conducted on medical data, studying how doctors classified a disease~\cite{Simpson2019GradMaskRO, viviano2019saliency}. Singh et al.~\cite{singh2020don} regularize feature representations of a category from its co-occurring context. Zhu et al.~\cite{zhu2019deformable} enrich classifier representations by mimicking detection models. Importantly, unlike all the above methods, our method incorporates the foreground features and the classifier confidence, rather than considering only the background features. Furthermore, we apply our method not during training, but as a short finetuning process that is feasible for large models.

\section{Method}
\label{sec:method}

Our approach aims to direct vision models such that their decision will be based on the features of the object rather than on other supportive background features. To achieve this, we employ additional supervision to distinguish between the foreground and background features. The method finetunes the model in a way that encourages the class relevance map, obtained through a relevance computation method, to roughly resemble the segmentation map. This way, the decision-making process is focused on the foreground. 
The relevance map employed by our method is calculated using a recent advancement in explainability for transformer-based architectures~\cite{Chefer_2021_ICCV}. A brief introduction of the explainability method used is provided in Appendix~\ref{sec:background}.

The method employs a small set of labeled segmentation maps for distinguishing between the foreground and background of the input image. Our first loss term discourages the model from considering mostly the background:
\begin{align}
    \mathcal{L}_{\text{bg}} = \text{MSE}\left(\mathbf{R}(i) \odot \bar{\mathbf{S}}(i), 0\right),
    \label{eq:background}
\end{align}
where $i$ is the input image, $\mathbf{R}(i)$ is the relevance map produced for $i$, $\bar{\mathbf{S}}(i)$ is the inverse of the segmentation map for $i$, and $\odot$ is the Hadamard product. 
Put differently, $\mathcal{L}_{\text{bg}}$ extracts the relevance values assigned to the background using the provided segmentation, and encourages those values to be close to $0$, which is the minimal possible relevance value.

Our second loss term encourages the model to consider as much information as possible from the foreground of the image:
\begin{align}
    \mathcal{L}_{\text{fg}} = \text{MSE}\left(\mathbf{R}(i) \odot {\mathbf{S}}(i), 1\right),
    \label{eq:foreground}
\end{align}
where ${\mathbf{S}}(i)$ is the foreground mask. This loss encourages the relevance of pixels inside the segmentation to be higher ($1$ is the maximal achievable relevance value). The overall explainability loss is constructed as follows:
\begin{align}
    \mathcal{L}_{\text{relevance}} = \lambda_{\text{bg} }\cdot\mathcal{L}_{\text{bg}} + \lambda_{\text{fg}}\cdot\mathcal{L}_{\text{fg}},
    \label{eq:expl-loss}
\end{align}
where $\lambda_{\text{bg}}, \lambda_{\text{fg}}$ are hyperparameters. All our experiments apply the same choice of  $\lambda_{\text{bg}} = 2, \lambda_{\text{fg}} = 0.3$. Note that the coefficient $\lambda_{\text{fg}}$ is much smaller than $\lambda_{\text{bg}}$. The reason is two-fold: (i) we find that the issue of overinterpreting the background is more common than the issue of partial relevance of foreground pixels, and (ii) $\mathcal{L}_{\text{fg}}$ implies a uniform relevance of $1$ for all foreground pixels, which may be detrimental, as we wish to allow the model to be able to focus on specific features of the object.   

Finally, in the absence of an additional regularization loss, the finetuning results in explanations that resemble the ground-truth segmentation, while the accuracy plummets due to the absence of encouragement to maintain high accuracy. 
Therefore, one must apply an additional loss term to ensure that the output distribution of the model remains similar to the original model. We opt to use a confidence-boosting loss for this purpose, which is constructed as follows:
\begin{align}
    \mathcal{L}_{\text{classification}} =\text{CE}\left(\mathcal{M}(i) , \arg\max(\mathcal{M}(i))\right),
    \label{eq:cls-loss}
\end{align}
where $\mathcal{M}$ notates the vision model, and $\arg\max(\mathcal{M}(i))$ is the class predicted by $\mathcal{M}$ for the input image $i$. $\mathcal{L}_{\text{classification}}$ calculates the cross-entropy loss between the output distribution of $\mathcal{M}$ and the one-hot distribution where the predicted class is assigned a probability of $1$. In other words, this loss encourages the confidence of the predicted class to increase.

The overall loss for the finetuning process is, therefore:
\begin{align}
    \mathcal{L} = \lambda_{\text{relevance}}\cdot\mathcal{L}_{\text{relevance}} + \lambda_{\text{classification}}\cdot\mathcal{L}_{\text{classification}},
    \label{eq:overall-loss}
\end{align}
where $\lambda_{\text{relevance}}=0.8$, and $\lambda_{\text{classification}}=0.2$ remain constant in all our experiments.

\section{Experiments}

\begin{figure*}[t!]
  \centering
  \resizebox{\linewidth}{!}{
\begin{tabular}{c@{~~}c@{~}c@{~}c@{~~~}c@{~}c@{~}c@{~~~}c@{~}c@{~}c}
& \multicolumn{3}{c}{Same prediction}& \multicolumn{3}{c}{Corrected prediction} & \multicolumn{3}{c}{Incorrect prediction}\\
\cmidrule(r){2-4}
\cmidrule(r){5-7}
\cmidrule(r){8-10}
& \small{Input} & \small {Original} & {\small{Ours}} &  \small{Input} & \small {Original} & {\small{Ours}} &  \small{Input} & \small {Original} & {\small{Ours}} \\
{\begin{turn}{90}~~~ ViT-B \end{turn}} & 
\includegraphics[width=0.1\linewidth, clip]{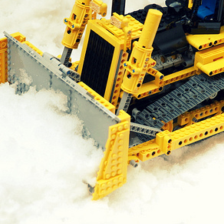}&
\includegraphics[width=0.1\linewidth, clip]{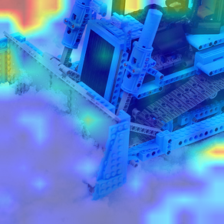}&
\includegraphics[width=0.1\linewidth, clip]{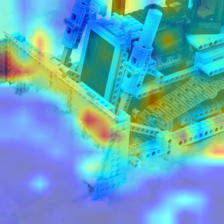}&
\includegraphics[width=0.1\linewidth, clip]{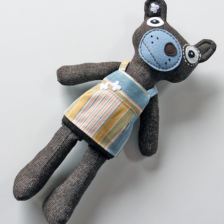}&
\includegraphics[width=0.1\linewidth, clip]{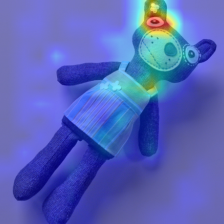}&
\includegraphics[width=0.1\linewidth, clip]{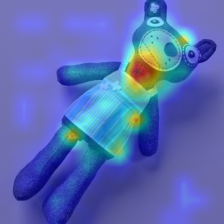}&
\includegraphics[width=0.1\linewidth, clip]{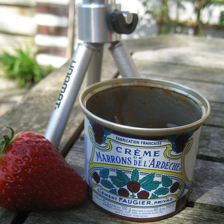}&
\includegraphics[width=0.1\linewidth, clip]{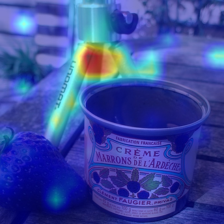}&
\includegraphics[width=0.1\linewidth, clip]{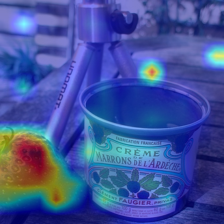}
\\
\multirow{2}{*}{\begin{turn}{90} Pred \end{turn}}
&
&{\small{Snowplow}} & {\small{Snowplow}} & & {\small{Can-}} & {\small{Teddy-}} &  & {\small{Tripod}} & {\small{Strawberry}}\\
& & & & & {\small{opener}}& {\small{bear}} &  & \\
{\begin{turn}{90}~~~ ViT-L \end{turn}} & 
\includegraphics[width=0.1\linewidth, clip]{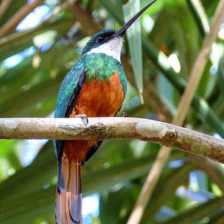}&
\includegraphics[width=0.1\linewidth, clip]{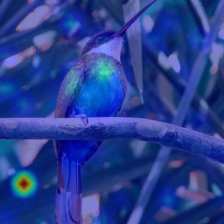}&
\includegraphics[width=0.1\linewidth, clip]{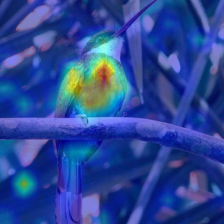}&
\includegraphics[width=0.1\linewidth, clip]{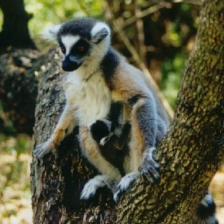}&
\includegraphics[width=0.1\linewidth, clip]{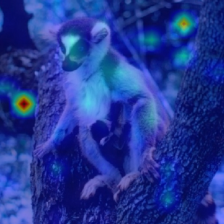}&
\includegraphics[width=0.1\linewidth, clip]{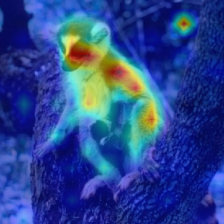}&
\includegraphics[width=0.1\linewidth, clip]{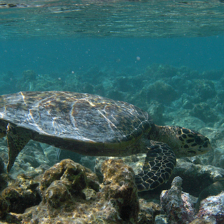}&
\includegraphics[width=0.1\linewidth, clip]{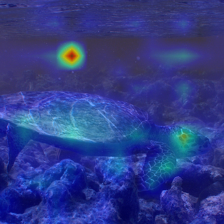}&
\includegraphics[width=0.1\linewidth, clip]{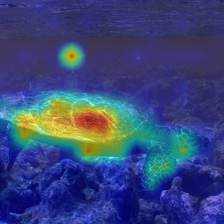}
\\
\multirow{2}{*}{\begin{turn}{90} Pred \end{turn}}
&
&{\small{Jacamar}} & {\small{Jacamar}} & & {\small{Indri}} & {\small{Lemur-}} &  & {\small{Coral-}} & {\small{Leather-}}\\
& & & & & {\small{}} & {\small{catta}} & & {\small{reef}} & {\small{back turtle}} 
\\
{\begin{turn}{90}~~~ AR-B \end{turn}} & 
\includegraphics[width=0.1\linewidth, clip]{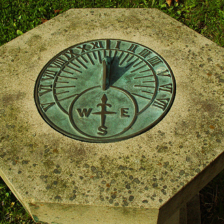}&
\includegraphics[width=0.1\linewidth, clip]{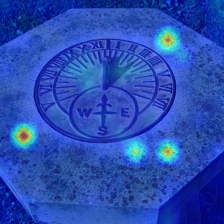}&
\includegraphics[width=0.1\linewidth, clip]{{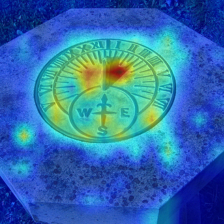}}&
\includegraphics[width=0.1\linewidth, clip]{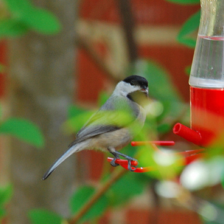}&
\includegraphics[width=0.1\linewidth, clip]{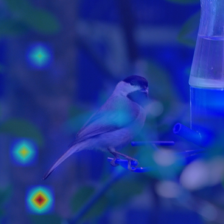}&
\includegraphics[width=0.1\linewidth, clip]{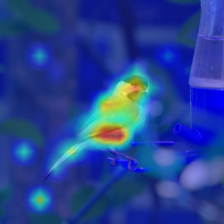}&
\includegraphics[width=0.1\linewidth, clip]{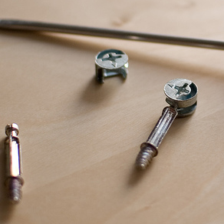}&
\includegraphics[width=0.1\linewidth, clip]{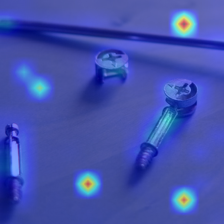}&
\includegraphics[width=0.1\linewidth, clip]{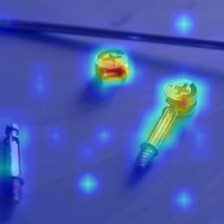}
\\
\multirow{2}{*}{\begin{turn}{90} Pred \end{turn}}
&
&{\small{Sundial}} & {\small{Sundial}} & & {\small{Humming-}} & {\small{Chickadee}} &  & {\small{Screwdr-}} & {\small{Screw}}
\\
& & & & &  {\small{bird}} & {\small{}} & &{\small{iver}} & {\small{}}  \\
{\begin{turn}{90}~~~ AR-L \end{turn}} & 
\includegraphics[width=0.1\linewidth, clip]{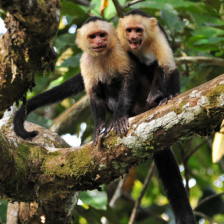}&
\includegraphics[width=0.1\linewidth, clip]{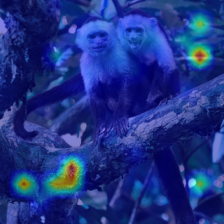}&
\includegraphics[width=0.1\linewidth, clip]{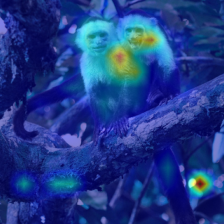}&
\includegraphics[width=0.1\linewidth, clip]{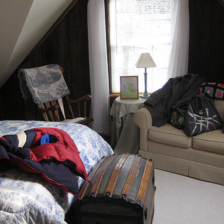}&
\includegraphics[width=0.1\linewidth, clip]{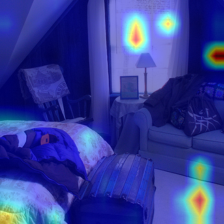}&
\includegraphics[width=0.1\linewidth, clip]{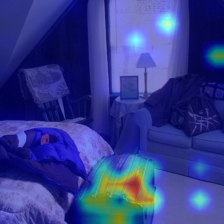}&
\includegraphics[width=0.1\linewidth, clip]{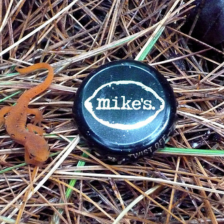}&
\includegraphics[width=0.1\linewidth, clip]{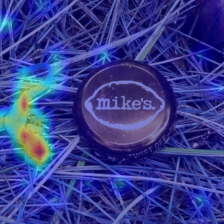}&
\includegraphics[width=0.1\linewidth, clip]{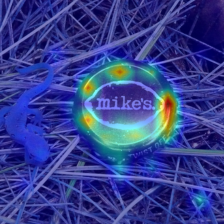}
\\
\multirow{2}{*}{\begin{turn}{90} Pred \end{turn}}
&
&{\small{Capuchin}} & {\small{Capuchin}} & & {\small{Quilt}} & {\small{Chest}} &  & {\small{Eft}} & {\small{Bottlecap}}\\
& & & & & & & & {\small{}} & {\small{}}
\\
{\begin{turn}{90}~ DeiT-B \end{turn}} & 
\includegraphics[width=0.1\linewidth, clip]{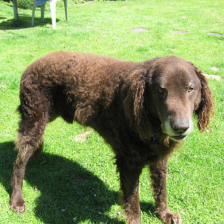}&
\includegraphics[width=0.1\linewidth, clip]{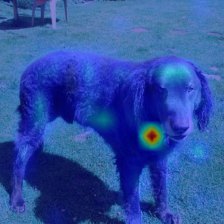}&
\includegraphics[width=0.1\linewidth, clip]{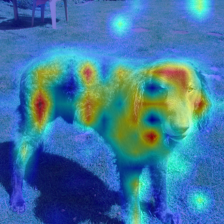}&
\includegraphics[width=0.1\linewidth, clip]{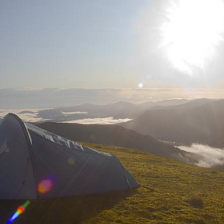}&
\includegraphics[width=0.1\linewidth, clip]{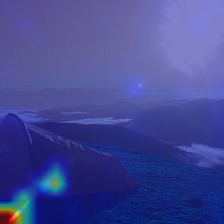}&
\includegraphics[width=0.1\linewidth, clip]{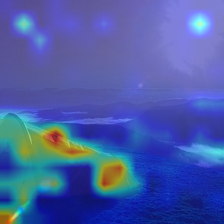}&
\includegraphics[width=0.1\linewidth, clip]{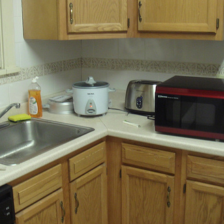}&
\includegraphics[width=0.1\linewidth, clip]{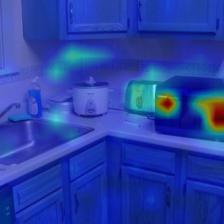}&
\includegraphics[width=0.1\linewidth, clip]{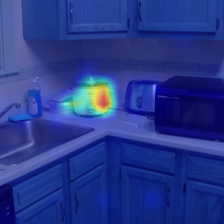}
\\
\multirow{2}{*}{\begin{turn}{90} Pred \end{turn}}
&
&{\small{Curly coat-}}  & {\small{Curly coat-}}  & & {\small{Bubble}} & {\small{Mountain-}} &  & {\small{Microwave}} & {\small{Crock-}}\\
& & {\small{ed retriever}} & {\small{ed retriever}} & {\small{}}& & {\small{tent}} & {\small{}} & & {\small{pot}}

\end{tabular}
}
\caption{Examples from the ImageNet validation set of cases where our method does not change the prediction, corrects the prediction, and ruins the prediction. Even in cases where our method changes a correct prediction, there is often a rationale  behind the modified prediction. The ``Pred" row specifies the predictions before and after our finetuning. The examples are presented for the base, large models of ViT~\cite{dosovitskiy2020image}, ViT AugReg~\cite{Steiner2021HowTT} (AR), and the base model of DeiT~\cite{touvron2020training}.
\label{fig:improved_expl}}
\end{figure*}
\begin{figure*}[t!]
  \centering
  \resizebox{\linewidth}{!}{
\begin{tabular}{c@{~~}c@{~}c@{~}c@{~~~}c@{~}c@{~}c@{~~~}c@{~}c@{~}c}
& \multicolumn{3}{c}{INet-A}& \multicolumn{3}{c}{ObjectNet} & \multicolumn{3}{c}{SI-rotation}\\
\cmidrule(r){2-4}
\cmidrule(r){5-7}
\cmidrule(r){8-10}
& \small{Input} & \small {Original} & {\small{Ours}} &  \small{Input} & \small {Original} & {\small{Ours}} &  \small{Input} & \small {Original} & {\small{Ours}} \\
{\begin{turn}{90}~~~ ViT-B \end{turn}} & 
\includegraphics[width=0.1\linewidth, clip]{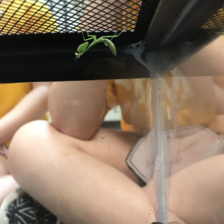}&
\includegraphics[width=0.1\linewidth, clip]{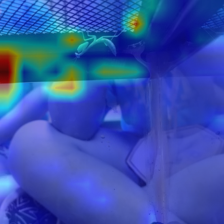}&
\includegraphics[width=0.1\linewidth, clip]{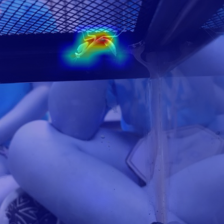}&
\includegraphics[width=0.1\linewidth, clip]{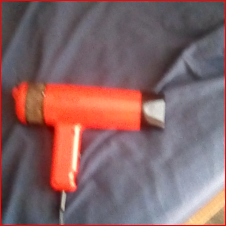}&
\includegraphics[width=0.1\linewidth, clip]{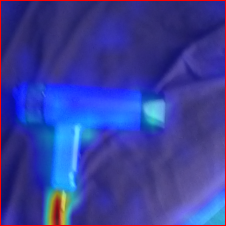}&
\includegraphics[width=0.1\linewidth, clip]{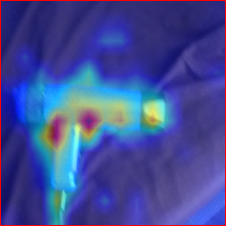}&
\includegraphics[width=0.1\linewidth, clip]{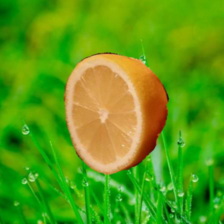}&
\includegraphics[width=0.1\linewidth, clip]{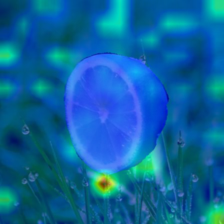}&
\includegraphics[width=0.1\linewidth, clip]{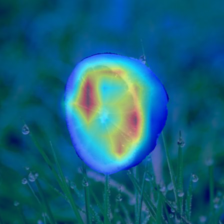}
\\
\multirow{2}{*}{\begin{turn}{90} Pred \end{turn}}
&
&{\small{Television}} & {\small{Mantis}} & & {\small{Drill}} & {\small{Hand-}} &  & {\small{Golf ball}} & {\small{Lemon}}\\
& & & & &  & {\small{blower}} & &  \\
{\begin{turn}{90}~~~ ViT-L \end{turn}} & 
\includegraphics[width=0.1\linewidth, clip]{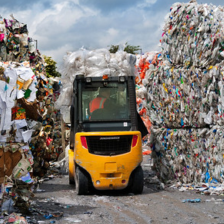}&
\includegraphics[width=0.1\linewidth, clip]{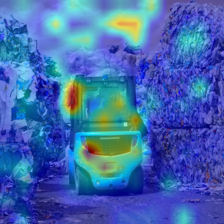}&
\includegraphics[width=0.1\linewidth, clip]{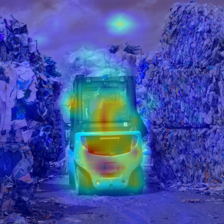}&
\includegraphics[width=0.1\linewidth, clip]{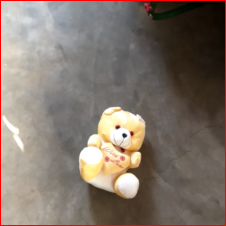}&
\includegraphics[width=0.1\linewidth, clip]{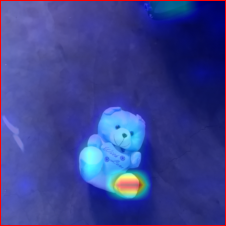}&
\includegraphics[width=0.1\linewidth, clip]{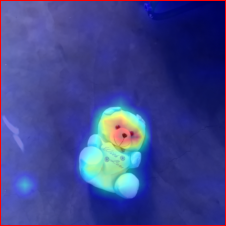}&
\includegraphics[width=0.1\linewidth, clip]{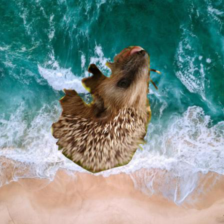}&
\includegraphics[width=0.1\linewidth, clip]{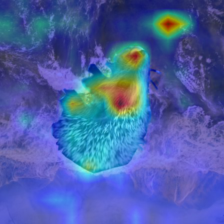}&
\includegraphics[width=0.1\linewidth, clip]{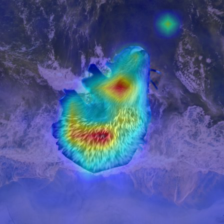}
\\
\multirow{2}{*}{\begin{turn}{90} Pred \end{turn}}
&
&{\small{Garbage-}} & {\small{Forklift}} & & {\small{Ping-pong-}} & {\small{Teddy}} &  & {\small{Sea Lion}} & {\small{Porcupine}}\\
& &{\small{truck}} & & & {\small{ball}} & & & 
\\
{\begin{turn}{90}~~~ AR-B \end{turn}} & 
\includegraphics[width=0.1\linewidth, clip]{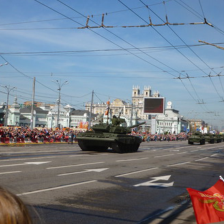}&
\includegraphics[width=0.1\linewidth, clip]{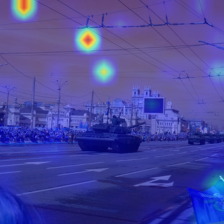}&
\includegraphics[width=0.1\linewidth, clip]{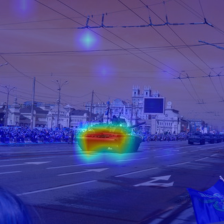}&
\includegraphics[width=0.1\linewidth, clip]{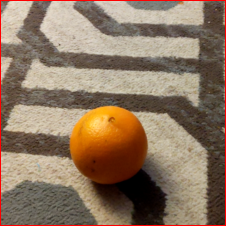}&
\includegraphics[width=0.1\linewidth, clip]{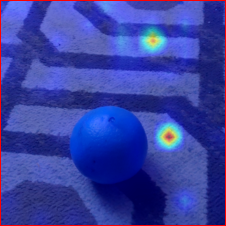}&
\includegraphics[width=0.1\linewidth, clip]{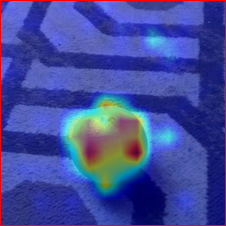}&
\includegraphics[width=0.1\linewidth, clip]{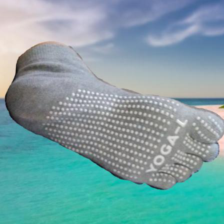}&
\includegraphics[width=0.1\linewidth, clip]{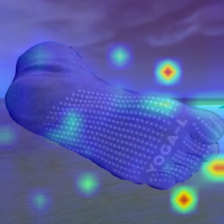}&
\includegraphics[width=0.1\linewidth, clip]{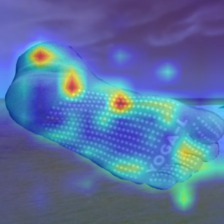}
\\
\multirow{2}{*}{\begin{turn}{90} Pred \end{turn}}
&
&{\small{Tram}} & {\small{Tank}} & & {\small{Maze}} & {\small{Orange}} &  & {\small{Swimmi-}} & {\small{Sock}}\\
& & & & &  {\small{}} & {\small{}} & & {\small{ng trunks}} \\
{\begin{turn}{90}~~~ AR-L \end{turn}} & 
\includegraphics[width=0.1\linewidth, clip]{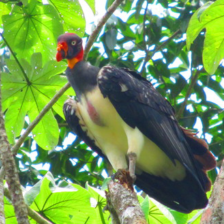}&
\includegraphics[width=0.1\linewidth, clip]{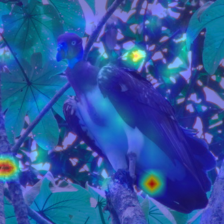}&
\includegraphics[width=0.1\linewidth, clip]{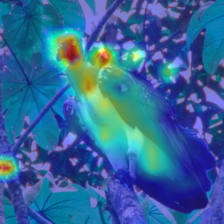}&
\includegraphics[width=0.1\linewidth, clip]{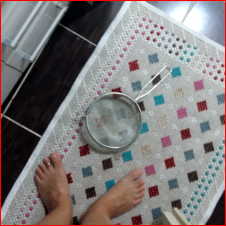}&
\includegraphics[width=0.1\linewidth, clip]{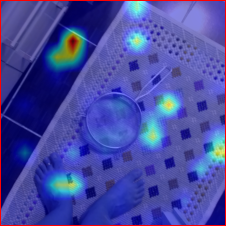}&
\includegraphics[width=0.1\linewidth, clip]{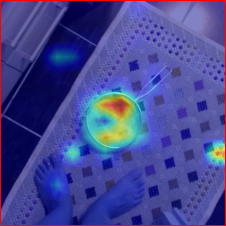}&
\includegraphics[width=0.1\linewidth, clip]{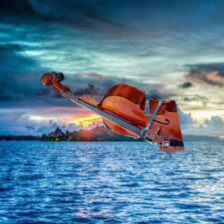}&
\includegraphics[width=0.1\linewidth, clip]{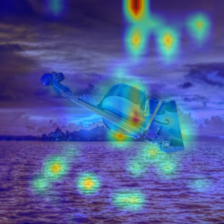}&
\includegraphics[width=0.1\linewidth, clip]{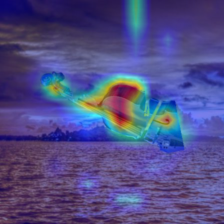}
\\
\multirow{2}{*}{\begin{turn}{90} Pred \end{turn}}
&
&{\small{Kite}} & {\small{Vulture}} & & {\small{Scale}} & {\small{Strainer}} &  & {\small{Lifeboat}} & {\small{Cello}}\\
& & & & & & & & 
\\
{\begin{turn}{90}~ DeiT-B \end{turn}} & 
\includegraphics[width=0.1\linewidth, clip]{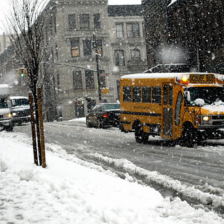}&
\includegraphics[width=0.1\linewidth, clip]{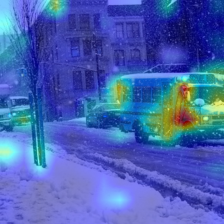}&
\includegraphics[width=0.1\linewidth, clip]{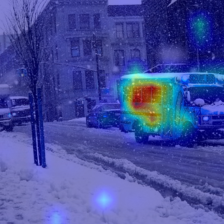}&
\includegraphics[width=0.1\linewidth, clip]{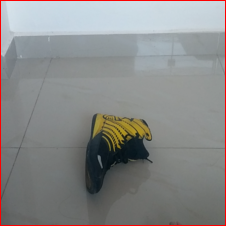}&
\includegraphics[width=0.1\linewidth, clip]{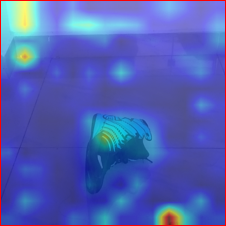}&
\includegraphics[width=0.1\linewidth, clip]{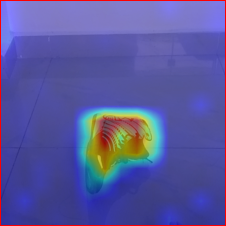}&
\includegraphics[width=0.1\linewidth, clip]{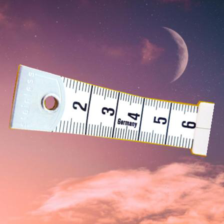}&
\includegraphics[width=0.1\linewidth, clip]{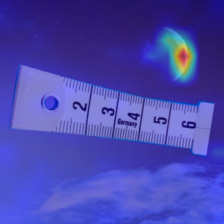}&
\includegraphics[width=0.1\linewidth, clip]{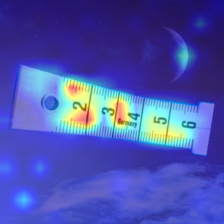}
\\

\multirow{2}{*}{\begin{turn}{90} Pred \end{turn}}
&
&{\small{Snowplow}} & {\small{School bus}} & & {\small{Bathtub}} & {\small{Running-}} &  & {\small{Balloon}} & {\small{Ruler}}\\
& & & & & & {\small{shoe}} & & 
\end{tabular}
}
\caption{Examples of cases where our method corrects wrong predictions, alongside the original and modified (after finetuning) explainability maps. The ``Pred" row specifies the predictions before and after our finetuning. The original classifiers focus on sparse or irrelevant data (e.g. the presence of snow leads DeiT-B to predict that a bus is a Snowplow, a porcupine is classified by ViT-L as a sea lion due to the presence of the ocean, a tank is classified by AR-B as a tram due to the presence of tram cables in the image, etc.). The examples are presented for the base, large models of ViT~\cite{dosovitskiy2020image}, ViT AugReg~\cite{Steiner2021HowTT} (AR), and DeiT~\cite{touvron2020training}.
\label{fig:robustness}}
\end{figure*}

The main hypothesis of this work is that improving the salient maps of ViTs trained on ImageNet will result in reduced overfitting, and better generalization to data from unseen distributions. We present a wide range of tests to confirm our hypothesis. 

First, we evaluate the improvement in robustness, i.e., the ability to maintain high accuracy under distribution shifts. 
The datasets with shifted distributions are only used for evaluation and contain both real-world datasets and synthetic ones. 
Second, we conduct segmentation tests following~\cite{chefer2020transformer} to assess the effect of our method on the level of agreement between the relevancy maps and the foreground segmentation maps.
Third, following~\cite{yosinski2014transferable}, our method employs samples from a subset of the classes during training, so that we can check whether the training classes (set A of the labels) differ from the set of classes not used during training (set B). Ideally, the method would have a positive effect on test samples from both sets A and B. 
Furthermore, in Appendix~\ref{sec:sensitivity} we evaluate how sensitive the method is to the number of samples per class, and to the number of classes in set A.

\noindent{\bf Baseline Methods\quad}
We focus on methods that resemble ours, i.e. methods that strive to correct overfitting by manipulating the saliency maps of the model. Our baselines include GradMask~\cite{Simpson2019GradMaskRO}, and Right for the Right Reasons~\cite{Ross2017RightFT} (RRR). Both GradMask and RRR were originally applied during training; in this work, however, we focus on large vision models that require significant resources for training, which makes this approach computationally impossible for us. Therefore, we apply their accuracy and relevancy losses within a finetuning process, similar to ours, in order to test their success in a fair way, while adhering to  computational limitations.

Both GradMask and RRR employ two loss functions. First, a classic cross-entropy loss with the ground-truth labels to ensure correct labeling, and second, a loss limiting the values of the gradients of irrelevant parts of the input. The latter resembles our background loss (Eq.~\ref{eq:background}), with gradients as the relevance map. We refer the reader to Appendix~\ref{sec:baselines} for the full description of the applied losses.

We note that while using the gradient of the output w.r.t. the input is common practice for interpreting CNNs, these gradients are less stable for transformer-based models. For example, results presented in~\cite{Liu2022RethinkingAE} demonstrate that for transformer-based models the classic Input$\times$Gradient method violates faithfulness. We found it difficult to grid-search hyperparameters to fit both accuracy and relevancy losses simultaneously for the baselines. Furthermore, we had to tune the hyperparameters for each model separately to obtain an improved relevance loss. Our method, on the other hand, uses the same hyperparameter choice (see Sec.~\ref{sec:method}) for all models, which makes it far more stable to use, thus allowing us to run experiments on large models as well (i.e. ViT-L, ViT AugReg-L). We refer the reader to Appendix~\ref{sec:hyperparameters} for the full description of hyperparameters used in ours experiments. 

\noindent{\bf Models and Training\quad} To demonstrate the effectiveness of our method, we experiment on three types of ViT-based models: vanilla ViT~\cite{dosovitskiy2020image}, ViT AugReg (AR)~\cite{Steiner2021HowTT}, and DeiT~\cite{Touvron2021TrainingDI} which presents techniques for efficiently training ViTs. For each model type, we experiment with different model sizes, in order to learn whether our method improves robustness across training techniques and model sizes. 
All models use $224$ resolution images with a patch size of $16 \times 16$. We use the implementation and pre-trained weights from~\cite{rw2019timm}. The small, base models are finetuned on a single RTX 2080 Ti GPU, and the large models on a single Tesla V100 GPU. 
All models are finetuned as described in Sec.~\ref{sec:method} for $50$ epochs, with a batch size of $8$. We use $3$ training images from $500$ ImageNet classes for our finetuning (overall- $1500$ samples), and another $414$ images as a validation set. In most experiments, we simply select the first $500$ classes from ImageNet-S to be used for training (set A). For results with multiple random seeds and multiple choices of the $500$ classes to use for training, see Appendix~\ref{sec:random_seeds}. The learning rate of each model is determined using a grid search between the values $5e-7$ and $5e-6$. As a rule of thumb, we select the highest learning rate for which the validation accuracy does not decrease by more than $2-3\%$ and $\mathcal{L_{\text{classification}}}$ does not increase. We find that our method is not sensitive to small shifts in the learning rate, thus this rule applies well to all models.

As mentioned in Sec.~\ref{sec:method}, we use segmentation maps to distinguish between the foreground and the background of an input image. Our experiments employ two options of obtaining segmentation maps for ImageNet training images. In the first option, we use human-annotated segmentation maps from the ImageNet-S dataset~\cite{gao2021luss}. The ImageNet-S dataset contains $10$ training samples with their segmentation maps for $919$ of the ImageNet classes. We employ a considerably smaller subset, as detailed above.
The second option does not employ extra supervision in the form of manually labeled images; instead, it employs the Tokencut object localization method presented in~\cite{wang2022tokencut} to produce foreground segmentation maps.

\begin{table}[t!]
\caption{Robustness evaluation for ViT~\cite{dosovitskiy2020image}, ViT AugReg~\cite{Steiner2021HowTT} (AR), and DeiT~\cite{Touvron2021TrainingDI} with our method, and the baseline methods GradMask~\cite{Simpson2019GradMaskRO} and Right for the Right Reason (RRR)~\cite{Ross2017RightFT}. ``Annotated segmentation" indicates whether we used annotated segmentation~\cite{gao2021luss} or unsupervised localization~\cite{wang2022tokencut}. ``Original" stands for the model without finetuning. The bottom rows indicate the average change caused by our method across all architectures (on some models the baselines could not be run successfully; therefore, we do not compute their average change). 
    }
    \begin{center}
    \resizebox{\linewidth}{!}{
    \begin{tabular}{@{}l@{~~~}c@{~~}c@{~~}c@{~}c@{~~}c@{~}c@{~~}c@{~}c@{~~}c@{~}c@{~~}c@{~}c@{~~}c@{~}c@{}}
        \toprule
        \multirow{2}{*}{Model} & \multirow{2}{*}{Method} & Annotated&\multicolumn{2}{c}{INet val} & \multicolumn{2}{c}{INet-A} & \multicolumn{2}{c}{INet-R}& \multicolumn{2}{c}{Sketch} 
        & \multicolumn{2}{c}{INet-v2}
        & \multicolumn{2}{c}{ObjNet} 
        \\
        {} & & segmentation& \begin{small}R@1\end{small}  & \begin{small}R@5\end{small} &\begin{small}R@1\end{small}  & \begin{small}R@5\end{small} &\begin{small}R@1\end{small}  & \begin{small}R@5\end{small} &\begin{small}R@1\end{small}  & \begin{small}R@5\end{small} &\begin{small}R@1\end{small}  & \begin{small}R@5\end{small} &\begin{small}R@1\end{small}  & \begin{small}R@5\end{small}   \\
        \midrule
        \multirow{5}{*}{\small{ViT-B}}  &\small{Original} & \xmark& {81.5} & {96.0}  & 16.0 & 37.0 & 33.8 & 48.5 & 35.4 & 57.4 & {71.1} & {89.9} & 35.1 & 56.4 \\
        &\small{GradMask} & \cmark & {81.8} & {96.1}   & 17.5 & 39.8 & 34.5 & 49.4 & 35.8 & 57.8 & \textbf{71.4} & \textbf{90.5} & 36.7 & 58.2  \\
        &\small{RRR} & \cmark & \textbf{81.9} & \textbf{96.2} 
        & 18.9 & 41.9 & 34.8 & 49.7 & 35.8 & 57.8 & \textbf{71.4} & \textbf{90.5} & 38.1 & 60.0\\
        &\small{Ours} &\cmark & 80.3 & 95.4 & \textbf{24.1} & \textbf{48.0} & \textbf{36.3} & \textbf{51.4} & \textbf{36.2}& \textbf{58.5} & 70.0& 89.4 & \textbf{42.2} & \textbf{65.1} \\
        &\small{Ours} &\xmark & 80.4 & 95.4 & 23.0 & 45.7 & 35.4 & 50.0 & 35.8& 58.2 & 69.8 & 89.4 &  40.8 & 64.0 \\
        \midrule
         &\small{Original} & \xmark& \textbf{82.9} &  \textbf{96.4} & 19.0 & 41.5 &  36.6 & 52.0 & 40.4 & 63.4 & {71.8}  &  {90.7} & 37.4 & 59.5 \\
        \small{ViT-L} &\small{Ours} &\cmark& 82.0 & 96.2 & \textbf{25.2} & 49.6 & 38.8 & 54.6 & {41.2}& {64.3} & 71.3 & 90.6 & {42.5} & {65.4}   \\
        &\small{Ours} &\xmark& 82.7 & \textbf{96.4} & \textbf{25.2} & \textbf{50.0} & \textbf{39.8} & \textbf{55.1} & \textbf{41.8} & \textbf{64.8} & \textbf{72.1} & \textbf{91.2} & \textbf{43.2} & \textbf{65.8} \\
        \midrule
         \multirow{5}{*}{\small{AR-S}}&\small{Original} & \xmark& {81.4} & \textbf{96.1}  & 13.0 & 33.9 & 31.2 & 47.1 & 32.8 & 54.2 &  {69.9}& {90.1} & 34.3 & 55.8  \\
         &\small{GradMask} & \cmark  & {81.3} & \textbf{96.1}  & 16.4 & 39.2 & 32.3 & 48.3 & 32.5 & 53.7 &  {70.1}& \textbf{90.3} & 37.6 & 60.2  \\
         &\small{RRR} & \cmark  & \textbf{81.5} & \textbf{96.1}  & 13.7 & 35.1 & 31.6 & 47.4 & 32.9 & 54.2 &  \textbf{70.3}& {90.1} & 35.1 & 56.7  \\
        & \small{Ours} &\cmark & 79.8 & 95.7 & 18.2 & 40.6 & \textbf{33.9} & \textbf{50.2} & 33.5& 55.4 & 69.6 & 90.0 & 38.7 & 61.1 \\
        &\small{Ours} &\xmark & 80.3 & 95.8 & \textbf{19.1} & \textbf{42.2} & 33.8 & 49.7 & \textbf{33.8} & \textbf{55.5} & 69.6 & {90.1} & \textbf{39.3} & \textbf{61.7}\\
        \midrule
        \multirow{5}{*}{\small{AR-B}}&\small{Original} & \xmark& {84.4} & {97.2} & 23.9 & 49.2 & 41.0 & 57.8 & 43.1 & 65.7 &  {73.8}& 92.3 & 41.4 & 63.7 \\
        &GradMask &\cmark  & {84.5} & \textbf{97.3} & 25.1 & 51.4 & 41.5 & 58.1 & 43.1 & {65.7} & {74.0} & \textbf{92.6} & 42.7 & 64.8 \\
        &\small{RRR} & \cmark & \textbf{84.6} & \textbf{97.3} 
        & 26.8 & 53.0 & 41.9 & 58.5 & 43.2 & 65.7 & \textbf{74.3} & \textbf{92.6} & 43.7 & 65.9\\
        & \small{Ours} &\cmark & 83.1 & 96.9 & \textbf{31.3} & {57.1} & \textbf{44.7} & \textbf{61.5} & 44.6& \textbf{67.4} & 73.5 & 92.0 & \textbf{47.1} & \textbf{70.0} \\
        &\small{Ours} &\xmark & 83.6 & 97.1 & 31.2 & \textbf{57.2} & 44.5 & 60.9 & \textbf{44.7} & \textbf{67.4} & 73.7 & {92.4} & 46.5 & 69.1 \\
        \midrule
        &\small{Original} & \xmark&
        \textbf{85.6} &  \textbf{97.8} & 34.7 & 61.0 & 48.8 & 64.9 & 51.8 & 73.6 & {75.8} & {93.4} & 46.5 & 68.3  \\
        \small{AR-L}& \small{Ours} &\cmark & 85.1 & 97.5 & 42.1 & 67.5 & \textbf{54.0} & \textbf{69.1} & \textbf{54.2}& \textbf{75.8} & {75.8}& {93.4} & 51.6 & 73.2  \\
        &\small{Ours} &\xmark & 85.4 & 97.6 & \textbf{42.4} & \textbf{68.0} & 53.8 & 69.0 & 54.1 & \textbf{75.8} & \textbf{76.1} & \textbf{93.6} & \textbf{52.0} & \textbf{73.5} \\
        \midrule
         \multirow{4}{*}{\small{DeiT-S}}&\small{Original} & \xmark& 78.1 & 93.7  & 8.3 & 23.5 & 28.2 & 41.9 & 28.8 & 46.7 & 66.5 & 86.6 & 28.3 & 47.3  \\
         &\small{GradMask} & \cmark & 77.0 & 93.6  & 7.9 & 24.7 & 26.6 & 40.5 & 26.0 & 43.5 & 64.5 & 85.6 & 28.2 & 48.6  \\
          &\small{RRR} & \cmark & 78.1 & 94.1  & 9.0 & 26.9 & 26.9 & 40.6 & 26.9 & 44.4 & 66.0 & 86.7 & 29.3 & 49.9  \\
        & \small{Ours} &\cmark & \textbf{78.6} & \textbf{94.5} & 10.1 & 29.0 & 29.3 & 43.6 & 29.1 & 47.8 & \textbf{67.3}& 87.3 & \textbf{31.6} & \textbf{53.0} \\
        &\small{Ours} &\xmark & \textbf{78.6} & 94.4 & \textbf{11.0} & \textbf{30.3} & \textbf{29.9} & \textbf{44.4} & \textbf{29.4} & \textbf{48.0} & 67.1 & \textbf{87.4} & \textbf{31.6} & 52.9\\
        \midrule
        \multirow{5}{*}{\small{DeiT-B}}&\small{Original} & \xmark& {80.8} & 94.2 & 12.9 & 31.0 & 30.9 & 44.2 & 31.2 & 48.6 & \textbf{69.7} & 86.8 & 31.4 & 48.5 \\
        &\small{GradMask} & \textbf{\cmark} & \textbf{81.1} & \textbf{95.3} & 15.1 & 36.9 & 31.0 & 45.5 & 31.2 & 49.1 & \textbf{69.7} & \textbf{88.7} & 33.5 & 53.1 \\
         &\small{RRR} & \cmark & {81.0} & {95.2}   & 14.8 & 37.0 & 30.7 & 45.1 & 30.9 & 48.8 & {69.5} & 88.6 & 33.6 & 53.3 \\
        & \small{Ours} &\cmark & 80.5 & 94.9 & 17.2 & 40.0 & 32.4 & 47.0 & 30.9& 49.2 & 69.1& 88.3 & 35.9 & 56.2  \\
        &\small{Ours} &\xmark & 80.5 & {95.0} & \textbf{18.3} & \textbf{40.9} & \textbf{32.8} & \textbf{47.5} & \textbf{31.5} & \textbf{49.9} & 69.3 & {88.5} & \textbf{36.3} & \textbf{56.6} \\
        \midrule
        \textbf{Avg.} & \small{Ours} &\cmark & {\color{red} -0.8} & 
        0.0
        & \dellarge{+5.8} & \dellarge{+7.8} & \dellarge{+2.7} & \dellarge{+3.0} & \dellarge{+0.9} & \dellarge{+1.3} & {\color{red} -0.3} & \dellarge{+0.2}& \dellarge{+5.0}& \dellarge{+6.4}\\
        \textbf{change} & \small{Ours} & \xmark & {\color{red} -0.5} & 0.0 & \dellarge{+6.1} & \dellarge{+8.2} & \dellarge{+2.8} & \dellarge{+2.9} & \dellarge{+1.1} & \dellarge{+1.4} &{\color{red} -0.1}& \dellarge{+0.4} & \dellarge{+5.0}& \dellarge{+6.3} \\
        \bottomrule
    \end{tabular}
    }
    \smallskip
    \smallskip
    \label{table:robustness}
    \end{center}
    \vspace{-24px}
\end{table}

\begin{table}[t!]
\caption{Robustness evaluation on the synthetic SI-Score dataset~\cite{Djolonga2021OnRA}, which tests changes in object position, rotation, and size using our method and the baseline methods GradMask~\cite{Simpson2019GradMaskRO}, Right for the Right Reasons (RRR)~\cite{Ross2017RightFT}. The models tested are ViT~\cite{dosovitskiy2020image}, ViT AugReg~\cite{Steiner2021HowTT} (AR), and DeiT~\cite{Touvron2021TrainingDI}. ``Annotated segmentation" indicates if we used annotated segmentation~\cite{gao2021luss} or unsupervised localization~\cite{wang2022tokencut}. ``Original" stands for the model without finetuning.}
    \centering
    \resizebox{\linewidth}{!}{
    \begin{tabular}{@{~}l@{~~~}c@{~~}c@{~~~~~}c@{~~~}c@{~~~}c@{~~~}c@{~~~}c@{~~}c@{}}
        \toprule
        \multirow{2}{*}{Model} & \multirow{2}{*}{Method} & Annotated&\multicolumn{2}{c}{SI-location} & \multicolumn{2}{c}{SI-rotation} & \multicolumn{2}{c}{SI-size}
        \\
        {} & & segmentation & \begin{small}R@1\end{small}  & \begin{small}R@5\end{small} &\begin{small}R@1\end{small}  & \begin{small}R@5\end{small} &\begin{small}R@1\end{small}  & \begin{small}R@5\end{small}  \\
        \midrule
        \multirow{5}{*}{ViT-B} &Original & \xmark & 33.3 & 52.2 & 39.1 & 58.3 & 55.6 & 76.2  \\
         & GradMask &\cmark & {34.6 \del{(+1.3)}} & {53.9 \del{(+1.7)}} & {40.7 \del{(+1.6)}} & {60.3 \del{(+2.0)}}& {57.0 \del{(+1.4)}} & {77.5 \del{(+1.3)}} \\
        & RRR &\cmark & {35.6 \del{(+2.3)}} & {55.0 \del{(+2.8)}} & {41.9 \del{(+2.8)}} & {61.8 \del{(+3.5)}}& {58.0 \del{(+2.4)}} & {78.4 \del{(+2.2)}} \\
        & Ours &\cmark & \textbf{38.6 \del{(+5.3)}} & \textbf{57.8 \del{(+5.6)}} & \textbf{46.2 \del{(+7.1)}} & \textbf{67.0 \del{(+8.7)}}& \textbf{61.0 \del{(+5.4)}} & \textbf{81.4 \del{(+5.2)}} \\
        & Ours & \xmark & 38.4 \del{(+5.1)} & 57.0 \del{(+4.8)} & 44.8 \del{(+5.7)}  & 65.2 \del{(+6.9)}  & 60.2 \del{(+4.6)}  & 80.6 \del{(+4.4)} \\
        \midrule
        &Original & \xmark & 31.6 & 50.3  & 40.7 & 60.1 & 54.8 & 75.6  \\
        ViT-L & Ours &\cmark& 36.3 \del{(+4.7)} & 56.2 \del{(+5.9)} & \textbf{45.3 \del{(+4.6)}} & 66.2 \del{(+6.1)}& 58.6 \del{(+3.8)} & 80.3 \del{(+4.7)}  \\
        & Ours & \xmark & \textbf{36.7 \del{(+5.1)}} & \textbf{56.3 \del{(+6.0)}} & \textbf{45.3 \del{(+4.6)}} & \textbf{66.6 \del{(+6.5)}} & \textbf{59.1 \del{(+4.3)}} & \textbf{80.5 \del{(+4.9)}}  \\
        \midrule
        \multirow{5}{*}{AR-S}&Original & \xmark & 32.4 & 51.7  & 40.6 & 59.6 & 55.4 & 75.7 \\
        & GradMask & \cmark & 34.3 \del{(+1.9)} & 53.9 \del{(+2.2)} & 43.3 \del{(+2.7)} & 63.0 \del{(+3.4)} & 58.0 \del{(+2.6)} & 78.3 \del{(+2.6)} \\
        & RRR & \cmark & 32.9 \del{(+0.5)} & 52.3 \del{(+0.6)} & 41.4 \del{(+0.8)} & 60.6 \del{(+1.0)} & 56.0 \del{(+0.6)} & 76.3 \del{(+0.6)} \\
        & Ours &\cmark & \textbf{36.8 \del{(+4.4)}} & \textbf{56.6 \del{(+4.9)}} & \textbf{47.6 \del{(+7.0)}} & \textbf{67.8 \del{(+8.2)}}& \textbf{61.3 \del{(+5.9)}} & \textbf{81.2 \del{(+5.5)}} \\
        & Ours & \xmark & 36.3 \del{(+3.9)} & 55.6 \del{(+3.9)} & 46.6 \del{(+6.0)} & 66.7 \del{(+7.1)} & 60.7 \del{(+5.3)} & 80.4 \del{(+4.7)} \\
        \midrule
        \multirow{5}{*}{AR-B}&Original & \xmark & 40.5 &  60.8 & 48.1 & 68.3 & 60.6 & 80.4   \\
        & GradMask &\cmark & 41.5 \del{(+1.0)}
        & 61.8 \del{(+1.0)} & 49.3 \del{(+1.2)} & 69.5 \del{(+1.2)}& 61.4 \del{(+0.8)} & 81.3 \del{(+0.9)} \\
        & RRR &\cmark & 42.4 \del{(+1.9)}
        & 62.7 \del{(+1.9)} & 50.4 \del{(+2.3)} & 70.7 \del{(+2.4)}& 62.1 \del{(+1.5)} & 82.0 \del{(+1.6)} \\
        & Ours &\cmark & 43.2 \del{(+2.7)}
        & 62.8 \del{(+2.0)} & 54.0 \del{(+5.9)} & 74.6 \del{(+6.3)}& 64.1 \del{(+3.5)} & 83.9 \del{(+3.5)} \\
        & Ours & \xmark & \textbf{44.3 \del{(+3.8)}} & \textbf{64.0 \del{(+3.2)}} & \textbf{54.6 \del{(+6.5)}} & \textbf{74.7 \del{(+6.4)}} & \textbf{64.5 \del{(+3.9)}} & \textbf{84.6 \del{(+4.2)}}  \\
        \midrule
        &Original & \xmark & 43.8 & 64.2  & 52.4 & 72.5 & 62.3 &  82.2  \\
        AR-L& Ours & \cmark & \textbf{48.3 \del{(+4.5)}} & \textbf{68.5 \del{(+4.3)}} & 57.0 \del{(+4.6)} & {77.2 \del{(+4.7)}}& {66.4 \del{(+4.1)}} & \textbf{86.0 \del{(+3.8)}} \\
        & Ours & \xmark & 47.4 \del{(+3.6)} &67.4 \del{(+3.2)} & \textbf{58.0 \del{(+5.6)}} & \textbf{78.1 \del{(+5.6)}} & \textbf{66.5 \del{(+4.2)}} & 85.6 \del{(+3.4)}\\
           \midrule
        \multirow{5}{*}{DeiT-S}&Original & \xmark & 30.7 &  50.4 & 36.7 & 54.3 & 51.6 & 72.0  \\
        & GradMask &\cmark & 32.0 \del{(+1.3)}
        & 50.7 \del{(+0.3)} & 38.9 \del{(+2.2)} & 56.7 \del{(+2.4)}& 54.1 \del{(+2.5)} & 74.0 \del{(+2.0)} \\
        & RRR &\cmark & 32.0 \del{(+1.3)}
        & 51.0 \del{(+0.6)} & 38.5 \del{(+1.8)} & 56.3 \del{(+2.0)}& 53.9 \del{(+2.3)} & 73.8 \del{(+1.8)} \\
        {}& Ours & \cmark & 32.3 \del{(+1.6)} & \textbf{51.5 \del{(+1.1)}} & 40.6 \del{(+3.9)} & 59.4 \del{(+5.1)} & 55.8 \del{(+4.2)} & \textbf{76.3 \del{(+4.3)}} \\
        & Ours & \xmark & \textbf{32.5 \del{(+1.8)}} & 51.4 \del{(+1.0)} & \textbf{41.0 \del{(+4.3)}} & \textbf{59.6 \del{(+5.3)}} & \textbf{56.0 \del{(+4.4)}} & 76.1 \del{(+4.1)}      \\
        \midrule
        \multirow{5}{*}{DeiT-B}&Original & \xmark & 34.5 & 54.6 & 39.3 & 56.3 & 54.6 & 73.4  \\
        {}& GradMask & \cmark & 34.1 {\color{red} (-0.4)} & 54.9 \del{(+0.3)} & 39.1 {\color{red} (-0.2)} & 58.3 \del{(+2.0)}& 55.2 \del{(+0.6)} & 75.8 \del{(+2.4)}  \\
         {}& RRR & \cmark & 34.4 {\color{red} (-0.1)} & 55.2 \del{(+0.6)} & 40.4 \del{(+1.1)} & 58.5 \del{(+2.2)}& 55.3 \del{(+0.7)} & 75.8 \del{(+2.4)}  \\
        {}& Ours & \cmark & 36.6 \del{(+2.1)} & 57.0 \del{(+2.4)} & 42.9 \del{(+3.6)} & 61.5 \del{(+5.2)}& 58.0 \del{(+3.4)} & 78.2 \del{(+4.8)}  \\
        & Ours& \xmark & \textbf{ 37.8 \del{(+3.3)}} & \textbf{58.1 \del{(+3.5)}} & \textbf{ 44.2 \del{(+4.9)}} & \textbf{62.7 \del{(+6.4)}} & \textbf{59.3 \del{(+4.7)}} & \textbf{79.0 \del{(+5.6)}}\\
        \bottomrule
    \end{tabular}
    }
    \smallskip
    \smallskip
    \label{table:SI}
    \vspace{-22px}
\end{table}

\noindent{\bf Results\quad} Tab.~\ref{table:robustness} presents the results of our method and the baseline methods applied to different ViT-based models. As can be seen, for real-world datasets, both adversarial datasets (INet-A) and datasets with random and controlled background, rotations, and view points (ObjectNet), our method significantly and consistently improves performance (average of $5.8\%, 5.0\%$ top-1 improvement on INet-A, ObjectNet, respectively). {For datasets that contain art, sculptures, sketches etc. (INet-R, INet-Sketch) the increase in accuracy is less steep ($2.7\%, 0.9\%$ averaged top-1 improvement, respectively). This can be intuitively attributed to the fact that art and sketches often feature the object without a background, or with a uniform background.} Additionally, as can be seen, although the baseline methods preserve accuracy on the datasets from the original ImageNet distribution (INet val set and INet-v2), they fall behind our method on real-world out-of-distribution datasets (INet-A and ObjectNet), indicating that the baselines are less successful in alleviating overfitting.

Following our method, there is a slight decrease in the performance of the models on data from the ImageNet distribution (INet val set and INet-v2), which can be attributed to the fact that we reduce some of the overfitting on the ImageNet data.  Fig.~\ref{fig:improved_expl} presents examples of all three possible prediction cases on the ImageNet validation set: cases where our method preserves the original classification, corrects the original prediction, and changes a correct prediction. As can be seen, while performance sometimes decreases with respect to the assigned label, in most cases the rationale of the class proposed by the modified model is clear, often indicating a more natural labeling option. In addition, in the cases where the prediction remains the same or is corrected, our method produces improved relevance maps.

Additionally, as shown in Tab.~\ref{table:SI} for the SI-Score dataset, which is a synthetic dataset designed for testing resilience for shifts in object locations, sizes, and rotations, there is a very steep improvement in performance across all models, while, once again, the baselines fall behind on all models and sizes.

Evidently, in both Tab.~\ref{table:robustness},\ref{table:SI} our method works just as well and often better when using the unsupervised segmentation maps. This means that our method may be applied without requiring any manual supervision, except for the image label.

Fig.~\ref{fig:robustness} presents example cases in which our method is able to correct the prediction of the original model on images from various robustness datasets. As can be seen, the original models tend to overinterpret the background, and therefore produce false classifications based on it. For example, a lemon is classified as a golf ball due to the grass in the background (third example in the first row), a tank is classified as a tram due to the tram cables at the top of the image (first example in the third row), and so on. Additional examples can be found in Appendix~\ref{sec:qualitative}.

\noindent{\bf Segmentation Tests\quad} Since our motivation is to encourage the relevance to focus less on the background and more on as much of the foreground as possible, we test the resemblance of the resulting relevance maps to the segmentation maps following~\cite{chefer2020transformer}. As can be seen in Tab.~\ref{table:segmentation}, our method significantly and consistently improves segmentation metrics on all models, indicating that our finetuning indeed achieves its goal.

\begin{table}[t!]
\caption{Evaluation of segmentation performance from relevance maps on the ImageNet-segmentation dataset~\cite{Guillaumin2014ImageNetAW} for ViT~\cite{dosovitskiy2020image}, ViT AugReg~\cite{Steiner2021HowTT} (AR), and DeiT~\cite{Touvron2021TrainingDI} before and after finetuning with our method. Metrics and dataset are taken from~\cite{chefer2020transformer}.
    }
    \begin{center}
    \begin{tabular}{l@{~~~}c@{~}c@{~~~}c@{~}c@{~~~}c@{~}c@{~~~}c@{~}c@{~~~}c@{~}c@{~~~}c@{~}c@{~~~}c@{~}c@{}}
        \toprule
          \multirow{2}{*}{Model} & \multicolumn{2}{c}{ViT-B}&\multicolumn{2}{c}{ViT-L} & \multicolumn{2}{c}{AR-S} & \multicolumn{2}{c}{AR-B}& \multicolumn{2}{c}{AR-L} 
        & \multicolumn{2}{c}{DeiT-S}
        & \multicolumn{2}{c}{DeiT-B} 
         \\
        & \begin{small}Orig\end{small} & \begin{small}Ours\end{small}& \begin{small}Orig\end{small}  & \begin{small}Ours\end{small} &\begin{small}Orig\end{small}  & \begin{small}Ours\end{small} &\begin{small}Orig\end{small}  & \begin{small}Ours\end{small} &\begin{small}Orig\end{small}  & \begin{small}Ours\end{small} &\begin{small}Orig\end{small}  & \begin{small}Ours\end{small} &\begin{small}Orig\end{small}  & \begin{small}Ours\end{small}  \\
        \midrule
         \small{Pixel acc.} & 76.3 & \textbf{82.1} & 73.4 & \textbf{82.5} & 76.7 & \textbf{83.3} & 76.6 & \textbf{81.2} & 65.2 & \textbf{78.9} & 78.7 & \textbf{80.8} & 79.0 & \textbf{81.3} \\
         \midrule
        \small{mIoU} & 58.3 & \textbf{65.8} & 54.4 & \textbf{66.4} & 57.7 & \textbf{67.7} & 57.1 & \textbf{64.6} & 43.6 & \textbf{61.0} & 60.7 & \textbf{64.0} & 61.6 & \textbf{64.7}\\
        \midrule
        \small{mAP} & 85.3 & \textbf{87.5} & 82.7 & \textbf{86.9} & 84.2 & \textbf{87.7} & 84.4 & \textbf{86.8} & 78.6 & \textbf{85.4} & 85.0 & \textbf{86.4} & 85.7 & \textbf{86.8}
        \\
        \bottomrule
    \end{tabular}
    \smallskip
    \smallskip
    \label{table:segmentation}
    \end{center}
    \vspace{-18px}
\end{table}

\begin{figure*}[t!]%
    \centering
    \begin{tabular}{c@{~~~~~~~~~~}c@{~~}c@{}}
    {{\includegraphics[width=6cm]{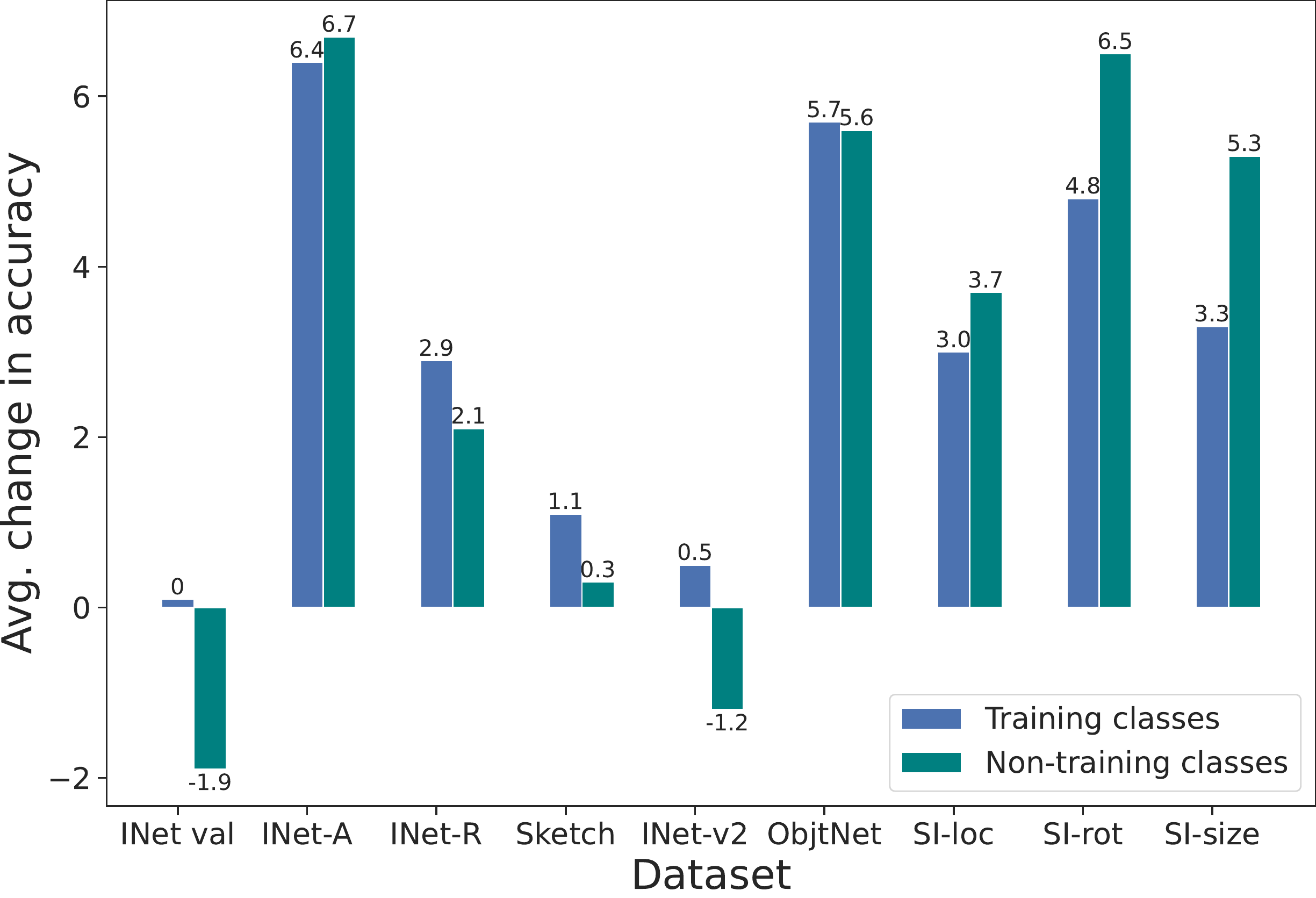}}}&
    {{\includegraphics[width=6cm]{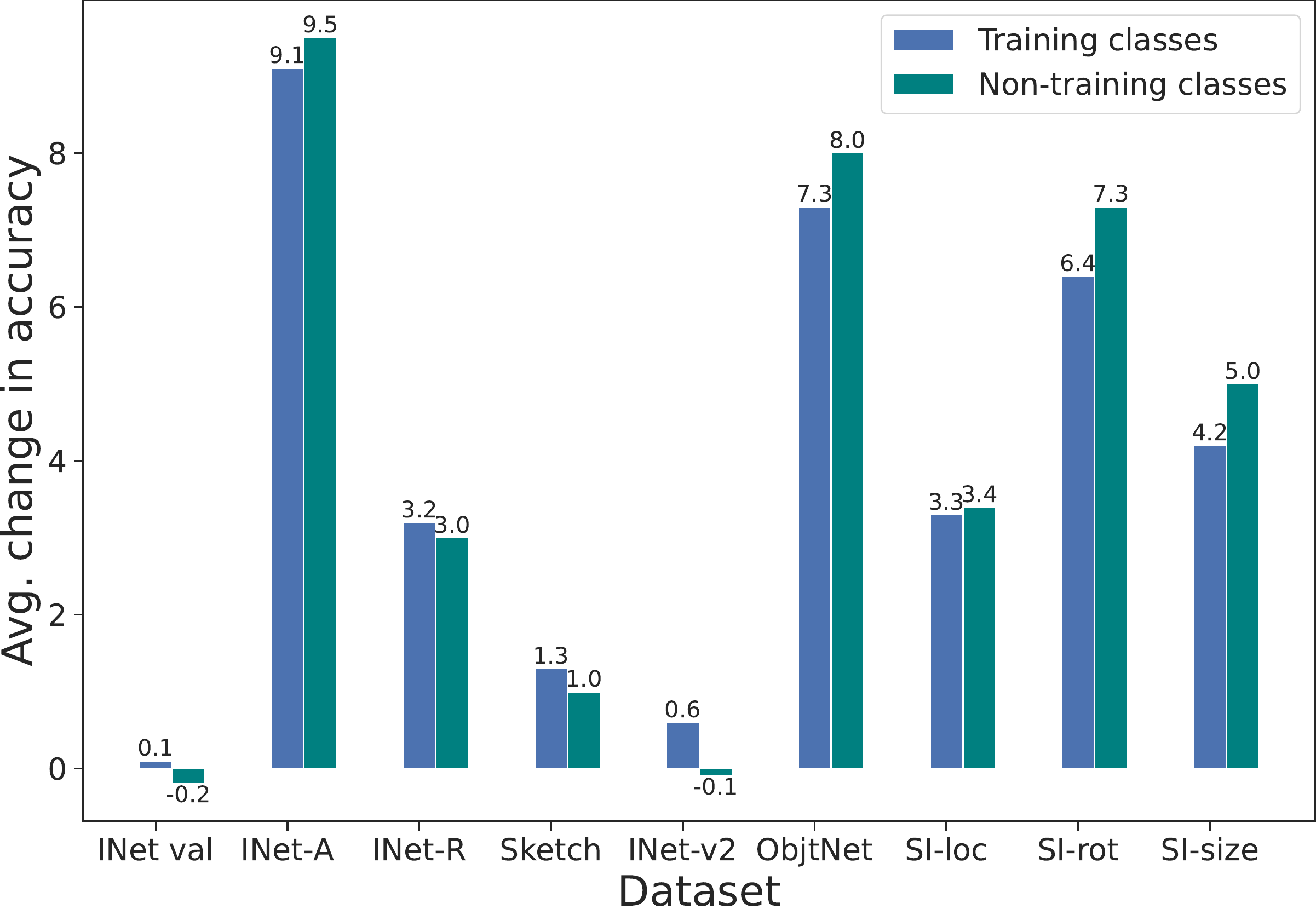}}}&
    \\
    (a) & (b)\\
    \end{tabular}
    \caption{Evaluation of the average change produced by our method on the training classes and on the classes that were not in the training set. Changes are averaged across the base models of ViT~\cite{dosovitskiy2020image}, ViT AugReg~\cite{Steiner2021HowTT}, and DeiT~\cite{touvron2020training}. (a) top-1 average change, (b) top-5 average change. 
    }%
    \label{fig:ab}%
    \vspace{-12px}
\end{figure*}

\noindent{\bf Comparing training classes to the other classes\quad} To ensure that the effects of our finetuning generalize to classes that were not included in the training set, we test the increase in robustness resulting from our method separately on the training classes and non-training classes. Fig.~\ref{fig:ab} presents the average improvement across the base models of ViT~\cite{dosovitskiy2020image}, ViT AugReg~\cite{Steiner2021HowTT}(AR), and DeiT~\cite{touvron2020training}. As can be seen, both subsets of classes demonstrate a very similar increase in accuracy for the robustness datasets, with the classes that belong to the training set demonstrating better performance on the datasets from the original ImageNet distribution (INet val, INet-v2), as can be expected due to the fact that they were represented in the training set. The full results of this experiment are presented in Appendix~\ref{sec:ab}.

\noindent{\bf Ablation Study\quad}
We conduct an ablation study to test the effect of each of our loss terms on the result of the finetuning process, by studying the impact of removing each loss term $\mathcal{L_{\text{classification}}}, \mathcal{L_{\text{bg}}}, \mathcal{L_{\text{fg}}}$. Our ablation study is conducted on the base models of ViT and DeiT, as they demonstrate different advantages for each loss term. Additionally, we perform an ablation to test the choice of confidence-boosting as the classification loss (Eq.~\ref{eq:cls-loss}), by replacing it with the classic cross-entropy loss with the ground-truth label. For brevity, Tab.~\ref{table:ablation} presents the top-1 accuracy results for each version of our method; the complementary table for top-5 accuracy can be found in Appendix~\ref{sec:ablation}. As can be seen, validation accuracy is lost mostly due to our background loss (Eq.~\ref{eq:background}), as when we remove it, the accuracy remains intact, and yet, as we hypothesized, it is more effective than the foreground loss (Eq.~\ref{eq:foreground}) in increasing the robustness, since when removing it, the accuracy on the out-of-distribution datasets drops significantly. Additionally, the DeiT ablation demonstrates that in some cases, the classification loss does not contribute to the increase in robustness (other than for INet-A), nor is it necessary for preserving the original ImageNet validation accuracy. However, for ViT, the classification loss is crucial for avoiding a significant loss of accuracy.

Finally, our ablation study demonstrates the benefit of using confidence boosting over the ground-truth for the classification loss. While the ground-truth variant preserves the original accuracy better (more so for ViT than for DeiT), using confidence boosting often improves robustness more significantly over using the ground-truth labels (see for example INet-A for ViT).

\begin{table}[t!]
\caption{Ablation study for our method on ViT~\cite{dosovitskiy2020image} and DeiT~\cite{touvron2020training} base models. The table presents top-1 accuracy. We check the impact of each term of our loss on robustness, as well as the choice of confidence boosting versus cross-entropy with the ground-truth label (labeled as w/ ground-truth).
    }
    \begin{center}
    \resizebox{\linewidth}{!}{
    \begin{tabular}{@{}l@{~~}c@{~~~~}c@{~~~~}c@{~~~~}c@{~~~~}c@{~~~~}c@{~~~~}c@{~~~~}c@{~~~~}c@{~~~~}c@{}}
        \toprule
           {Model} & {Method} &{INet val} & {INet-A} & {INet-R}& {Sketch} 
        & {INet-v2}
        & {ObjNet} & {SI-loc.} & {SI-rot.}  & {SI-size} 
        \\
        \midrule
        \multirow{6}{*}{\small{\begin{turn}{90} ViT-B \end{turn}}}  &\small{Original} & 81.5 & 16.0 & 33.8 & 35.4 & 71.1 & 35.1 & 33.3 & 39.1 & 55.6  \\
        &\small{Ours}  & 80.3 & \textbf{24.1} & \textbf{36.3} & 36.2 & 70.0 & \textbf{42.2} & 38.6 & \textbf{46.2} & \textbf{61.0} \\
        &\small{w/o $\mathcal{L_{\text{classification}}}$ (Eq.~\ref{eq:cls-loss}}) & 77.8 & 17.9 & 34.2 & 34.8 & 67.4 & 37.6 & 37.2 & 43.0 & 58.4 \\
        &\small{w/o $\mathcal{L_{\text{bg}}}$ (Eq.~\ref{eq:background})} & 81.5 & 19.1 & 35.9 & \textbf{36.4} & \textbf{71.4} & 39.7 & 34.9 & 42.3 & 58.4   \\
        &\small{w/o $\mathcal{L_{\text{fg}}}$ (Eq.~\ref{eq:foreground}}) & 80.2 & \textbf{24.1} & 34.3 & 35.2 & 69.6 & 41.8 & \textbf{39.2} & 45.6 & 60.7  \\
        &\small{w/ ground-truth} & \textbf{81.7} & 21.5 & 35.5 & 35.8 & 71.2 & 40.3 & 37.8 & 44.5 & 60.1   \\
        \midrule
        \multirow{6}{*}{\small{\begin{turn}{90} DeiT-B \end{turn}}} & \small{Original} & 80.8 & 12.9 & 30.9 & 31.2 & 69.7 & 31.4 & 34.5 & 39.3 & 54.6  \\
        & \small{Ours}  & 80.5 & 17.2 & 32.4 & 30.9 & 69.1 & 35.9 & 36.6 & 42.9 & 58.0    \\
        &\small{w/o $\mathcal{L_{\text{classification}}}$ (Eq.~\ref{eq:cls-loss}}) & \textbf{81.0} & 15.5 & \textbf{33.3} & \textbf{32.1} & \textbf{69.8} & \textbf{36.1} & \textbf{39.0} & \textbf{44.0} & \textbf{58.3}\\
        &\small{w/o $\mathcal{L_{\text{bg}}}$ (Eq.~\ref{eq:background})}  & \textbf{81.0} & 13.2 & 30.5 & 30.2 & 69.6 & 33.0 & 32.7 & 39.7 & 54.5 \\
        &\small{w/o $\mathcal{L_{\text{fg}}}$ (Eq.~\ref{eq:foreground}}) & 80.3 & \textbf{17.9} & 32.6 & 31.0 & 69.0 & 35.8 & 37.3 & 43.0 & 58.2  \\
        &\small{w/ ground-truth} & 80.7 & 17.2 & 31.9 & 31.3 & 69.2 & 35.6 & 36.7 & 42.8 & 57.6\\
        \bottomrule
    \end{tabular}
    }
    \smallskip
    \smallskip
    \label{table:ablation}
    \end{center}
    \vspace{-24px}
\end{table}

\section{Discussion and limitations}

A surge of works have explored the benefits of using large-scale datasets of unlabeled samples from the internet, and many techniques have been proposed to train vision models in an unsupervised or self-supervised manner~\cite{caron2021emerging,Goyal2022VisionMA}. As a result, it is increasingly important to develop methods for boosting model robustness without requiring any labels. We note that our method is compatible with such frameworks. As can be seen from Tab.~\ref{table:robustness},~\ref{table:SI}, our method performs well, even when applied with foreground masks obtained in an unsupervised manner, using Tokencut. Additionally, all of our losses can be applied without knowledge of the ground-truth label, since our classification loss $\mathcal{L_{\text{classification}}}$ uses the predicted label, and the relevance maps could be propagated w.r.t. the predicted class.  

Recently, it has been discovered that the effect of augmentation-based regularization is extremely uneven across classes~\cite{balestriero2022effects}. While such a regularization improves performance on average, some classes benefit significantly from it, while other classes suffer from a large drop in accuracy. 
This variation is also apparent in the effect of our method. Above we have established that the method helps both classes which are included in the training set of the finetuning and classes which are not. However, in both sets there is considerable variability in the effect of individual classes. This makes sense, since some classes present more well-localized objects and some classes require more reliance on the context of the object. See Appendix~\ref{sec:limitations} for these results.

\section{Conclusions}
Can segmentation information help image categorization? Intuitively, the answer has to be positive. However, despite some effort to manually segment images of classification datasets, the obtained improvement, if any, is soon overtaken by better image-level methods. 

Here, we propose a generic way to improve classification accuracy that can be applied to virtually any image classifier. Applied to transformers, we present evidence to show that while the accuracy on the original dataset does not improve, there is an increase in accuracy for out-of-distribution test sets.
The method optimizes the relevancy maps directly based on intuitive desiderata. It opens a new way for improving accuracy using explainability methods, which are currently seldom used for improving downstream tasks other than seeding weakly supervised segmentation methods.

\bibliographystyle{plain}
\bibliography{relevancy}

\newpage
\appendix

\section{Reinterpreting overinterpretation}
\label{sec:sis}
One way to assess the salient behavior of a model is to study Sufficient Input Subsets (SIS), i.e., the minimal number of pixels necessary for a confident prediction~\cite{carter2019made}. A gradient signal can be utilized to find the sufficient pixels~\cite{overinterpretation}. Finding an SIS for a class can imply that the classifier has overinterpreted its input since it can make a confident accurate decision using a small, sparse subset of pixels, which does not appear meaningful to humans. 

We study the SIS with gradients approach for ViT models, and find that it can be misleading. Specifically, SIS can be regarded as an adversarial method that can lead to high-confidence classification of {\em any label} from a sparse set of pixels.

\begin{figure*}[h!]
  \centering
  \resizebox{\linewidth}{!}{
\begin{tabular}{c@{~~}c@{~~}c@{~~}c@{~~}c@{~~}c@{~~}c@{~~}c}
& \multicolumn{5}{c}{SIS subset}\\
\cmidrule(r){2-7}
Original & Original class & \multirow{2}{*}{Gibbon} & Black- & Common- & \multirow{2}{*}{Shovel} & Australian \\
classification & SIS & & widow & iguana &  &terrier\\
\includegraphics[width=0.13\linewidth, clip]{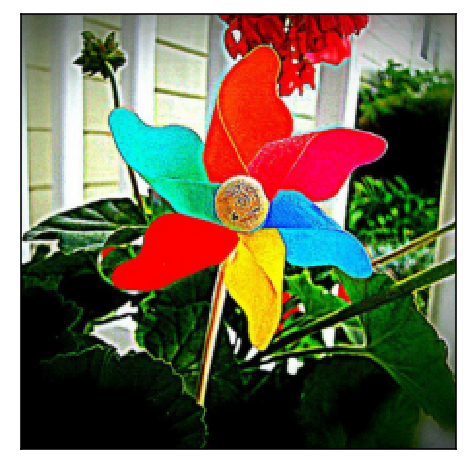}&
\includegraphics[width=0.13\linewidth, clip]{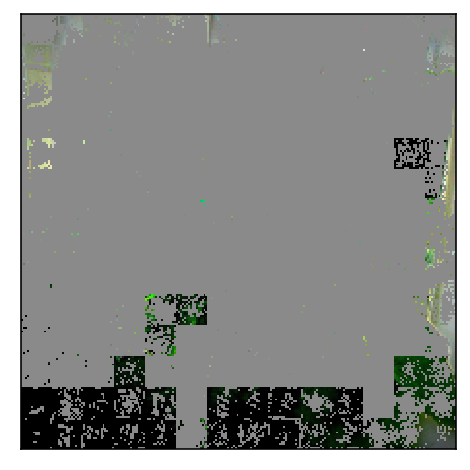}&
\includegraphics[width=0.13\linewidth, clip]{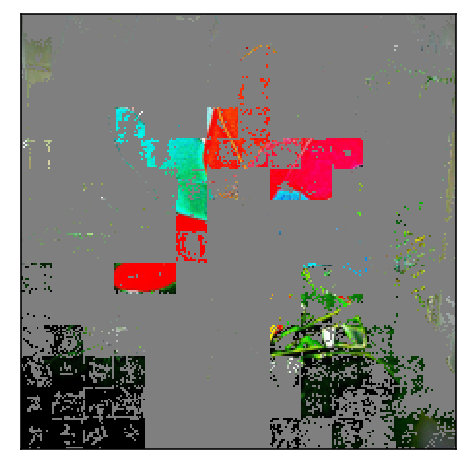}&
\includegraphics[width=0.13\linewidth, clip]{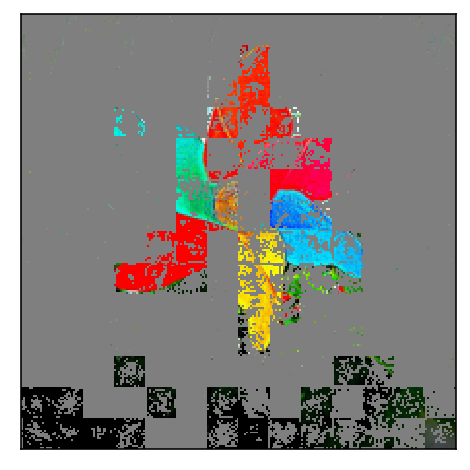}&
\includegraphics[width=0.13\linewidth, clip]{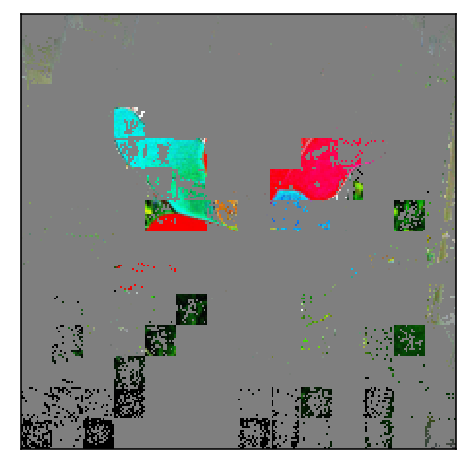}&
\includegraphics[width=0.13\linewidth, clip]{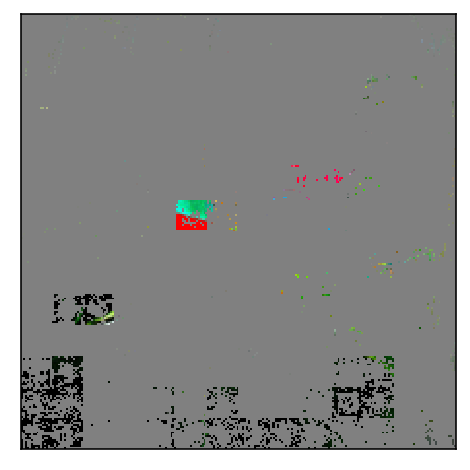}&
\includegraphics[width=0.13\linewidth, clip]{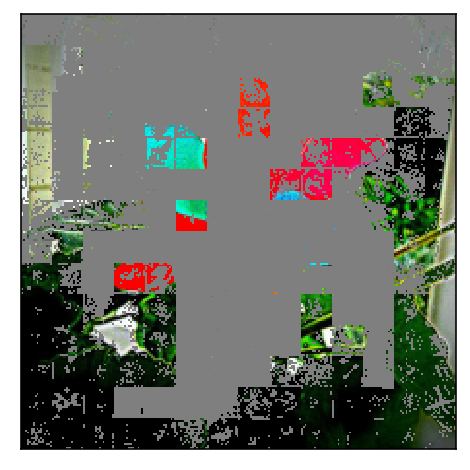}
\\
\small{Pinwheel} &
\small{Pinwheel} &
\small{Gibbon} &
\small{Black window} &
\small{Common iguana} &
\small{Shovel} &
\small{Australian terrier} &
\\
\small{$99.9\%$} &
\small{$97.9\%$} &
\small{$93.1\%$}
&
\small{$91.1\%$}
&
\small{$94.1\%$}
&
\small{$92.5\%$}
&
\small{$90.5\%$}
\\
\includegraphics[width=0.13\linewidth, clip]{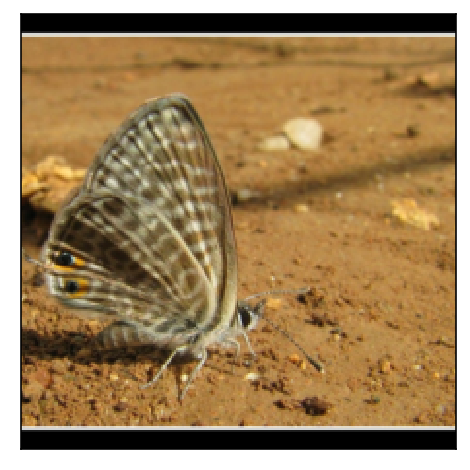}&
\includegraphics[width=0.13\linewidth, clip]{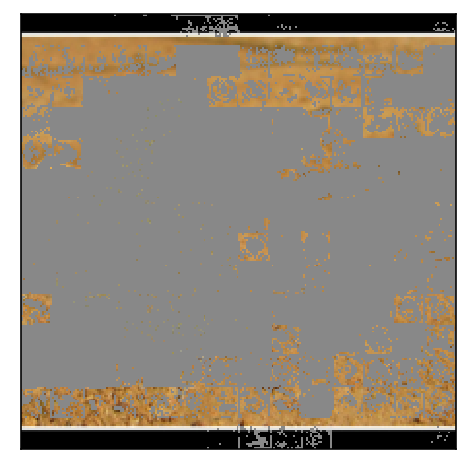}&
\includegraphics[width=0.13\linewidth, clip]{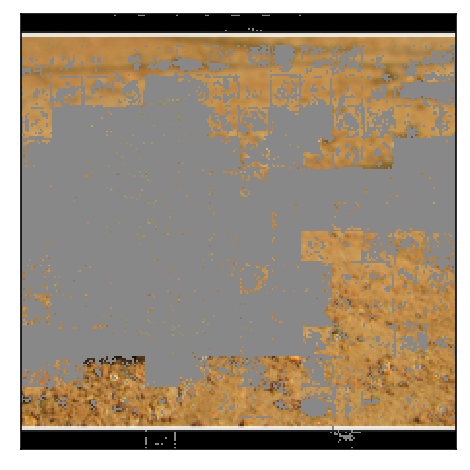}&
\includegraphics[width=0.13\linewidth, clip]{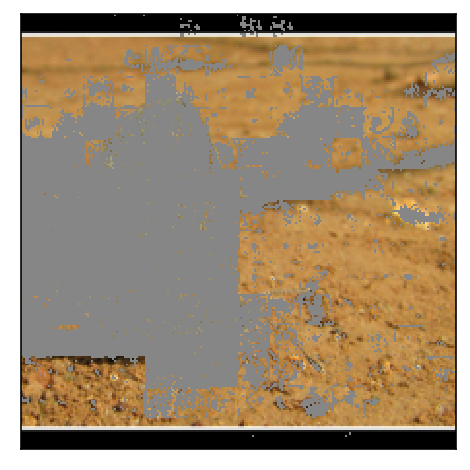}&
\includegraphics[width=0.13\linewidth, clip]{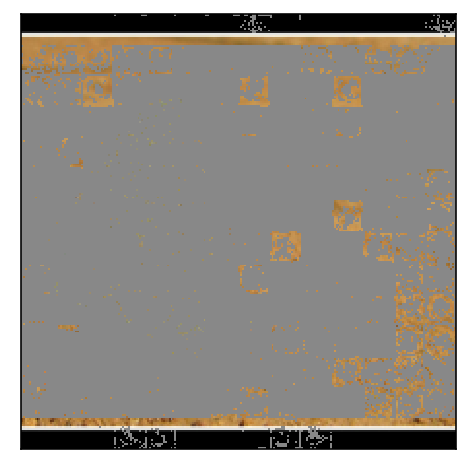}&
\includegraphics[width=0.13\linewidth, clip]{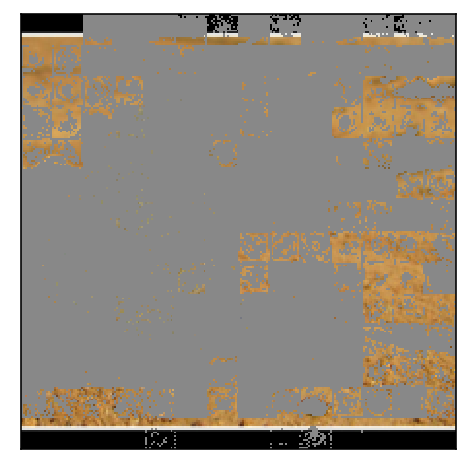}&
\includegraphics[width=0.13\linewidth, clip]{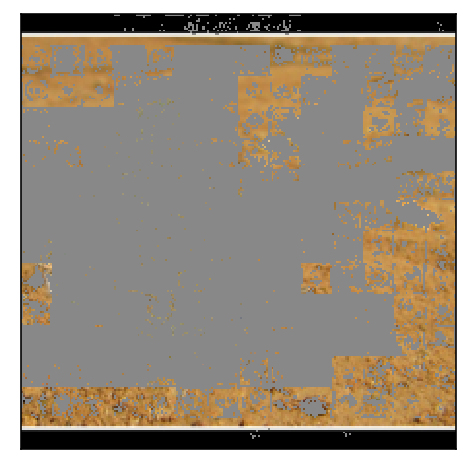}
\\
\small{Lycaenid} &
\small{Lycaenid} &
\small{Gibbon} &
\small{Black window} &
\small{Common iguana} &
\small{Shovel} &
\small{Australian terrier} &
\\
\small{$99.6\%$} &
\small{$93.0\%$}
&
\small{$96.3\%$}
&
\small{$94.3\%$}
&
\small{$98.0\%$}
&
\small{$98.3\%$}
& 
\small{$97.2\%$}
\\
\includegraphics[width=0.13\linewidth, clip]{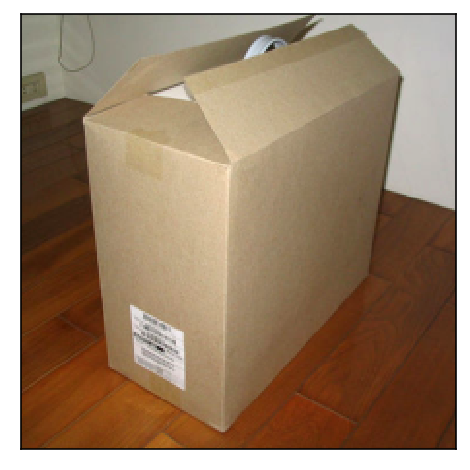}&
\includegraphics[width=0.13\linewidth, clip]{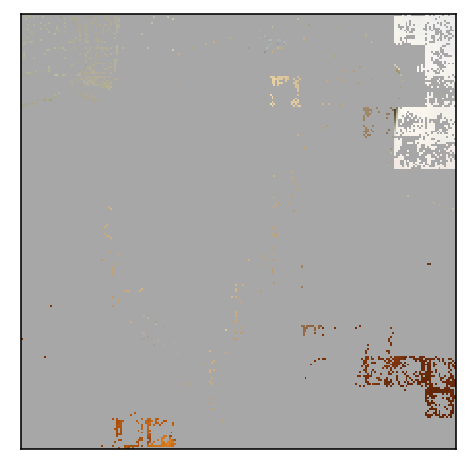}&
\includegraphics[width=0.13\linewidth, clip]{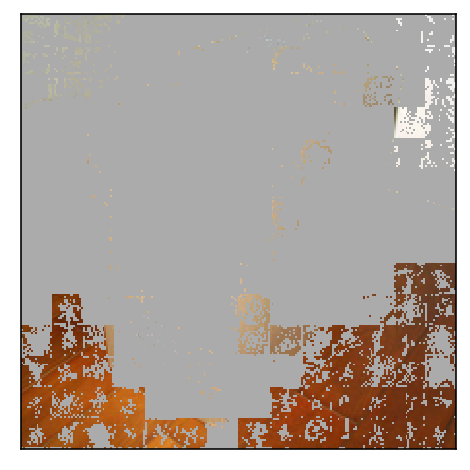}&
\includegraphics[width=0.13\linewidth, clip]{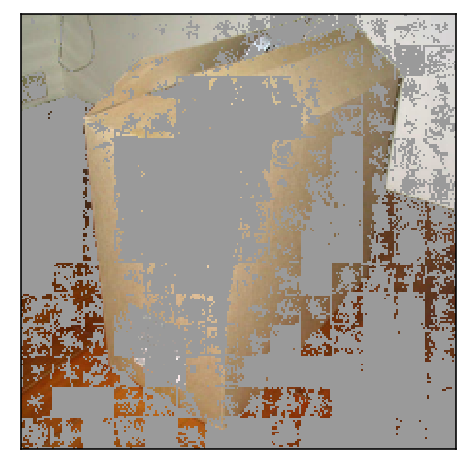}&
\includegraphics[width=0.13\linewidth, clip]{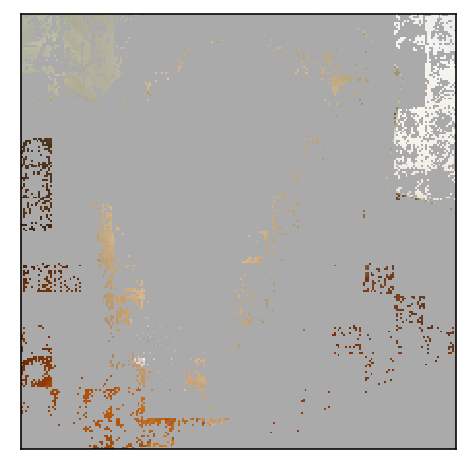}&
\includegraphics[width=0.13\linewidth, clip]{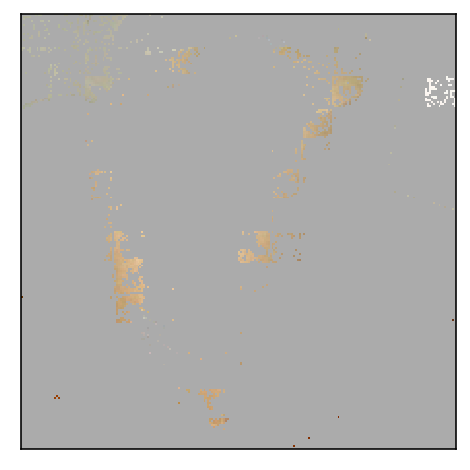}&
\includegraphics[width=0.13\linewidth, clip]{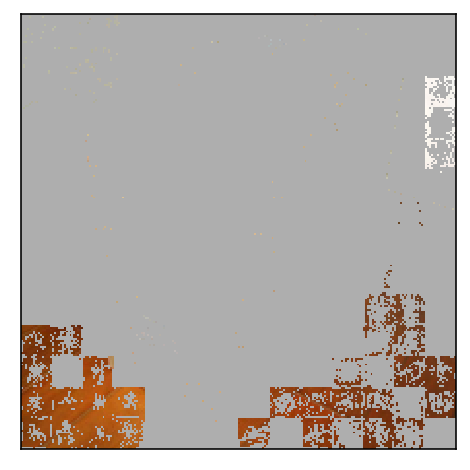}
\\
\small{Carton} &
\small{Carton} &
\small{Gibbon} &
\small{Black window} &
\small{Common iguana} &
\small{Shovel} &
\small{Australian terrier} &
\\
\small{$99.6\%$} &
\small{$90.1\%$} &
\small{$97.9\%$}
&
\small{$91.8\%$}
&
\small{$93.0\%$}
&
\small{$96.8\%$}
&
\small{$93.2\%$}
\\
\includegraphics[width=0.13\linewidth, clip]{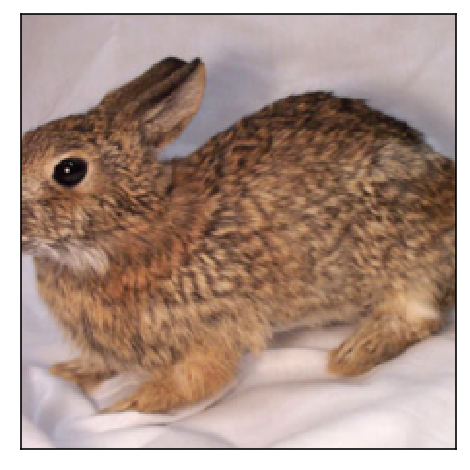}&
\includegraphics[width=0.13\linewidth, clip]{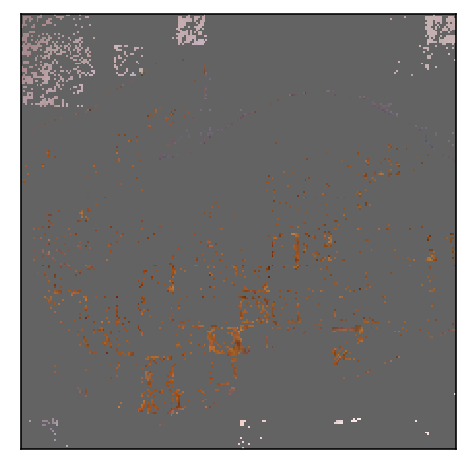}&
\includegraphics[width=0.13\linewidth, clip]{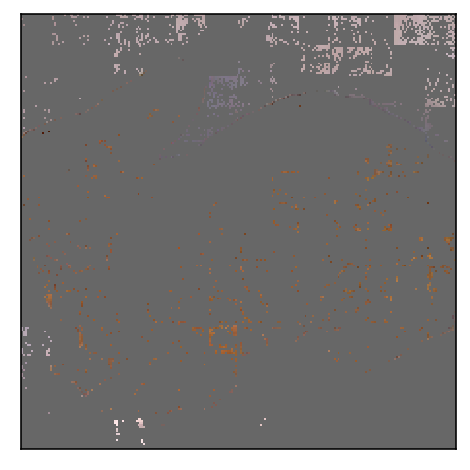}&
\includegraphics[width=0.13\linewidth, clip]{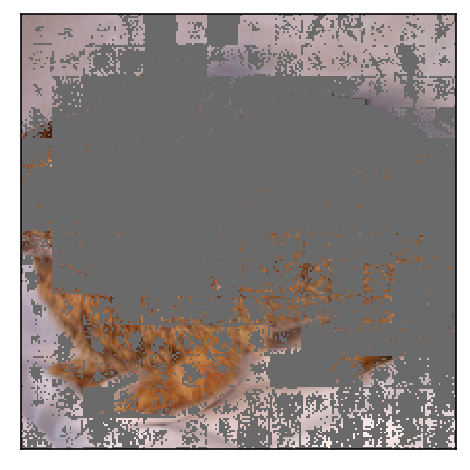}&
\includegraphics[width=0.13\linewidth, clip]{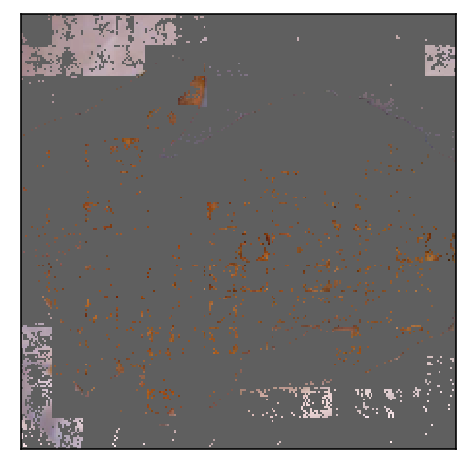}&
\includegraphics[width=0.13\linewidth, clip]{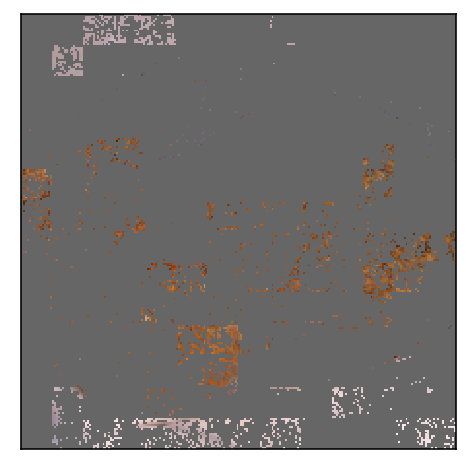}&
\includegraphics[width=0.13\linewidth, clip]{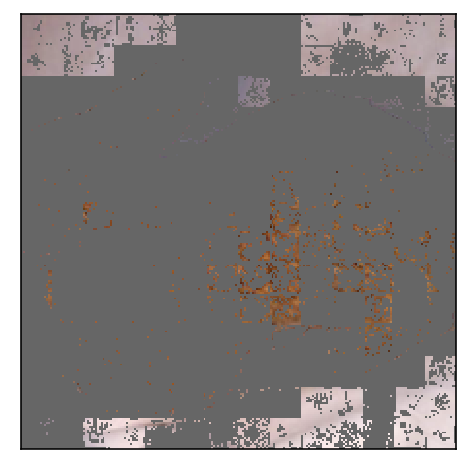}
\\
\small{Wood rabbit} &
\small{Wood rabbit} &
\small{Gibbon} &
\small{Black window} &
\small{Common iguana} &
\small{Shovel} &
\small{Australian terrier} &
\\
\small{$93.4\%$} &
\small{$96.2\%$} &
\small{$93.1\%$}
&
\small{$94.7\%$}
&
\small{$95.4\%$}
&
\small{$92.4\%$}
&
\small{$92.1\%$}
\end{tabular}
}
\caption{Examples of SIS subsets for $4$ random images from the ImageNet validation set with $5$ random ImageNet classes. As can be seen, an SIS subset can be found for all images with all classes with confidence higher than $90\%$, demonstrating that SIS subsets do not necessarily explain the prediction of the model.
\label{fig:overinterpretation}}
\vspace{-16px}
\end{figure*}

Fig.~\ref{fig:overinterpretation} demonstrates the resulting SIS pixels for $4$ randomly selected images from the ImageNet validation set, with $5$ randomly selected ImageNet labels- Gibbon, Black-widow, Common Iguana, Shovel, and Australian Terrier. As can be seen, we were able to find a subset of pixels for each random image and random label that made the classifier predict the random label with a confidence higher than $90\%$. 

This can indicate that there is no guarantee that the classifier used the unnatural cues to decide for a given class based on the existence of a corresponding SIS, i.e. the subset of pixels considered an SIS is not necessarily indicative of the pixels used for the original decision, and resembles an adversarial attack in the sense that it can make the model predict unexpected outputs from a given input. Thus, we opt to use datasets designed specifically for testing robustness~\cite{Geirhos2020ShortcutLI, recht2019imagenet, Hendrycks_2021_CVPR, NEURIPS2019_3eefceb8, barbu2019objectnet}. 

\clearpage

\section{Background: explainability for Vision Transformers}
\label{sec:background}
The Vision Transformer architecture (ViT) splits an image $i$ into $n_i$ fixed-size patches.

Our method employs the Generic Attention Explainability (GAE) method~\cite{Chefer_2021_ICCV} to create a relevance map for the input image tokens. In essence, GAE produces a relevance map $\bar{\mathbf{A}}$ for each self-attention layer. The final relevance map is the result of aggregating all the self-attention layers' relevance maps into a single map $\mathbf{R}(i)$ for the entire network in a forward pass using matrix multiplication.

The aggregated map is initialized using identity matrix: $\mathbf{R}(i) := {I}_{n_i}$.  The relevance map of each self-attention layer of the Vision Transformer is calculated as follows:

 \begin{align}
   \label{eq:modified_att}
    \nabla \mathbf{A} := \frac{\partial \mathcal{M}(i)}{\partial \mathbf{A}}&;& 
    \mathbf{\bar{A}} &= \mathbb{E}_h ((\nabla \mathbf{A} \odot \mathbf{A})^+),
\end{align}
 where $\mathbf{A}$ is the attention map of the current layer, $\odot$ is the Hadamard product, $\mathcal{M}$ is the Vision Transformer, $\mathcal{M}(i)$ is the logit that corresponds to the class we wish to visualize, and $\mathbb{E}_h$ is the mean across the heads dimension. Put differently, the method integrates the gradients of the desired output logit w.r.t. the attention map, in order to average across the attention heads and produce a single unified attention relevance map. The unified attention map $\bar{\mathbf{A}}$ is considered the relevance map of the attention layer. 

Finally, to incorporate each layer's explainability map into the accumulated relevance maps the following propagation rule for self-attention layers is applied during a forward pass on the attention layers:
\begin{align}
    \mathbf{R}(i) \leftarrow  \mathbf{R}(i) + \mathbf{\bar{A}} \cdot  \mathbf{R}(i),
\end{align}
where $\mathbf{\bar{A}}$ is the attention relevance map for the self-attention layers, which is calculated using Eq.~\ref{eq:modified_att}. To extract the relevance of each image token, the row of $\mathbf{R}(i)$ that corresponds to the $[\texttt{CLS}]$ token is used since the $[\texttt{CLS}]$ token alone determines the classification.

In this work, rather than using the relevance map as a form of explainability for a fixed model, we regularize the network to obtain the desired relevance map.

\clearpage

\section{Baseline description}
\label{sec:baselines}
Both GradMask and RRR employ two loss functions. First, a classic cross-entropy loss with the ground-truth labels to ensure correct labeling, and second, gradients of the output of the model w.r.t. the input are used as a form of explanation, to limit the relevance scores of irrelevant parts of the input. This notion resembles our background loss (Eq.~\ref{eq:background}) with gradients as the relevance map.

In the following, we describe the loss terms applied in each method.
For both methods, a standard cross-entropy loss with the ground truth class is applied to maintain accuracy, i.e.:
\begin{align}
    \mathcal{L}_{\text{classification}} =\text{CE}\left(\mathcal{M}(i) , y_i\right),
    \label{eq:gt_loss}
\end{align}
where $i$ is the input image, and $y_i$ is a one-hot vector, where the ground truth classification of $i$ is assigned the value $1$.

GradMask applies a gradient-based loss to ensure that the gradients of the background are close to $0$:
\begin{align}
    \mathcal{L}_{\text{bg}} =\norm{\frac{\partial\hat{y}_i}{\partial i}\cdot \bar{\mathbf{S}}(i)}_2,
    \label{eq:gradmask_bg}
\end{align}
where $\hat{y}_i$ is the predicted output for the ground-truth class, and $\bar{\mathbf{S}}(i)$ is, as before, the reversed segmentation map for the ground-truth class.
Eq.~\ref{eq:gradmask_bg} resembles our background loss (Eq.~\ref{eq:background}) with simple gradients w.r.t. the input image instead of the relevance map produced by GAE.

Similarly, Right for the Right Reasons (RRR) applies a loss to restrain the magnitude of explanations outside the relevant information. Their relevance loss is obtained as follows:
\begin{align}
    \mathcal{L}_{\text{bg}} =\left(\frac{\partial\sum_{k=1}^K\log({p}_{i,k})}{\partial i}\cdot \bar{\mathbf{S}}(i)\right)^2,
    \label{eq:rrr_bg}
\end{align}
where 
$k=1,...,K$ are the possible output classes, and ${p}_{i,k}$ is the probability assigned to the $k$-th class for image $i$ by the model.

For both GradMask and RRR, the following loss is used $
    \mathcal{L}_{\text{final}} =\lambda_{\text{bg}}\cdot \mathcal{L}_{\text{bg}} + \lambda_{\text{classification}}\cdot \mathcal{L}_{\text{classification}}$,
where $\lambda_{\text{bg}}, \lambda_{\text{classification}}$ are hyperparameters. We note that while using the gradient of the output w.r.t. the input is common practice for interpreting CNNs, these gradients are less stable for transformer-based models. For example, results presented in~\cite{Liu2022RethinkingAE} demonstrate that for transformer-based models the classic Input$\times$Gradient method violates faithfulness. 

In our experiments, we found it difficult to grid-search hyperparameters to fit $\mathcal{L}_{\text{bg}}$ and $\mathcal{L}_{\text{classification}}$. Furthermore, we had to tune $\lambda_{\text{bg}}, \lambda_{\text{classification}}$ for each model separately to obtain an improvement for $\mathcal{L}_{\text{bg}}$. Our method, on the other hand, uses the same hyperparameter choice (see Sec.~\ref{sec:method}), which makes it far more stable to use, thus allowing us to run experiments on large models as well. We refer the reader to Appendix~\ref{sec:hyperparameters} for the full description of hyperparameters used in our experiments. 

\clearpage

\section{Hyperparameters}
\label{sec:hyperparameters}
In Tab.~\ref{table:hyperparam-ours} we present the hyperparameter selection for all our experiments. Our method is stable and uses the same selection in all cases, other than the learning rate. The learning rates range from $6e-7$ to $3e-6$, allowing for a quick and easy grid search.

Tab.~\ref{table:hyperparam-RRR},~\ref{table:hyperparam-GradMask} represent the hyperparameters of RRR, GradMask, respectively. For RRR, we had to tune the parameters per model, and the method is sensitive to the specific selection. For GradMask, we had to carefully tune the learning rate for each model. We found that the results of GradMask are sensitive to the specific learning rate selection, and to minor changes in the learning rate. We found it difficult to get the background loss to converge in both cases.
\begin{table}[h!]
\caption{
    Hyperparameter selection for our method for all models- ViT~\cite{dosovitskiy2020image}, ViT AugReg~\cite{Steiner2021HowTT}, and DeiT~\cite{touvron2020training}. All hyperparameters are fixed except for the learning rate.
    }
    \begin{center}
    \begin{tabular}{@{}l@{~~~}c@{~~~~}c@{~~~~}c@{~~~~}c@{~~~~}c@{}}
        \toprule
        Model& $\lambda_{\text{classification}}$ & $\lambda_{\text{relevance}}$  & $\lambda_{\text{bg}}$ & $\lambda_{\text{fg}}$& Learning rate \\
        \midrule
        ViT-B & 0.2 & 0.8 & 2 & 0.3 & $3e-6$\\
        \midrule
        ViT-L & 0.2 & 0.8 & 2 & 0.3 & $9e-7$\\
        \midrule
        AR-S & 0.2 & 0.8 & 2 & 0.3 & $2e-6$\\
        \midrule
        AR-B & 0.2 & 0.8 & 2 & 0.3 & $6e-7$\\
        \midrule
        AR-L & 0.2 & 0.8 & 2 & 0.3 & $9e-7$\\
        \midrule
        DeiT-S & 0.2 & 0.8 & 2 & 0.3 & $1e-6$\\
        \midrule
        DeiT-B & 0.2 & 0.8 & 2 & 0.3 & $8e-7$\\
        \bottomrule
    \end{tabular}
    \smallskip
    \smallskip
    \label{table:hyperparam-ours}
    \end{center}
\end{table}

\begin{table}[h!]
\caption{
    Hyperparameter selection for the Right for the Right Reasons~\cite{Ross2017RightFT} method for all models- ViT~\cite{dosovitskiy2020image}, ViT AugReg~\cite{Steiner2021HowTT}, and DeiT~\cite{touvron2020training}. Hyperapramters vary according to the model.
    }
    \begin{center}
    \begin{tabular}{@{}l@{~~~}c@{~~~~}c@{~~~~}c@{}}
        \toprule
        Model& $\lambda_{\text{classification}}$ & $\lambda_{\text{bg}}$ & Learning rate \\
        \midrule
        ViT-B & $2e-6$ & $1e-10$ & $2e-6$ \\
        \midrule
        AR-S & $2e-8$ & $1e-8$ & $1e-5$ \\
        \midrule
        AR-B & $2e-7$ & $1e-8$ & $5e-7$  \\
        \midrule
        DeiT-S & $2e-6$ & $1e-10$ & $1e-5$   \\
        \midrule
        DeiT-B & $2e-6$ & $1e-10$ & $5e-6$  \\
        \bottomrule
    \end{tabular}
    \smallskip
    \smallskip
    \label{table:hyperparam-RRR}
    \end{center}
\end{table}

\begin{table}[]
\caption{
    Hyperparameter selection for the GradMask~\cite{Simpson2019GradMaskRO} method for all models- ViT~\cite{dosovitskiy2020image}, ViT AugReg~\cite{Steiner2021HowTT}, and DeiT~\cite{touvron2020training}.
    }
    \begin{center}
    \begin{tabular}{@{}l@{~~~}c@{~~~~}c@{~~~~}c@{}}
        \toprule
        Model& $\lambda_{\text{classification}}$ & $\lambda_{\text{bg}}$ & Learning rate \\
        \midrule
        ViT-B & $3e-9$ & $50$ & $0.0002$ \\
        \midrule
        AR-S & $3e-9$ & $50$ & $0.0005$ \\
        \midrule
        AR-B & $3e-9$ & $50$ & $1e-5$   \\
        \midrule
        DeiT-S  & $3e-9$ & $50$ & $0.005$   \\
        \midrule
        DeiT-B  & $3e-9$ & $50$ & $0.001$    \\
        \bottomrule
    \end{tabular}
    \smallskip
    \smallskip
    \label{table:hyperparam-GradMask}
    \end{center}
\end{table}

\clearpage

\section{Random seed selection of the training classes}
\label{sec:random_seeds}

Tab.~\ref{table:seeds_top1}, and Tab.~\ref{table:seeds_top5} present the results for the main experiment for multiple seeds, extending Tab.~1 and Tab.~2 of the main text. Each seed changes the 500 random classes used for finetuning with exactly the same hyperparameters.

As can be seen, the Standard Deviation is not large, especially in comparison to the performance gap. In all shifted-distribution datasets, the original result, before our intervention, is outside the standard error range for the multiple-seed experiments.

\begin{table}[h!]
            \caption{Results using $3$ random seeds. The table presents the average and standard deviation of the top-1 accuracy per dataset.
    }
    \begin{center}
    \resizebox{\linewidth}{!}{
    \begin{tabular}{@{}l@{~~}c@{~~~}c@{~~~}c@{~~~}c@{~~~}c@{~~~}c@{~~~}c@{~~~~}c@{~~~}c@{~~~}c@{}}
        \toprule
        {Model} & {Method} &{INet val} & {INet-A} & {INet-R}& {Sketch} 
        & {INet-v2}
        & {ObjNet} & {SI-loc.} & {SI-rot.}  & {SI-size} 
        \\
        \midrule
        \multirow{2}{*}{\small ViT-B }  &\small{Original} & \textbf{81.5} & 16.0 & 33.8 & 35.4 & \textbf{71.1} & 35.1 & 33.3 & 39.1 & 55.6 \\
&\small{Ours} & 80.3$\pm$0.1 & \textbf{23.6$\pm$0.7} & \textbf{36.1$\pm$0.3} & \textbf{36.3$\pm$0.1} & 70.1$\pm$0.3 & \textbf{41.7$\pm$0.6} & \textbf{38.6$\pm$0.2} & \textbf{46.1$\pm$0.4} & \textbf{60.9$\pm$0.4}
\\
\midrule
\multirow{2}{*}{\small ViT-L }  &\small{Original} & \textbf{82.9} & 19.0 & 36.6 & 40.4 & \textbf{71.8} & 37.4 & 31.6 & 40.7 & 54.8 \\
&\small{Ours} & 82.0$\pm$0.1 & \textbf{25.0$\pm$0.3} & \textbf{38.6$\pm$0.4} & \textbf{41.1$\pm$0.1} & 71.2$\pm$0.2 & \textbf{42.6$\pm$0.2} & \textbf{36.6$\pm$0.4} & \textbf{45.3$\pm$0.4} & \textbf{58.6$\pm$0.5} 
\\
\midrule

\multirow{2}{*}{\small AR-S }  &\small{Original} & \textbf{81.4} & 13.0 & 31.2 & 32.8 & \textbf{69.9} & 34.3 & 32.4 & 40.6 & 55.4 \\
&\small{Ours} & 79.8$\pm$0.2 & \textbf{18.3$\pm$0.8} & \textbf{34.1$\pm$0.2} & \textbf{33.6$\pm$0.1} & 69.4$\pm$0.2 & \textbf{39.1$\pm$0.4} & \textbf{36.4$\pm$0.9} & \textbf{47.6$\pm$0.9} & \textbf{61.0$\pm$0.8} \\
\midrule
\multirow{2}{*}{\small AR-B }  &\small{Original} & \textbf{84.4} & 23.9 & 41.0 & 43.1 & \textbf{73.8} & 41.4 & 40.5 & 48.1 & 60.6 \\
&\small{Ours} & 83.0$\pm$0.1 & \textbf{30.9$\pm$0.3} & \textbf{45.3$\pm$0.7} & \textbf{44.9$\pm$0.4} & 73.3$\pm$0.2 & \textbf{47.0$\pm$0.6} & \textbf{43.8$\pm$0.7} & \textbf{54.7$\pm$0.6} & \textbf{64.6$\pm$0.5} \\
\midrule
\multirow{2}{*}{\small AR-L }  &\small{Original} & \textbf{85.6} & 34.7 & 48.8 & 51.8 & \textbf{75.8} & 46.5 & 43.8 & 52.4 & 62.3 \\
&\small{Ours} & 84.8$\pm$0.3 & \textbf{42.2$\pm$0.4} & \textbf{53.9$\pm$0.3} & \textbf{54.0$\pm$0.3} & 75.7$\pm$0.1 & \textbf{51.7$\pm$0.2} & \textbf{48.6$\pm$0.6} & \textbf{57.8$\pm$0.8} & \textbf{66.6$\pm$0.3} \\

\midrule
\multirow{2}{*}{\small DeiT-S }  &\small{Original} & 78.1 & 8.3 & 28.2 & 28.8 & 66.5 & 28.3 & 30.7 & 36.7 & 51.6 \\
&\small{Ours} & \textbf{78.7$\pm$0.1} & \textbf{10.4$\pm$0.3} & \textbf{29.3$\pm$0.1} & \textbf{29.0$\pm$0.2} & \textbf{67.3$\pm$0.3} & \textbf{31.5$\pm$0.1} & \textbf{32.2$\pm$0.2} & \textbf{40.6$\pm$0.4} & \textbf{55.6$\pm$0.4} \\
\midrule
\multirow{2}{*}{\small DeiT-B }  &\small{Original} & \textbf{80.8} & 12.9 & 30.9 & \textbf{31.2} & \textbf{69.7} & 31.4 & 34.5 & 39.3 & 54.6 \\
&\small{Ours} & 80.6$\pm$0.2 & \textbf{17.2$\pm$0.2} & \textbf{32.7$\pm$0.3} & \textbf{31.2$\pm$0.3} & 69.3$\pm$0.2 & \textbf{35.9$\pm$0.2} & \textbf{37.0$\pm$0.4} & \textbf{43.3$\pm$0.4} & \textbf{58.3$\pm$0.3} \\

        \bottomrule
    \end{tabular}
    }
    \smallskip
    \smallskip
    \label{table:seeds_top1}
    \end{center}
    \vspace{-24px}
\end{table}

\begin{table}[h!]
            \caption{Results using $3$ random seeds. The table presents the average and standard deviation of the top-5 accuracy per dataset.
    }
    \begin{center}
    \resizebox{\linewidth}{!}{
    \begin{tabular}{@{}l@{~~}c@{~~~}c@{~~~}c@{~~~}c@{~~~}c@{~~~}c@{~~~}c@{~~~}c@{~~~}c@{~~~}c@{}}
        \toprule
        {Model} & {Method} &{INet val} & {INet-A} & {INet-R}& {Sketch} 
        & {INet-v2}
        & {ObjNet} & {SI-loc.} & {SI-rot.}  & {SI-size} 
        \\
        \midrule
\multirow{2}{*}{\small ViT-B }  &\small{Original} & \textbf{96.0} & 37.0 & 48.5 & 57.4 & \textbf{89.9} & 56.4 & 52.2 & 58.3 & 76.2 \\
&\small{Ours} & 95.4$\pm$0.0 & \textbf{47.1$\pm$0.8} & \textbf{51.3$\pm$0.2} & \textbf{58.7$\pm$0.3} & 89.4$\pm$0.1 & \textbf{64.7$\pm$0.5} & \textbf{57.5$\pm$0.3} & \textbf{66.6$\pm$0.5} & \textbf{81.2$\pm$0.3} \\
\midrule
\multirow{2}{*}{\small ViT-L }  &\small{Original} & \textbf{96.4} & 41.5 & 52.0 & 63.4 & \textbf{90.7} & 59.5 & 50.3 & 60.1 & 75.6 \\
&\small{Ours} & 96.2$\pm$0.0 & \textbf{49.2$\pm$0.6} & \textbf{54.6$\pm$0.3} & \textbf{64.3$\pm$0.1} & 90.6$\pm$0.3 & \textbf{65.5$\pm$0.2} & \textbf{56.4$\pm$0.3} & \textbf{66.3$\pm$0.5} & \textbf{80.3$\pm$0.3} \\
\midrule
\multirow{2}{*}{\small AR-S }  &\small{Original} & \textbf{96.1} & 33.9 & 47.1 & 54.2 & \textbf{90.1} & 55.8 & 51.7 & 59.6 & 75.7 \\
&\small{Ours} & 95.6$\pm$0.1 & \textbf{40.9$\pm$1.1} & \textbf{50.2$\pm$0.4} & \textbf{55.3$\pm$0.1} & 89.9$\pm$0.1 & \textbf{61.8$\pm$0.6} & \textbf{56.1$\pm$0.8} & \textbf{67.6$\pm$0.8} & \textbf{81.2$\pm$0.5} \\
\midrule
\multirow{2}{*}{\small AR-B }  &\small{Original} & \textbf{97.2} & 49.2 & 57.8 & 65.7 & \textbf{92.3} & 63.7 & 60.8 & 68.3 & 80.4 \\
&\small{Ours} & 96.8$\pm$0.1 & \textbf{56.8$\pm$0.3} & \textbf{62.0$\pm$0.7} & \textbf{67.9$\pm$0.6} & 91.9$\pm$0.1 & \textbf{69.9$\pm$0.5} & \textbf{63.5$\pm$0.8} & \textbf{75.1$\pm$0.6} & \textbf{84.4$\pm$0.5} \\

\midrule
\multirow{2}{*}{\small AR-L }  &\small{Original} & \textbf{97.8} & 61.0 & 64.9 & 73.6 & \textbf{93.4} & 68.3 & 64.2 & 72.5 & 82.2 \\
&\small{Ours} & 97.4$\pm$0.1 & \textbf{67.3$\pm$0.6} & \textbf{69.2$\pm$0.4} & \textbf{75.7$\pm$0.2} & 93.3$\pm$0.1 & \textbf{73.5$\pm$0.4} & \textbf{68.8$\pm$0.7} & \textbf{78.1$\pm$0.8} & \textbf{86.3$\pm$0.3} \\

\midrule
\multirow{2}{*}{\small DeiT-S }  &\small{Original} & 93.7 & 23.5 & 41.9 & 46.7 & 86.6 & 47.3 & 50.4 & 54.3 & 72.0 \\
&\small{Ours} & \textbf{94.5$\pm$0.1} & \textbf{29.0$\pm$0.5} & \textbf{43.6$\pm$0.0} & \textbf{47.6$\pm$0.4} & \textbf{87.4$\pm$0.1} & \textbf{53.0$\pm$0.1} & \textbf{51.3$\pm$0.2} & \textbf{59.3$\pm$0.5} & \textbf{76.1$\pm$0.3} \\
\midrule
\multirow{2}{*}{\small DeiT-B }

&\small{Original} & 94.2 & 31.0 & 44.2 & 48.6 & 86.8 & 48.5 & 54.6 & 56.3 & 73.4 \\ 
&\small{Ours} & \textbf{95.0$\pm$0.1} & \textbf{40.1$\pm$0.2} & \textbf{47.4$\pm$0.4} & \textbf{49.7$\pm$0.4} & \textbf{88.4$\pm$0.1} & \textbf{56.3$\pm$0.2} & \textbf{57.4$\pm$0.4} & \textbf{61.9$\pm$0.5} & \textbf{78.4$\pm$0.3} \\
        \bottomrule
    \end{tabular}
    }
    \smallskip
    \smallskip
    \label{table:seeds_top5}
    \end{center}
    \vspace{-24px}
\end{table}

\clearpage

\section{Additional qualitative results}
\label{sec:qualitative}
\begin{figure*}[h!]
  \centering
  \resizebox{\linewidth}{!}{
\begin{tabular}{c@{~~}c@{~}c@{~}c@{~~~}c@{~}c@{~}c@{~~~}c@{~}c@{~}c}
& \multicolumn{3}{c}{INet-A}& \multicolumn{3}{c}{ObjectNet} & \multicolumn{3}{c}{SI-rotation}\\
\cmidrule(r){2-4}
\cmidrule(r){5-7}
\cmidrule(r){8-10}
& \small{Input} & \small {Original} & {\small{Ours}} &  \small{Input} & \small {Original} & {\small{Ours}} &  \small{Input} & \small {Original} & {\small{Ours}} \\
{\begin{turn}{90}~~~ ViT-B \end{turn}} & 
\includegraphics[width=0.1\linewidth, clip]{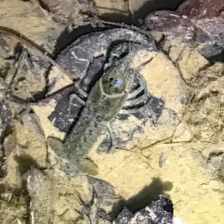}&
\includegraphics[width=0.1\linewidth, clip]{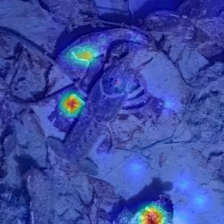}&
\includegraphics[width=0.1\linewidth, clip]{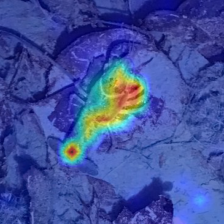}&
\includegraphics[width=0.1\linewidth, clip]{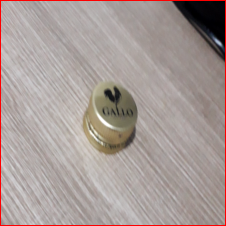}&
\includegraphics[width=0.1\linewidth, clip]{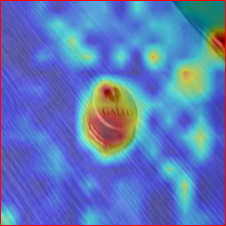}&
\includegraphics[width=0.1\linewidth, clip]{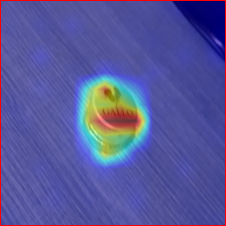}&
\includegraphics[width=0.1\linewidth, clip]{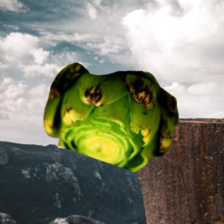}&
\includegraphics[width=0.1\linewidth, clip]{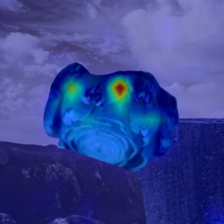}&
\includegraphics[width=0.1\linewidth, clip]{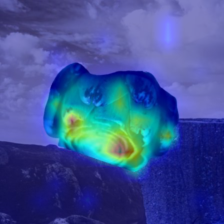}
\\
\multirow{2}{*}{\begin{turn}{90} Pred \end{turn}}
&
&{\small{Sea snake}} & {\small{Crayfish}} & & {\small{Electric}} & {\small{Bottle-}} &  & {\small{Green}} & {\small{Artichoke}}\\
& & & & & {\small{guitar}}  & {\small{cap}} & & \small{Mamba}  \\
{\begin{turn}{90}~~~ ViT-L \end{turn}} & 
\includegraphics[width=0.1\linewidth, clip]{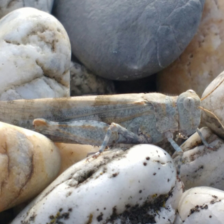}&
\includegraphics[width=0.1\linewidth, clip]{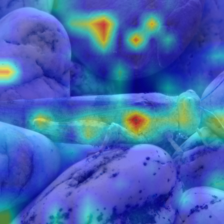}&
\includegraphics[width=0.1\linewidth, clip]{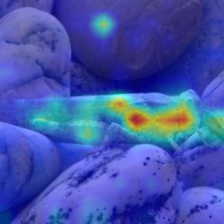}&
\includegraphics[width=0.1\linewidth, clip]{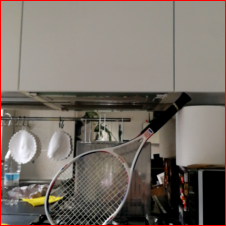}&
\includegraphics[width=0.1\linewidth, clip]{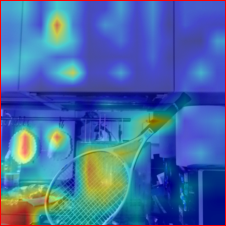}&
\includegraphics[width=0.1\linewidth, clip]{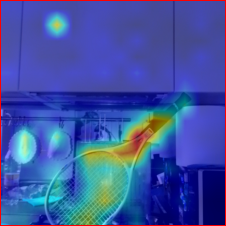}&
\includegraphics[width=0.1\linewidth, clip]{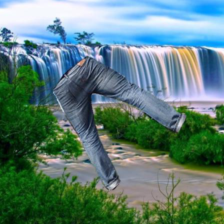}&
\includegraphics[width=0.1\linewidth, clip]{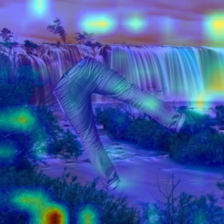}&
\includegraphics[width=0.1\linewidth, clip]{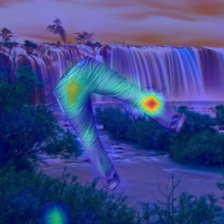}
\\
\multirow{2}{*}{\begin{turn}{90} Pred \end{turn}}
&
&{\small{Rock-}} & {\small{Grass-}} & & {\small{Strainer}} & {\small{Racket}} &  & {\small{lakeside}} & {\small{Jeans}}\\
& &{\small{crab}} & \small{hopper}& &  & & & {\small{}}
\\
{\begin{turn}{90}~~~ AR-B \end{turn}} & 
\includegraphics[width=0.1\linewidth, clip]{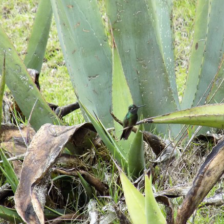}&
\includegraphics[width=0.1\linewidth, clip]{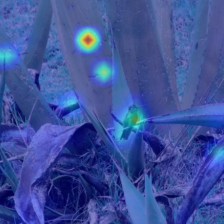}&
\includegraphics[width=0.1\linewidth, clip]{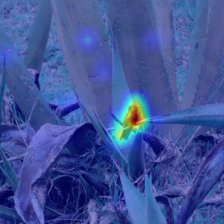}&
\includegraphics[width=0.1\linewidth, clip]{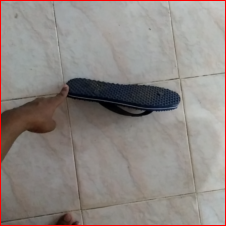}&
\includegraphics[width=0.1\linewidth, clip]{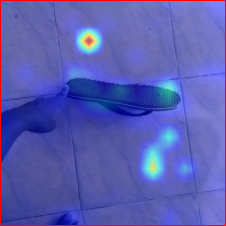}&
\includegraphics[width=0.1\linewidth, clip]{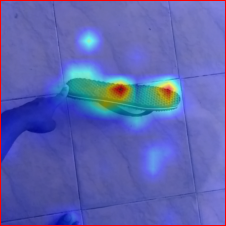}&
\includegraphics[width=0.1\linewidth, clip]{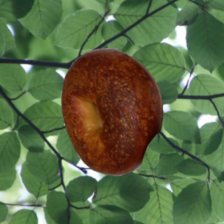}&
\includegraphics[width=0.1\linewidth, clip]{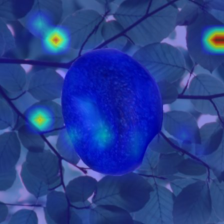}&
\includegraphics[width=0.1\linewidth, clip]{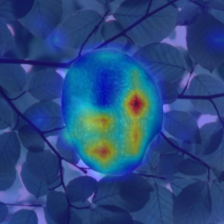}
\\
\multirow{2}{*}{\begin{turn}{90} Pred \end{turn}}
&
&{\small{Vine-}} & {\small{Humming-}} & & {\small{Mop}} & {\small{Sandal}} &  & {\small{Horse-}} & {\small{Bagel}}\\
& &{\small{snake}} & \small{bird} & &  {\small{}} & {\small{}} & &{\small{chestnut}}  \\
{\begin{turn}{90}~~~ AR-L \end{turn}} & 
\includegraphics[width=0.1\linewidth, clip]{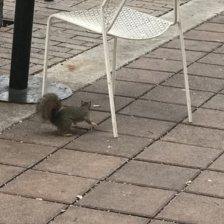}&
\includegraphics[width=0.1\linewidth, clip]{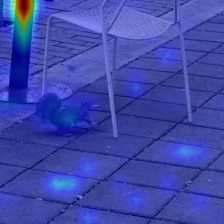}&
\includegraphics[width=0.1\linewidth, clip]{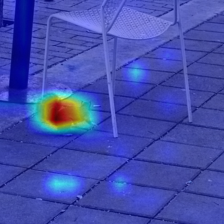}&
\includegraphics[width=0.1\linewidth, clip]{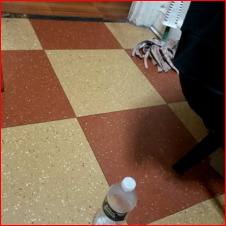}&
\includegraphics[width=0.1\linewidth, clip]{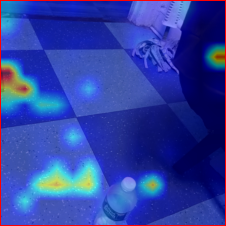}&
\includegraphics[width=0.1\linewidth, clip]{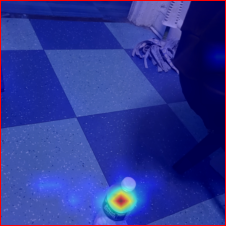}&
\includegraphics[width=0.1\linewidth, clip]{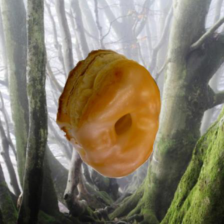}&
\includegraphics[width=0.1\linewidth, clip]{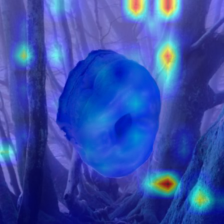}&
\includegraphics[width=0.1\linewidth, clip]{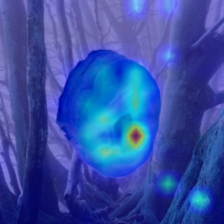}
\\
\multirow{2}{*}{\begin{turn}{90} Pred \end{turn}}
&
&{\small{Trash-}} & {\small{Fox-}} & & {\small{Vacuum}} & {\small{Water-}} &  & {\small{Hen of the}} & {\small{Bagel}}\\
& & \small{can}& \small{squirrel} & & &\small{bottle} & & \small{woods}
\\
{\begin{turn}{90}~ DeiT-B \end{turn}} & 
\includegraphics[width=0.1\linewidth, clip]{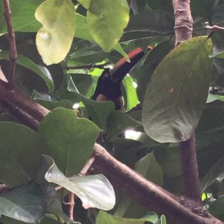}&
\includegraphics[width=0.1\linewidth, clip]{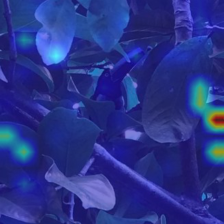}&
\includegraphics[width=0.1\linewidth, clip]{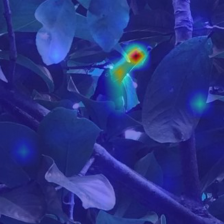}&
\includegraphics[width=0.1\linewidth, clip]{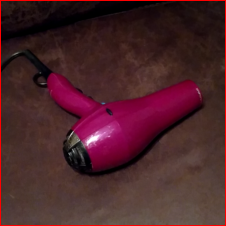}&
\includegraphics[width=0.1\linewidth, clip]{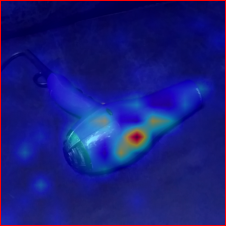}&
\includegraphics[width=0.1\linewidth, clip]{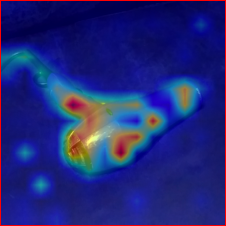}&
\includegraphics[width=0.1\linewidth, clip]{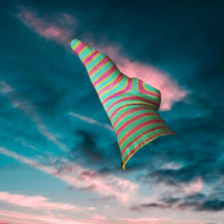}&
\includegraphics[width=0.1\linewidth, clip]{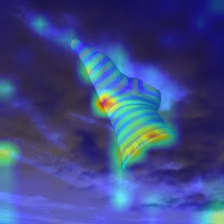}&
\includegraphics[width=0.1\linewidth, clip]{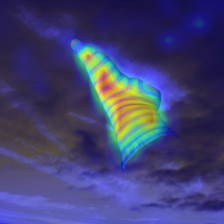}
\\
\multirow{2}{*}{\begin{turn}{90} Pred \end{turn}}
&
&{\small{Lemon}} & {\small{Toucan}} & & {\small{Maraca}} & {\small{Hand-}} &  & {\small{Parachute}} & {\small{Sock}}\\
& & & & & & {\small{blower}} & & 
\end{tabular}
}
\caption{Examples of cases where our method corrects wrong predictions, alongside the original and modified (after finetuning) explainability maps (please zoom in for a better view). The ``Pred" row specifies the predictions before and after our finetuning. The examples demonstrate cases where the original classifier relies on partial or irrelevant data, while our method rectifies the classification to be based on the object. The examples are presented for the large and base models of ViT~\cite{dosovitskiy2020image}, ViT AugReg~\cite{Steiner2021HowTT} (AR), and the base model of DeiT~\cite{touvron2020training}.
\label{fig:robustness_full}}

\end{figure*}

Fig.~\ref{fig:robustness_full} provides more examples of cases where the salient behavior of the models prevents them from producing correct predictions. For example, an artichoke is classified as a green mamba due to partial consideration of the foreground (third example in the first row), a bagel is classified as a horse-chestnut due to the leaves in the background (third example in the third row), a racket is classified as a strainer due to the kitchen setting (second example in the second row), and a grasshopper is classified as a rock crab due to the rocks in the background (first example in the second row). By correcting the relevance maps, our method assists the models to achieve an accurate prediction.

\clearpage

\section{Sensitivity tests}
\label{sec:sensitivity}
\begin{figure*}[h!]%
    \centering
    \begin{tabular}{c@{~~~~~~~~~~~~~~~~}c@{}}
    {{\includegraphics[width=5cm]{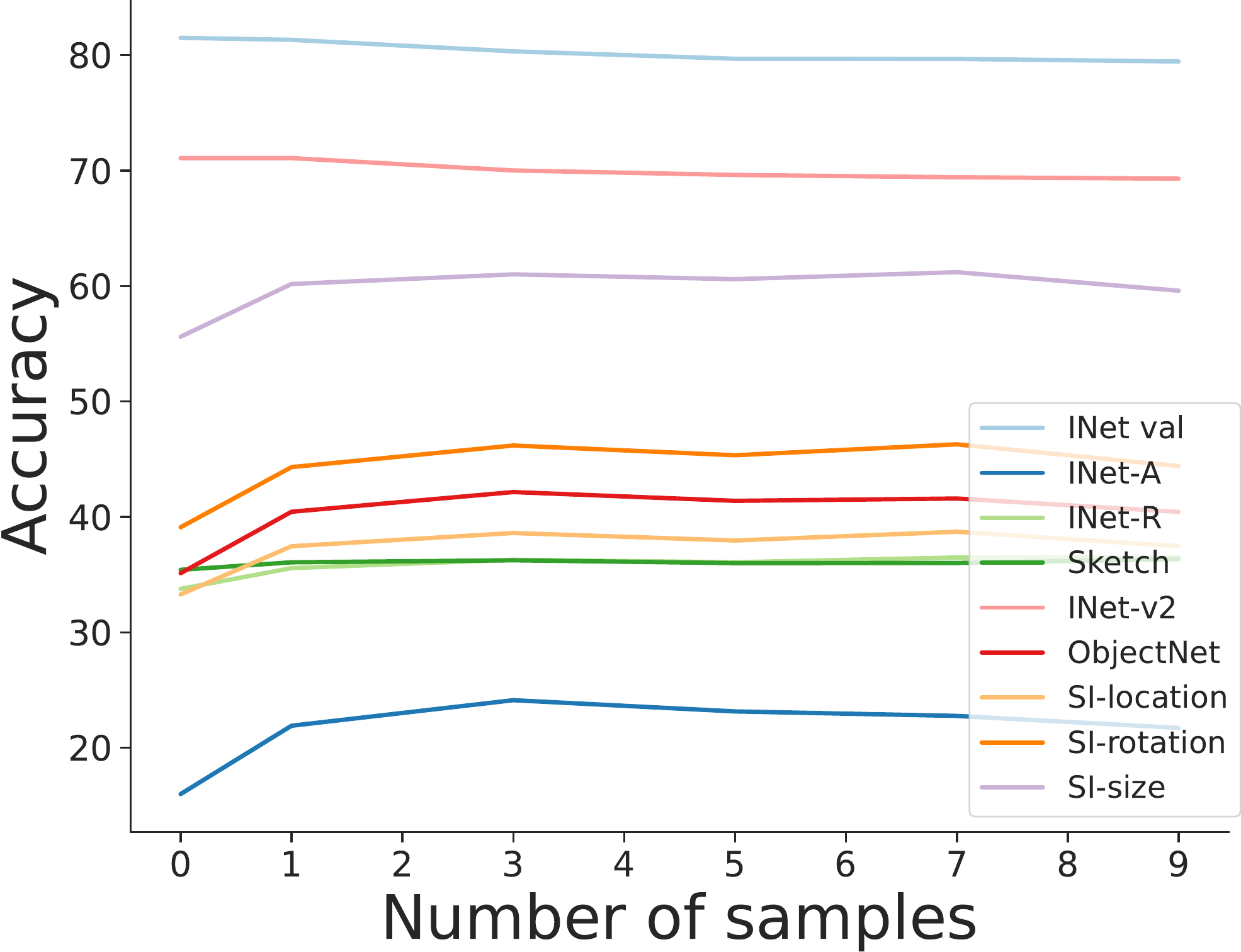} }}&
    {{\includegraphics[width=5cm]{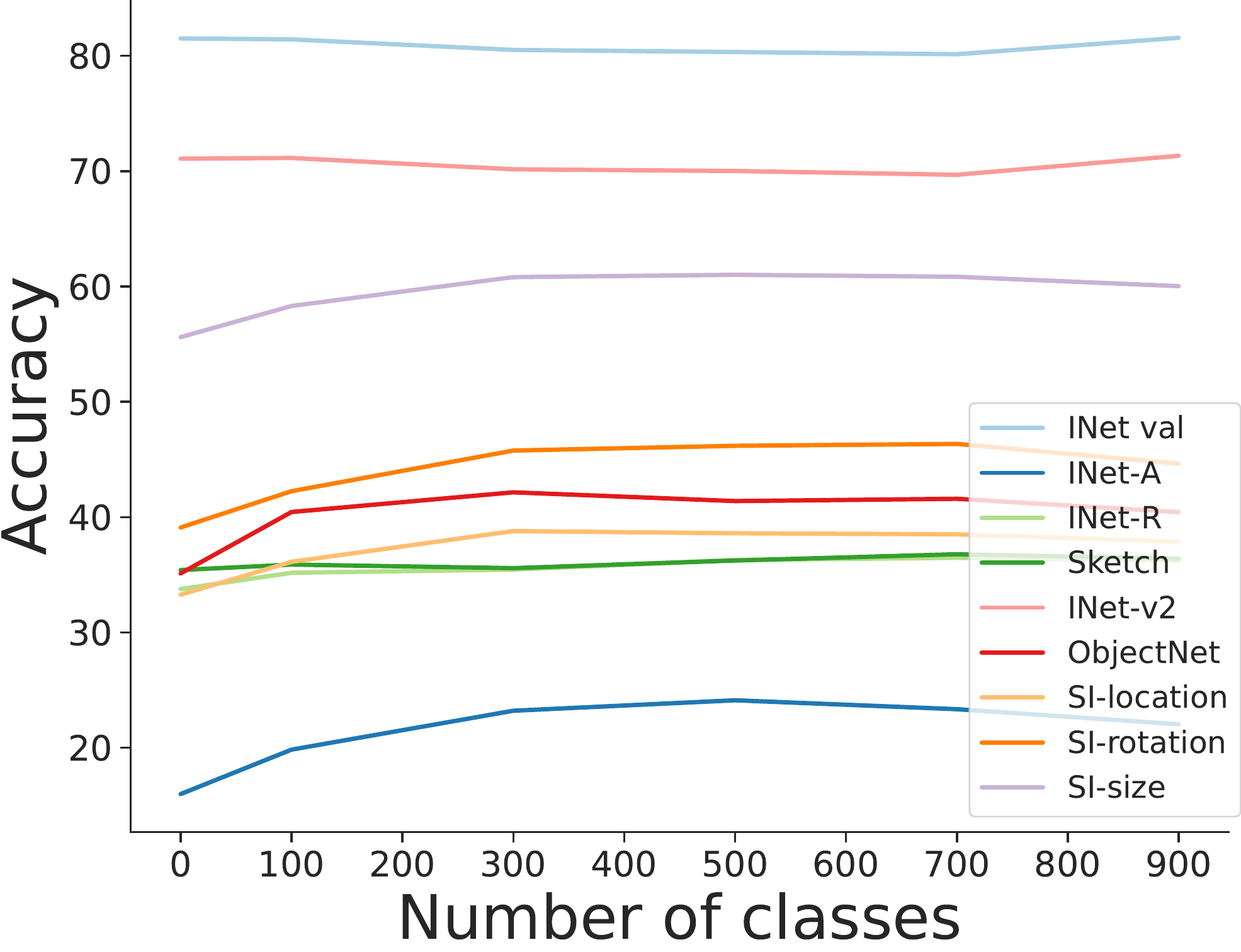}}}
    \\
    (a) & (b)\\
    \end{tabular}
    \caption{Evaluation of our method's sensitivity to (a) the number of samples used per class in the training set (we use $n=3$), and (b) number of training classes (we use $c=500$), on ViT-B~\cite{dosovitskiy2020image}. As can be seen, all the presented combinations of hyperparameters result in a significant increase in robustness, and a relatively modest decrease in INet val, INet-v2 accuracy.}%
    \label{fig:sensitivity-num-samples}%
    \vspace{-14px}
\end{figure*}

Fig.~\ref{fig:sensitivity-num-samples}, presents sensitivity tests evaluating our robustness results on increasing number of samples per class (panel a), and increasing number of classes (panel b). Evidently, three samples for half the classes suffice to achieve the maximal improvement in robustness, while minimally harming the performance on ImageNet-based datasets (ImageNet val, ImagNet-v2). Using only two samples or considerably fewer classes does not harm performance much.

\clearpage

\section{Comparing training classes to the other classes}
\label{sec:ab}
Tab.~\ref{table:AB-test},~\ref{table:AB-test-SI} contain the full results of our experiment comparing the effect of our method on the classes in the training set and the classes outside of it. As can be seen, both the training and non-training classes benefit very similarly from applying our method when evaluated in terms of robustness on both real-world and synthetic datasets.

\begin{table}[h!]
\caption{
    Robustness evaluation on the classes that were included in the training set of our finetuning and classes that were not. The results are presented for the base models of ViT~\cite{dosovitskiy2020image}, ViT AugReg~\cite{Steiner2021HowTT}, and DeiT~\cite{touvron2020training}. The last row indicates the average change for the training and non-training classes across the models for each dataset. 
    }
    \begin{center}
    \resizebox{\linewidth}{!}{
    \begin{tabular}{@{}l@{~~~}c@{~~}c@{~~}c@{~}c@{~~}c@{~}c@{~~}c@{~}c@{~~}c@{~}c@{~~}c@{~}c@{~~}c@{~}c@{}}
        \toprule
        \multirow{2}{*}{Model} & Train & &\multicolumn{2}{c}{INet val} & \multicolumn{2}{c}{INet-A} & \multicolumn{2}{c}{INet-R}& \multicolumn{2}{c}{Sketch} 
        & \multicolumn{2}{c}{INet-v2}
        & \multicolumn{2}{c}{ObjNet} 
        \\
        {} & classes& &  \begin{small}R@1\end{small}  & \begin{small}R@5\end{small} &\begin{small}R@1\end{small}  & \begin{small}R@5\end{small} &\begin{small}R@1\end{small}  & \begin{small}R@5\end{small} &\begin{small}R@1\end{small}  & \begin{small}R@5\end{small} &\begin{small}R@1\end{small}  & \begin{small}R@5\end{small} &\begin{small}R@1\end{small}  & \begin{small}R@5\end{small}   \\
        \midrule
        \multirow{4}{*}{\small{ViT-B}}  & \multirow{2}{*}{\cmark} & \small{Original} & \textbf{82.0} & \textbf{96.0} & 14.8 & 35.2 & 32.9 & 47.4 & 36.8 & 58.4 & 71.3 & 89.8 & 36.0 & 55.8   \\
         & &\small{Ours} & \textbf{82.0} & 95.7 & \textbf{22.9} & \textbf{46.1} & \textbf{35.5} & \textbf{50.5} & \textbf{38.1} & \textbf{59.7}  & \textbf{71.4} & \textbf{89.9}& \textbf{43.5} & \textbf{64.6}   \\
        \cmidrule(r){2-15}
        & \multirow{2}{*}{\xmark} & \small{Original} & \textbf{81.0} & \textbf{95.9} & 17.2 & 38.7  & 34.7 & 49.6 & 34.0 & 56.4 & \textbf{70.9} & \textbf{90.1} & 33.9 & 57.0 \\
         & &\small{Ours} & 78.6 & 95.1 & \textbf{25.3} & \textbf{49.8} & \textbf{37.1} & \textbf{52.3} & \textbf{34.4}  & \textbf{57.4} & 68.6 & 88.9 & \textbf{40.3} & \textbf{65.7}   \\
         \midrule
         \multirow{4}{*}{\small{AR-B}}  & \multirow{2}{*}{\cmark} & \small{Original} & \textbf{83.8} & \textbf{97.2} & 25.3 & 51.6  & 39.8 & 56.9 & 44.7 & 66.5 & 74.4 & \textbf{92.3} & 39.8 & 64.9 \\
         & &\small{Ours} & 81.2 & 96.7  & \textbf{32.2} & \textbf{59.1} & \textbf{44.6} & \textbf{60.9} & \textbf{47.2}  & \textbf{68.7} & \textbf{74.9} & {92.2} & \textbf{45.2} & \textbf{70.9} \\
         \cmidrule(r){2-15}
        & \multirow{2}{*}{\xmark} & \small{Original} & \textbf{85.0} & \textbf{97.3} & 22.6 & 46.7 & 42.3 & 58.7 & 41.5 & 64.9 & \textbf{73.3} & \textbf{92.4} & 42.6 & 62.7  \\
         & &\small{Ours} & \textbf{85.0} & 97.2  & \textbf{30.3} & \textbf{54.9} & \textbf{44.8} & \textbf{62.2} & \textbf{42.0} & \textbf{66.1}  & {72.1} & {91.9} & \textbf{48.4} & \textbf{69.2} \\
        \midrule
        \multirow{4}{*}{\small{DeiT-B}}  & \multirow{2}{*}{\cmark} & \small{Original} & \textbf{81.7} & 94.4 & 11.7 & 29.4 & 29.6 & 42.9 & \textbf{32.0} & 49.3 & \textbf{70.3} & 86.5 &  32.7 & 48.8\\
         & &\small{Ours} & \textbf{81.7} & \textbf{95.2} & \textbf{15.2} & \textbf{37.5} & \textbf{30.9} & \textbf{45.5} & 31.5 & \textbf{49.6} & 69.4 & \textbf{88.4} & \textbf{36.6} & \textbf{55.3}\\
        \cmidrule(r){2-15}
        & \multirow{2}{*}{\xmark} & \small{Original} & \textbf{79.9} & 94.0 & 14.1 & 32.5 & 32.4 & 45.7 & \textbf{30.4} & 48.0 & \textbf{69.0} & 87.0 & 29.8 & 48.1 \\
         & &\small{Ours} & 79.3 & \textbf{94.7}  & \textbf{19.1} & \textbf{42.5} & \textbf{33.9} & \textbf{48.5} & {30.3}  & \textbf{48.8} & {68.9} & \textbf{88.3} & \textbf{34.8} & \textbf{57.3} \\
         \midrule
        \textbf{Avg.} & \cmark & & {0} & \dellarge{+0.1} & \dellarge{+6.4} & \dellarge{+9.1} & \dellarge{+2.9} & \dellarge{+3.2} & \dellarge{+1.1} & \dellarge{+1.3}& \dellarge{+0.5}& \dellarge{+0.6} & \dellarge{+5.7} & \dellarge{+7.3}\\
        \textbf{change} & \xmark & & {\color{red} -1.9} & {\color{red} -0.2} & \dellarge{+6.7} & \dellarge{+9.5}  & \dellarge{+2.1} & \dellarge{+3.0} & \dellarge{+0.3} & \dellarge{+1.0}& {\color{red} -1.2}& {\color{red} -0.1} & \dellarge{+5.6} & \dellarge{+8.0} \\
        \bottomrule
    \end{tabular}
    }
    \smallskip
    \smallskip
    \label{table:AB-test}
    \end{center}
\end{table}

\begin{table}[h!]
\caption{
    Robustness evaluation on the classes that were included in the training set of our finetuning and classes that were not for the synthetic datasets. The results are presented for the base models of ViT~\cite{dosovitskiy2020image}, ViT AugReg~\cite{Steiner2021HowTT} (AR), and DeiT~\cite{touvron2020training}. The last row indicates the average change for the training and non-training classes across the models for each dataset.
    }
    \begin{center}
    \begin{tabular}{@{}l@{~~~}c@{~~}c@{~~}c@{~}c@{~~}c@{~}c@{~~}c@{~}c@{}}
        \toprule
        \multirow{2}{*}{Model} & Train & &  \multicolumn{2}{c}{SI-loc.} 
        & \multicolumn{2}{c}{SI-rot.}
        & \multicolumn{2}{c}{SI-size} 
        \\
        {} & classes& &  \begin{small}R@1\end{small}  & \begin{small}R@5\end{small} &\begin{small}R@1\end{small}  & \begin{small}R@5\end{small} &\begin{small}R@1\end{small}  & \begin{small}R@5\end{small} \\
        \midrule
        \multirow{4}{*}{\small{ViT-B}}  & \multirow{2}{*}{\cmark} & \small{Original} &  34.2 & 52.5 & 40.5 & 60.3 & 56.5 & 76.1 \\
         & &\small{Ours} & \textbf{38.6}& \textbf{58.0} & \textbf{46.3} & \textbf{68.1}  & \textbf{60.4} & \textbf{80.4}  \\
        \cmidrule(r){2-9}
        & \multirow{2}{*}{\xmark} & \small{Original} & 32.0 & 51.8 & 37.1 & 55.6 & 54.4 & 76.3 \\
         & &\small{Ours} &  \textbf{38.6}& \textbf{57.5} & \textbf{46.0} & \textbf{65.6} & \textbf{61.9} & \textbf{82.9} \\
         \midrule
        \multirow{4}{*}{\small{AR-B}}  & \multirow{2}{*}{\cmark} & \small{Original} & 41.2 & 60.7 & 50.0 & 69.2 & 61.6 & 79.9 \\
         & &\small{Ours} & \textbf{43.8}& \textbf{62.7} & \textbf{55.5} & \textbf{75.3} & \textbf{64.7} & \textbf{83.6} \\
         \cmidrule(r){2-9}
        & \multirow{2}{*}{\xmark} & \small{Original} & 39.7 & 60.8 & 45.5 & 67.0 & 59.1 & 81.0 \\
         & &\small{Ours} & \textbf{42.5}& \textbf{62.9} & \textbf{51.9} & \textbf{73.7} & \textbf{63.2} & \textbf{84.4} \\
        \midrule
        \multirow{4}{*}{\small{DeiT-B}}  & \multirow{2}{*}{\cmark} & \small{Original} & 35.2 & 54.2 & 40.3 & 56.3 & 56.1 & 72.3 \\
         & &\small{Ours} & \textbf{37.1} & \textbf{56.6} & \textbf{43.5} & \textbf{61.6} & \textbf{59.0} & \textbf{77.0}\\
         \cmidrule(r){2-9}
        & \multirow{2}{*}{\xmark} & \small{Original} & 33.6 & 55.1 & 38.0 & 56.1 & 52.5 & 74.8 \\
         & &\small{Ours} & \textbf{35.8} & \textbf{57.5} & \textbf{42.1} & \textbf{61.4} & \textbf{56.7} & \textbf{79.7} \\
         \midrule
        \textbf{Avg.} & \cmark & & \dellarge{+3.0} & \dellarge{+3.3} & \dellarge{+4.8} & \dellarge{+6.4}& \dellarge{+3.3}& \dellarge{+4.2}\\
        \textbf{change} & \xmark & & \dellarge{+3.7} & \dellarge{+3.4} & \dellarge{+6.5} & \dellarge{+7.3}& \dellarge{+5.3}& \dellarge{+5.0} \\
        \bottomrule
    \end{tabular}
    \smallskip
    \smallskip
    \label{table:AB-test-SI}
    \end{center}
\end{table}

\clearpage

\section{Ablation study}
\label{sec:ablation}
Tab.~\ref{table:ablation_r5} presents the top-5 accuracy results of our ablation study to complement Tab.~\ref{table:ablation}. As can be seen, the top-5 results are consistent with the top-1 results from Tab.~\ref{table:ablation}.
\begin{table}[h!]
            \caption{Ablation study for our method on the ViT~\cite{dosovitskiy2020image} and DeiT~\cite{touvron2020training} base models. The table presents top-5
            accuracy. We check the impact of each term of our loss on the robustness of the model, as well as the choice of confidence boosting versus cross-entropy with the ground-truth label (labeled as w/ ground-truth).
    }
    \begin{center}
    \resizebox{\linewidth}{!}{
    \begin{tabular}{@{}l@{~~}c@{~~~~}c@{~~~~}c@{~~~~}c@{~~~~}c@{~~~~}c@{~~~~}c@{~~~~}c@{~~~~}c@{~~~~}c@{}}
        \toprule
            {Model} & {Method} &{INet val} & {INet-A} & {INet-R}& {Sketch} 
        & {INet-v2}
        & {ObjNet} & {SI-loc.} & {SI-rot.}  & {SI-size} 
        \\
        \midrule
        \multirow{6}{*}{\small{\begin{turn}{90} ViT-B \end{turn}}}  &\small{Original} & 96.0 & 37.0 & 48.5 & 57.4 & 89.9 & 56.4 & 52.2 & 58.3 & 76.2  \\
        &\small{Ours} & 95.4 & \textbf{48.0} & \textbf{51.4} & \textbf{58.5} & 89.4 & \textbf{65.1} & 57.8 & \thanks{67.0} & \textbf{81.4}  \\
        &\small{w/o $\mathcal{L_{\text{classification}}}$ (Eq.~\ref{eq:cls-loss}}) & 94.3 & 38.9 & 50.3 & 57.9 & 87.4 & 61.6 & 56.2 & 63.8 & 78.9 \\
        &\small{w/o $\mathcal{L_{\text{bg}}}$ (Eq.~\ref{eq:background})} &  \textbf{96.1} & 41.8 & 50.9 & \textbf{58.5} & \textbf{90.5} & 62.1 & 54.1 & 62.7 & 78.9  \\
        &\small{w/o $\mathcal{L_{\text{fg}}}$ (Eq.~\ref{eq:foreground})} & 95.3 & 47.2 & 48.7 & 57.0 & 89.2 & 64.4 & \textbf{58.3} & 66.2 & 81.2  \\
        &\small{w/ ground-truth} & \textbf{96.1} & 45.2 & 50.4 & 57.9 & 90.4 & 63.2 & 57.1 & 64.8 & 80.3  \\
        \midrule
        \multirow{6}{*}{\small{\begin{turn}{90} DeiT-B \end{turn}}} & \small{Original} &  94.2 & 31.0 & 44.2 & 48.6 & 86.8 & 48.5 & 54.6 & 56.3 & 73.4\\
        & \small{Ours}  & 94.9 & 40.0 & 47.0 & 49.2 & 88.3 & 56.2 & 57.0 & 61.5 & 78.2\\
        &\small{w/o $\mathcal{L_{\text{classification}}}$ (Eq.~\ref{eq:cls-loss}})  & \textbf{95.2} & 37.4 & \textbf{48.4} & \textbf{51.4} & \textbf{88.9} & \textbf{56.8} & \textbf{60.1} & \textbf{63.1} & \textbf{79.5} \\
        &\small{w/o $\mathcal{L_{\text{bg}}}$ (Eq.~\ref{eq:background})}  & \textbf{95.2} & 34.4 & 44.7 & 47.9 & 88.3 & 52.4 & 53.2 & 57.5 & 74.7 \\
        &\small{w/o $\mathcal{L_{\text{fg}}}$ (Eq.~\ref{eq:foreground})} & 94.9 & \textbf{40.1} & 47.2 & 49.3 & 88.1 & 56.4 & 57.6 & 61.4 & 78.5 \\
        &\small{w/ ground-truth} & 95.0 & 39.6 & 46.5 & 49.6 & 88.2 & 55.9 & 57.3 & 61.2 & 78.0  \\
        \bottomrule
    \end{tabular}
    }
    \smallskip
    \smallskip
    \label{table:ablation_r5}
    \end{center}
    \vspace{-24px}
\end{table}

\clearpage

\section{Per class robustness analysis}
\label{sec:limitations}

There is considerable variability in the effect of our method between the different classes. This is similar to the varying effect of regularization reported by Balestriero et al.~\cite{balestriero2022effects} and it is reasonable to assume that it is a common phenomenon among regularization techniques.

\begin{figure}[!htb]
    \centering
    \includegraphics[width=1\textwidth]{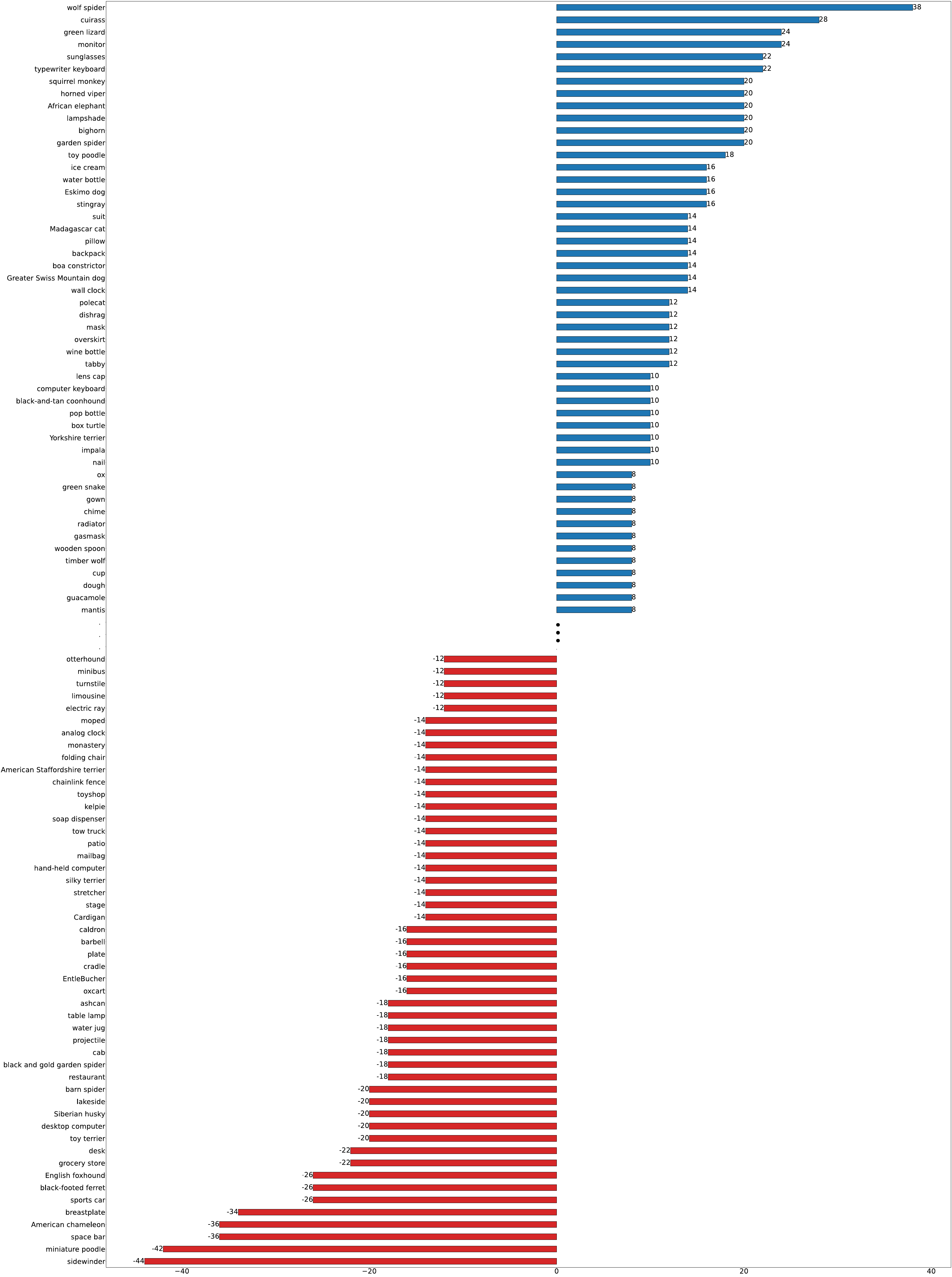}
    \caption{The 50 classes which most benefited and the 50 classes that were most harmed by the finetune procedure for ImageNet.}
    \label{fig:bar-val}
\end{figure}

\begin{figure}[!htb]
    \centering
    \includegraphics[width=1\textwidth]{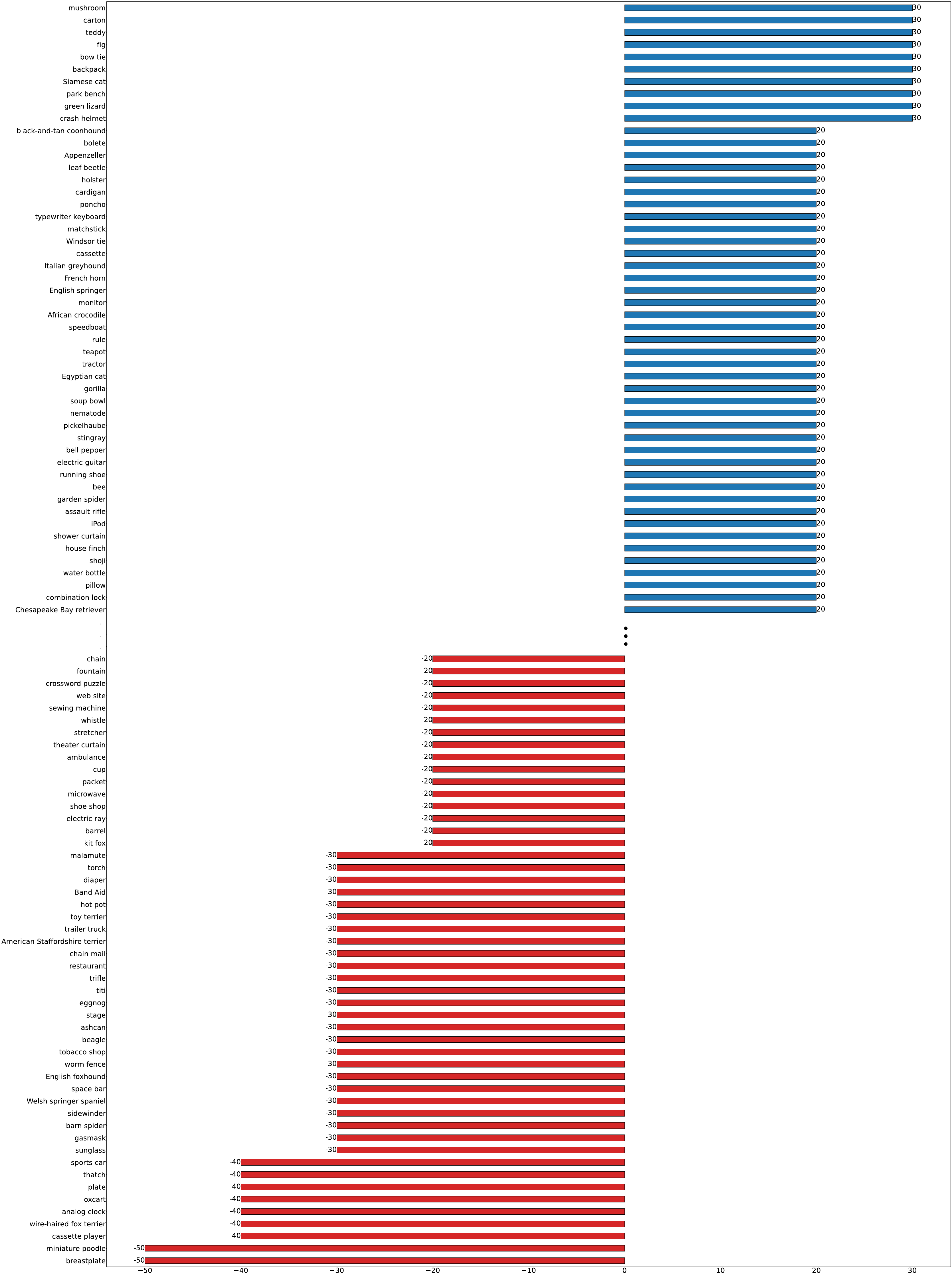}
    \caption{The 50 classes which most benefited and the 50 classes that were most harmed by the finetune procedure for ImageNet-v2.}
    \label{fig:bar-v2}
\end{figure}

\begin{figure}[!htb]
    \centering
    \includegraphics[width=1\textwidth]{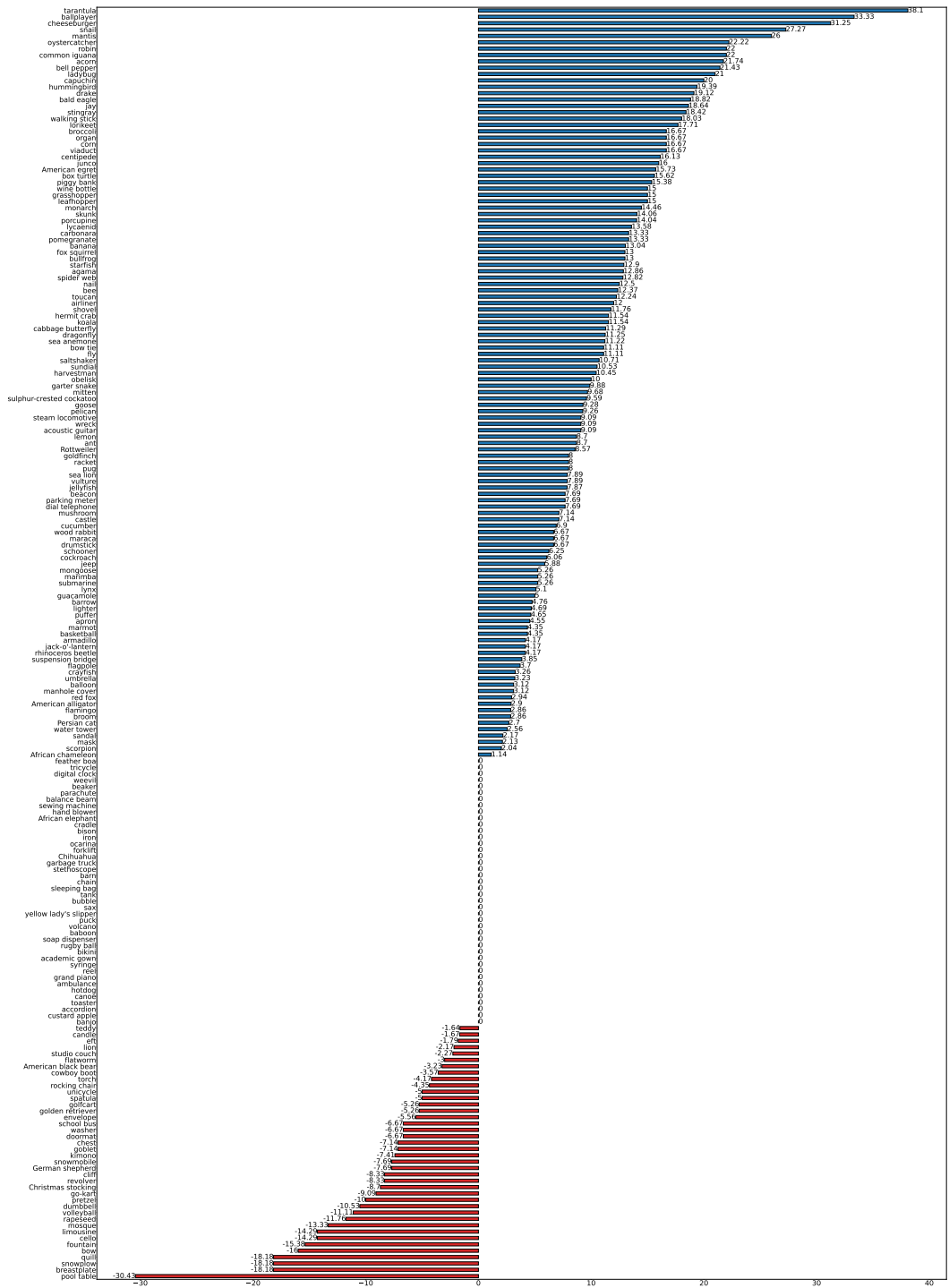}
    \caption{The effect of the finetune procedure on each class for ImageNet-A (there are some classes with zero effect).}
    \label{fig:bar-a}
\end{figure}

\begin{figure}[!htb]
    \centering
    \includegraphics[width=1\textwidth]{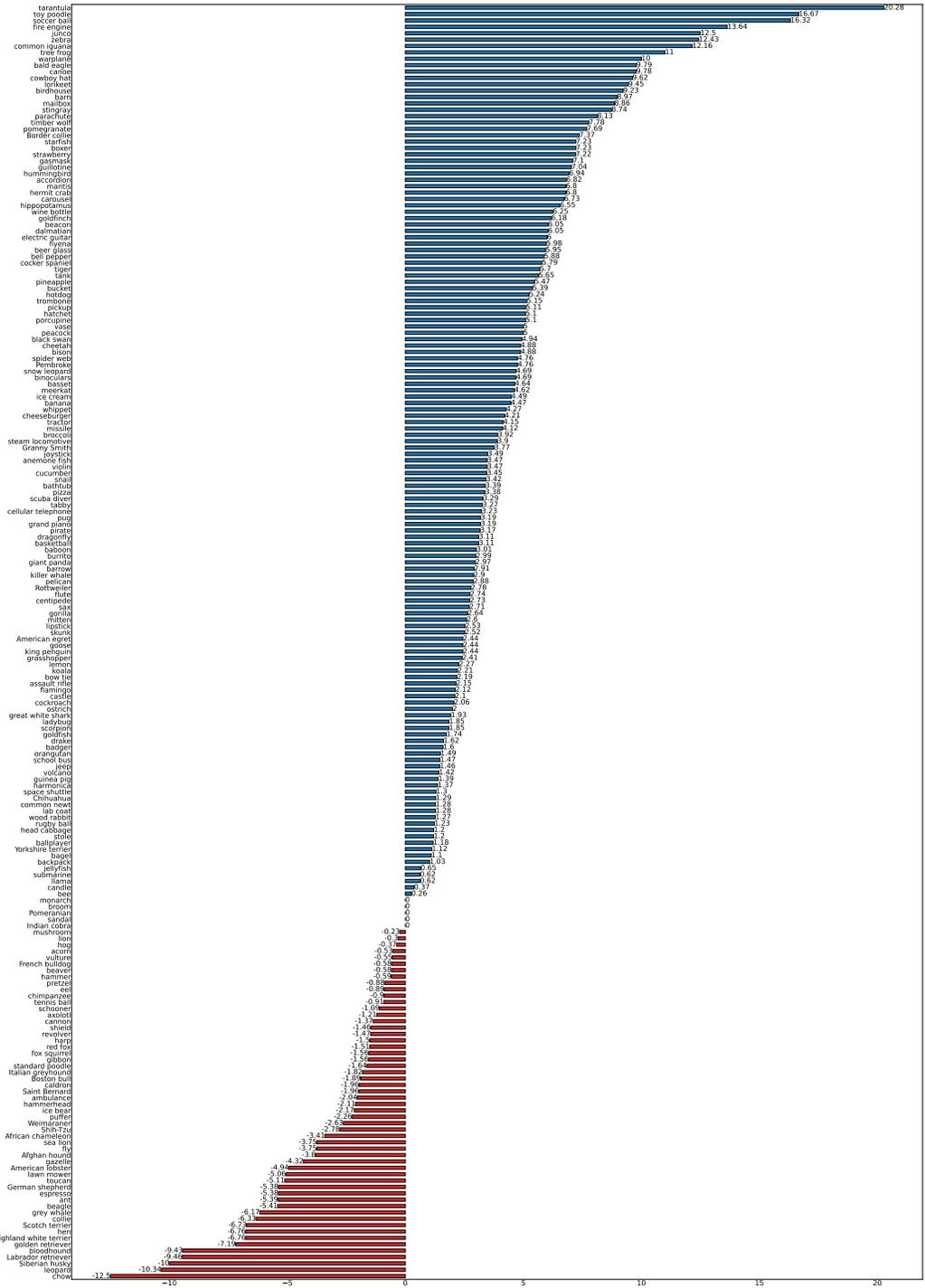}
    \caption{The effect of the finetune procedure on each class for ImageNet-R (there are some classes with zero effect).}
    \label{fig:bar-r}
\end{figure}

\begin{figure}[!htb]
    \centering
    \includegraphics[width=1\textwidth]{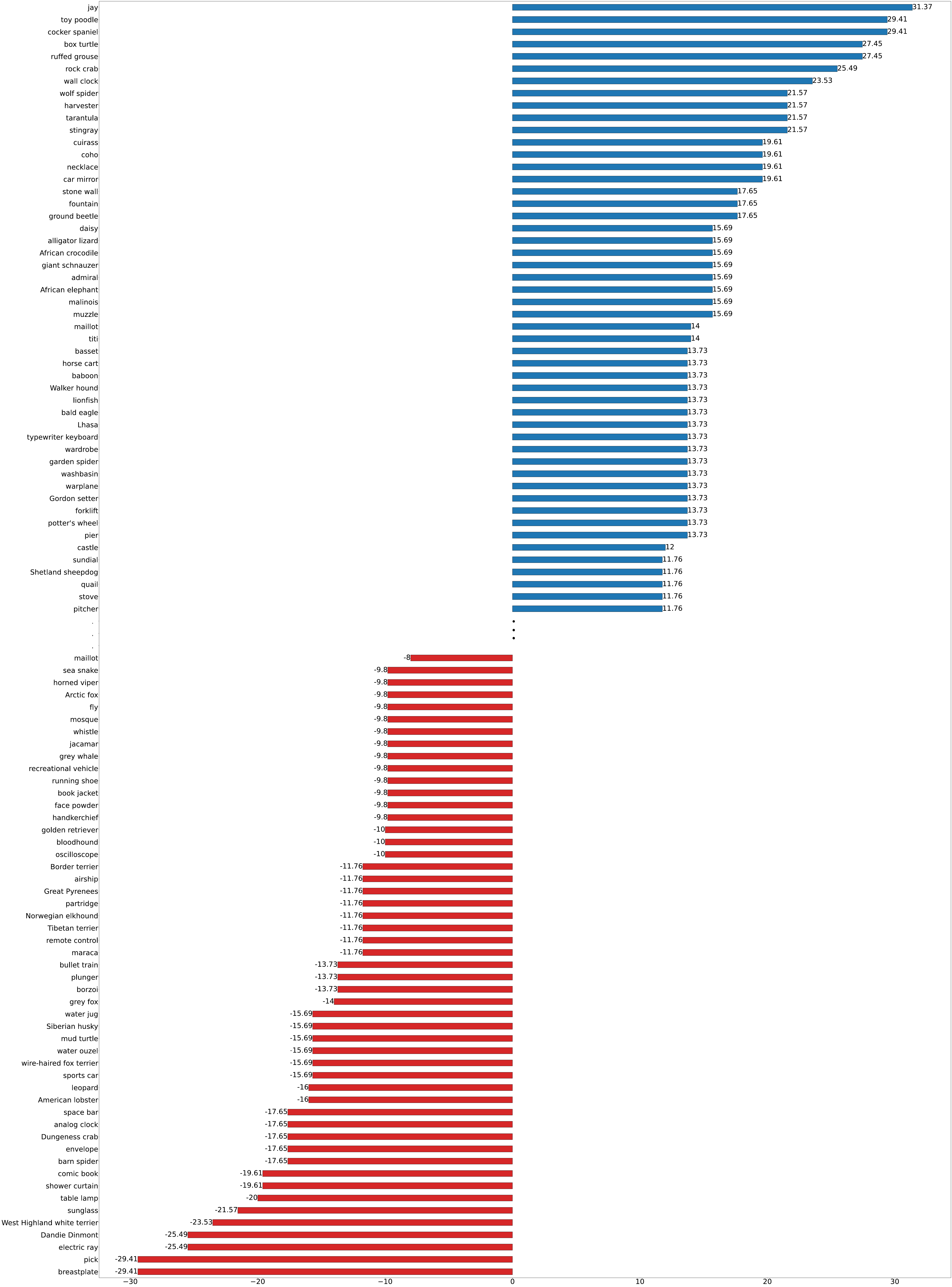}
    \caption{The 50 classes which most benefited and the 50 classes that were most harmed by the finetune procedure for ImageNet-sketch.}
    \label{fig:bar-sketch}
\end{figure}

\begin{figure}[!htb]
    \centering
    \includegraphics[width=1\textwidth]{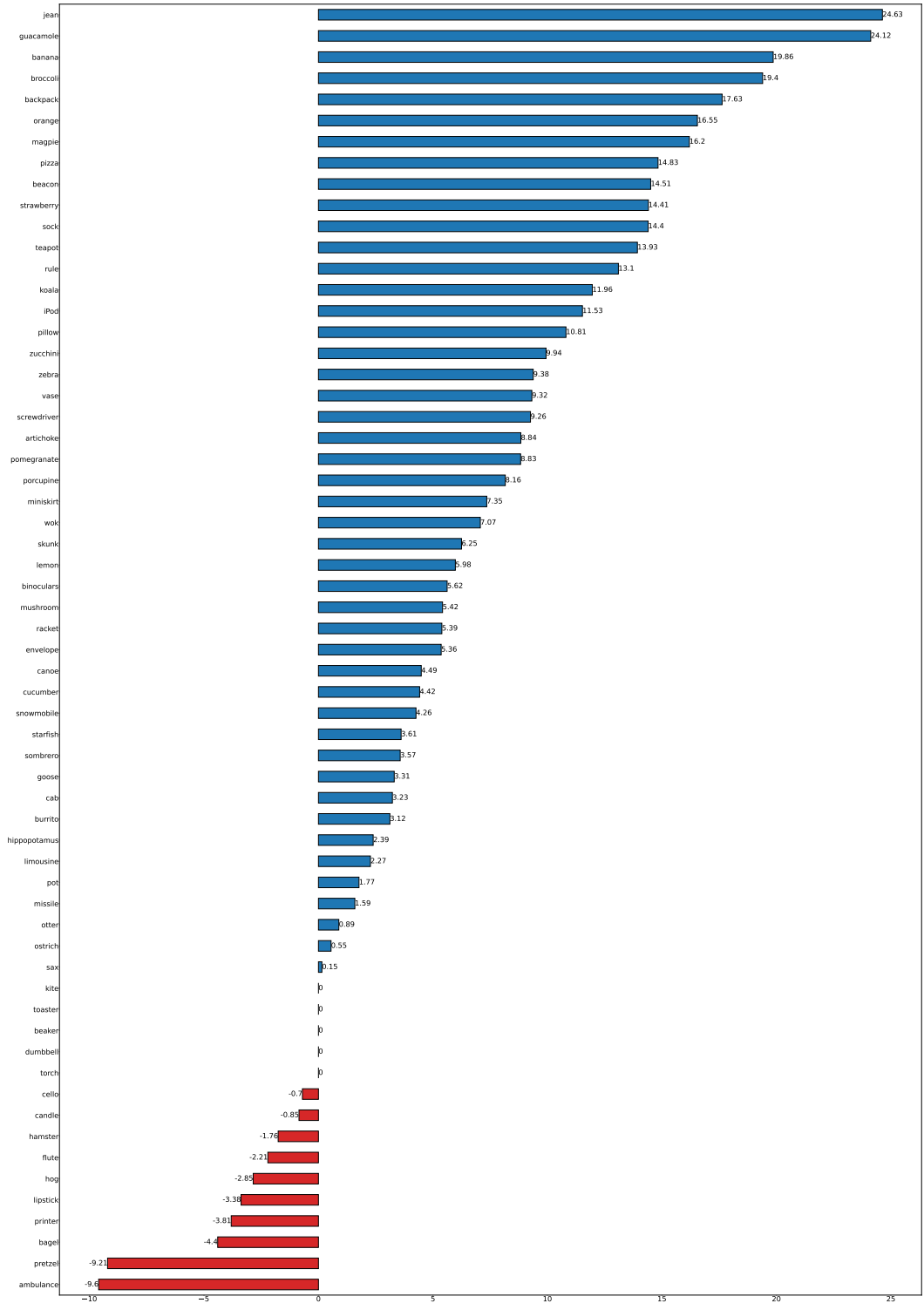}
    \caption{The effect of the finetune procedure on each class for SI-rotation (there are some classes with zero effect).}
    \label{fig:bar-rot}
\end{figure}

\begin{figure}[!htb]
    \centering
    \includegraphics[width=1\textwidth]{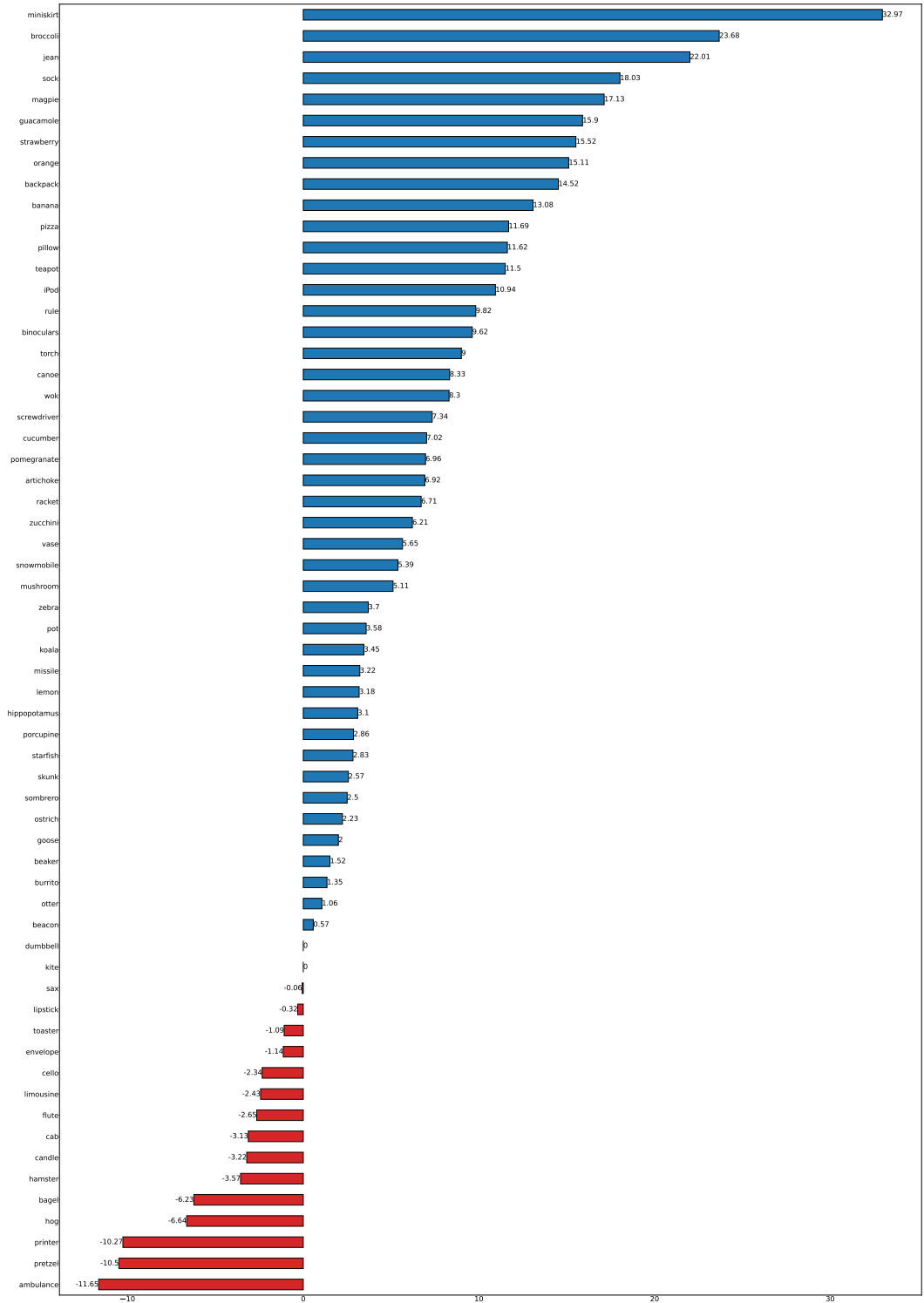}
    \caption{The effect of the finetune procedure on each class for SI-size (there are some classes with zero effect).}
    \label{fig:bar-size}
\end{figure}

Fig.~\ref{fig:bar-val}-\ref{fig:bar-size} depict the classes that benefited the most by our finetune process on ViT-B and those that were harmed the most. For datasets with $1,000$  classes, we present the $50$ classes with the most beneficial or harmful change for readability. Other datasets are presented with all their classes.

Inspecting the classes with the largest amount of change, we can observe a few trends. In some cases, classes with relatively small objects considerably benefit, and classes with objects that either reside in the background or benefit from the context are harmed, as can be expected. In some cases, classes with a small number of test samples, regardless of the type of object, lead to a higher absolute change, due to statistical reasons.

In order to study the limitations of our method, we inspect the classes where it fails the most for INet-A and INet-V2. These two test sets represent different amounts of domain shift from the original ImageNet (INet-V2 is very close to ImageNet, while INet-A is out-of-distribution). As shown in the paper, our method increases robustness to domain shifts. 

\begin{figure*}[h]
  \centering
  \resizebox{\linewidth}{!}{
\begin{tabular}{c@{~~}c@{~}c@{~}c@{~~~}c@{~}c@{~}c@{~~~}c@{~}c@{~}c}
& \small{Input} & \small {Original} & {\small{Ours}} &  \small{Input} & \small {Original} & {\small{Ours}} &  \small{Input} & \small {Original} & {\small{Ours}} \\
{\begin{turn}{90}~~~ INet-A \end{turn}} & 
\includegraphics[width=0.1\linewidth, clip]{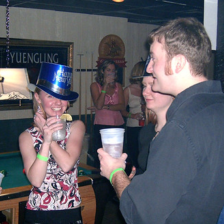}&
\includegraphics[width=0.1\linewidth, clip]{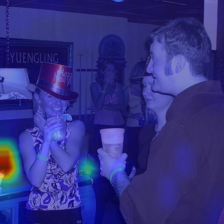}&
\includegraphics[width=0.1\linewidth, clip]{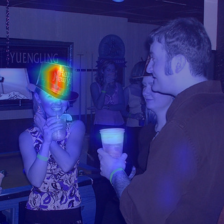}&
\includegraphics[width=0.1\linewidth, clip]{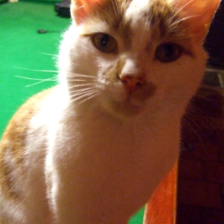}&
\includegraphics[width=0.1\linewidth, clip]{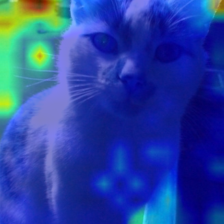}&
\includegraphics[width=0.1\linewidth, clip]{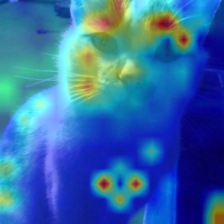}&
\includegraphics[width=0.1\linewidth, clip]{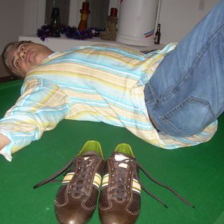}&
\includegraphics[width=0.1\linewidth, clip]{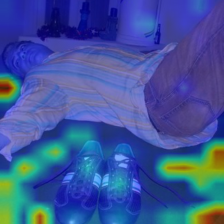}&
\includegraphics[width=0.1\linewidth, clip]{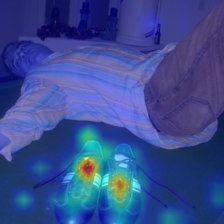}
\\
\multirow{2}{*}{\begin{turn}{90} Pred \end{turn}}
&
&{\small{Pool-}} & {\small{Bucket}} & & {\small{Pool-}} & {\small{Egyptian-}} &  & {\small{Pool-}} & {\small{Running-}}\\
& & {\small{table}} & {\small{}} &  & {\small{table}} & {\small{cat}} &  & {\small{table}} & {\small{shoe}}  \\
{\begin{turn}{90}~~~ INet-A \end{turn}} & 
\includegraphics[width=0.1\linewidth, clip]{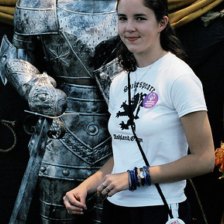}&
\includegraphics[width=0.1\linewidth, clip]{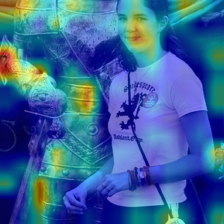}&
\includegraphics[width=0.1\linewidth, clip]{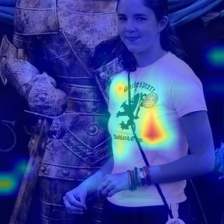}&
\includegraphics[width=0.1\linewidth, clip]{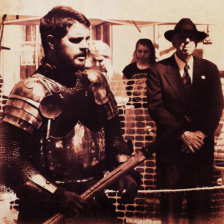}&
\includegraphics[width=0.1\linewidth, clip]{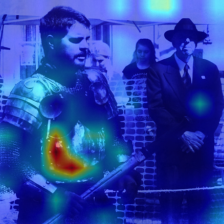}&
\includegraphics[width=0.1\linewidth, clip]{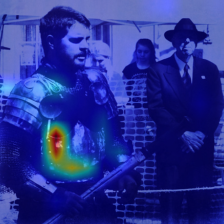}
\\
\multirow{2}{*}{\begin{turn}{90} Pred \end{turn}}
&
&{\small{Breastplate}} & {\small{T-shirt}} & & {\small{Breastplate}} & {\small{Cuirass}} &  & \\
& & & & &  {\small{}} & {\small{}}  \\

{\begin{turn}{90}~ INet-v2 \end{turn}} & 
\includegraphics[width=0.1\linewidth, clip]{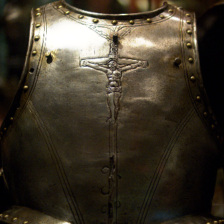}&
\includegraphics[width=0.1\linewidth, clip]{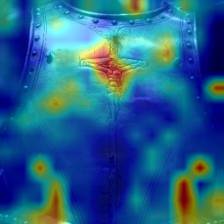}&
\includegraphics[width=0.1\linewidth, clip]{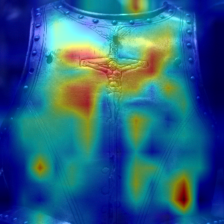}&
\includegraphics[width=0.1\linewidth, clip]{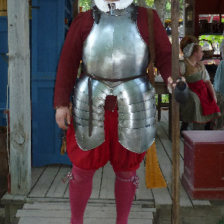}&
\includegraphics[width=0.1\linewidth, clip]{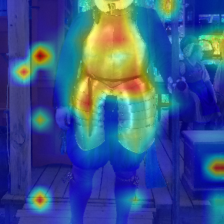}&
\includegraphics[width=0.1\linewidth, clip]{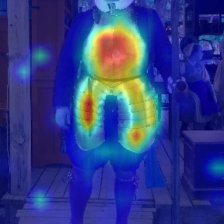}&
\includegraphics[width=0.1\linewidth, clip]{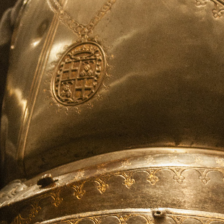}&
\includegraphics[width=0.1\linewidth, clip]{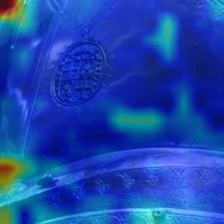}&
\includegraphics[width=0.1\linewidth, clip]{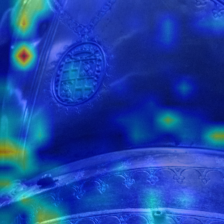}
\\
\multirow{2}{*}{\begin{turn}{90} Pred \end{turn}}
&
&{\small{Breastplate}} & {\small{Cuirass}} & & {\small{Breastplate}} & {\small{Cuirass}} &  & {\small{Breastplate}} & {\small{Cuirass}}\\
& & & & & & {\small{}} & & 
\\

{\begin{turn}{90}~ INet-v2 \end{turn}} & 
\includegraphics[width=0.1\linewidth, clip]{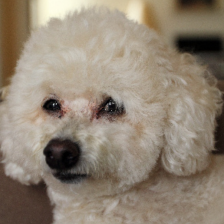}&
\includegraphics[width=0.1\linewidth, clip]{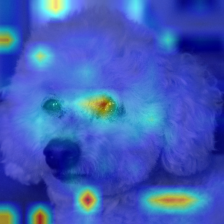}&
\includegraphics[width=0.1\linewidth, clip]{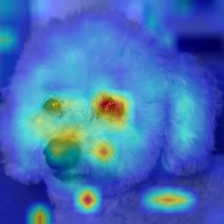}&
\includegraphics[width=0.1\linewidth, clip]{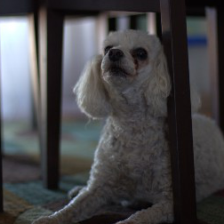}&
\includegraphics[width=0.1\linewidth, clip]{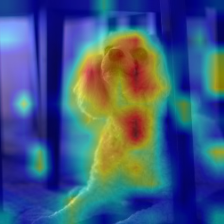}&
\includegraphics[width=0.1\linewidth, clip]{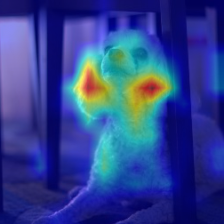}&
\includegraphics[width=0.1\linewidth, clip]{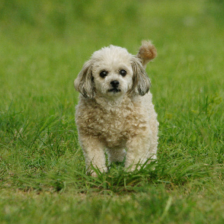}&
\includegraphics[width=0.1\linewidth, clip]{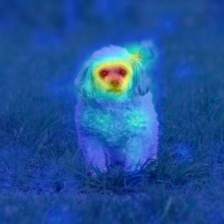}&
\includegraphics[width=0.1\linewidth, clip]{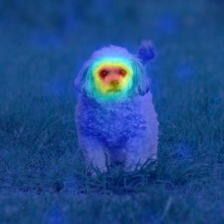}
\\
\multirow{2}{*}{\begin{turn}{90} Pred \end{turn}}
&
&{\small{Miniature-}} & {\small{Toy-}} & & {\small{Miniature-}} & {\small{Toy-}} &  & {\small{Miniature-}} & {\small{Lhasa-}}\\
& & {\small{poodle}}& {\small{poodle}} & &{\small{poodle}} & {\small{poodle}} & & {\small{poodle}} & {\small{apso}} 

\end{tabular}
}
\caption{Examples of the $2$ classes most harmed by our method for a dataset that is out of distribution (INet-A), and a dataset that has a similar distribution to ImageNet (INet-v2). For each dataset, we show the first three examples where our method modified a correct prediction and made it wrong (by showing the first three samples, we demonstrate that these samples are typical and not cherry-picked). As can be seen, in most cases the heatmaps are improved by our method, and the wrong prediction has a rationale (see text for examples). For INet-A there are only $2$ mistakes in the Breastplate class (out of a total of $11$ examples in the dataset), therefore the corresponding row (second row) only contains 2 examples.
\label{fig:per_class}}
\end{figure*}

Fig.~\ref{fig:per_class} depicts the two classes, out of each of the two test sets (INet-A and INet-V2) that were harmed the most: pool table and breastplate for INet-A and breastplate and miniature poodle for INet-v2. From each class, we show the effect of our method on the first three samples in which a correct classification before finetuning turned into a wrong classification (so we avoid cherry-picking). 

The pool table samples tend to show other objects in the foreground. In the top-left sample, the object is identified as a bucket (it is a hat, which is the 2nd top prediction). In the other two samples for pool tables and INet-A, the foreground object is correctly identified. In all three cases, the heatmap explains the result. In the breastplate case, there are only 11 test samples, two of which were wrongly classified by our method as other, reasonable classes. 

For INet-v2, following our method, some breastplates are identified as cuirass (oxford dictionary: a piece of armor consisting of breastplate and backplate fastened together), which is a similar class. In the 2nd most affected class, miniature poodles are identified as toy poodles, or as a Lhasa Apso, which are similar dog breeds.


\end{document}